%% file: main.tex
\documentclass[12pt,english]{article}

\input{newcmnds}

\allowdisplaybreaks

\begin{document}
	\title{Toeplitz Least Squares Problems, \\ Fast Algorithms and Big Data}
	\author{
		Ali Eshragh\thanks{School of Information and Physical Sciences, University of Newcastle, NSW, Australia, and International Computer Science Institute, Berkeley, CA, USA. Email:  \tt{ali.eshragh@newcastle.edu.au}}
		\and
		Oliver Di Pietro\thanks{School of Information and Physical Sciences, University of Newcastle, NSW, Australia. Email:  \tt{Oliver.DiPietro@uon.edu.au}} 
		\and 
		Michael A. Saunders\thanks{Department of Management Science and Engineering, Stanford University, CA, USA. Email: \tt{saunders@stanford.edu}}
	}
	\date{\today -- Version $1$}
	\maketitle

%%%%%%%%%%%% Abstract %%%%%%%%%%%%%%

\input{abstract}

%----------------------
%\section{Introduction}
%----------------------
\input{1Introduction}

%------------------------------------------
%\section{Ordinary Least Squares Problems}
%------------------------------------------
\input{2OLS_Problems}

%------------------------------------------
%\section{Toeplitz OLS Problems}
%------------------------------------------
\input{3Toeplitz_OLS_Problems}
\input{4numerical_results}

%------------------------------------------
%\section{Conclusion}
%------------------------------------------
\input{5Conclusion}

%%%%%%%%%%%%%%%% References %%%%%%%%%%%%%%%%%%%%%%%%%%%%
\footnotesize
\bibliographystyle{plain}
\bibliography{Biblio}

\end{document}

%% file: newcmnds.tex
\usepackage{algorithm}
\usepackage{booktabs}
\usepackage{graphicx}
\usepackage{amsmath}
\usepackage{amssymb}
\usepackage{latexsym}
\usepackage{crop}
\usepackage{algorithmic,algorithm}
\usepackage{multirow}
\usepackage{bm}
\usepackage{bbm}
\usepackage{enumerate}
\usepackage{url}
\usepackage{array}
\usepackage{paralist}
\usepackage{diagbox}
\usepackage{wasysym}
\usepackage{booktabs}
\usepackage[dvipsnames]{xcolor}
\usepackage[colorlinks = true, pdfstartview = FitV, linkcolor = blue, citecolor = blue, urlcolor = blue]{hyperref}

\usepackage{listings}

\usepackage[utf8]{inputenc}
\usepackage{rotating}

\usepackage[capitalise]{cleveref}
\crefname{equation}{}{}
\crefname{figure}{Figure}{Figures}
\creflabelformat{equation}{\textup{(#2#1#3)}}
\crefname{assumption}{Assumption}{Assumptions}
\crefname{condition}{Condition}{Conditions}

\usepackage{xspace}

\renewcommand\th{\textsuperscript{th}\xspace}
\usepackage{fullpage}
\usepackage{multirow}

\usepackage[sort,nocompress]{cite}
\usepackage{arydshln}
\usepackage{enumitem}
\setlist[enumerate,1]{leftmargin=*,wide=0em, noitemsep,nolistsep, label = {\bfseries \arabic*.}}
\setlist[itemize,1]{leftmargin=*,wide=0em, noitemsep,nolistsep}
\newlist{abbrv}{itemize}{1}
\setlist[abbrv,1]{label=,labelwidth=1in,align=parleft,itemsep=0.1\baselineskip,leftmargin=!}
\usepackage{pifont}% http://ctan.org/pkg/pifont
\newcommand {\W}  {\mathcal{W}}

\makeatletter
\newcommand*{\transpose}{%
	{\mathpalette\@transpose{}}%
}
\newcommand*{\@transpose}[2]{%
	\raisebox{\depth}{$\m@th#1\intercal$}%
}
\makeatother

\renewcommand {\AA}  { {\bm{A}} }
\newcommand {\BB}  { {\bm{B}} }
\newcommand {\CC}  { {\bm{C}} }
\newcommand {\DD}  { {\bm{D}} }

\newcommand {\GG}  { {\bm{G}} }
\newcommand {\HH}  { {\bm{H}} }
\newcommand {\VV}  { {\bm{V}} }

\newcommand {\bSS}  { {\bm{S}} }
\newcommand {\TT}  { {\bm{T}} }

\newcommand {\XX}  { {\bm{X}} }

\newcommand {\zz}  { {\bm z} }

\newcommand {\xx}  { {\bm x} }
\newcommand {\yy}  { {\bm y} }
\newcommand {\rr}  { {\bm r} }

\newcommand{\R}{{\rm I\!R}}

\newcommand {\vv}  { {\bm v} }
\newcommand {\ww}  { {\bm w} }

\newcommand {\bb}  { {\bm b} }

\newcommand {\ee}  { {\bm e} }

\renewcommand{\Pr}{\hbox{\bf{Pr}}}

\newcommand*\xbar[1]{%
	\hbox{%
		\vbox{%
			\hrule height 0.5pt % The actual bar
			\kern0.5ex%         % Distance between bar and symbol
			\hbox{%
				\kern-0.1em%      % Shortening on the left side
				\ensuremath{#1}%
				\kern-0.1em%      % Shortening on the right side
			}%
		}%
	}%
} 

\definecolor{forestgreen}{rgb}{0.13, 0.55, 0.13}

\definecolor{amber}{rgb}{1.0, 0.75, 0.0}

\definecolor{bananayellow}{rgb}{.8, 0.6, 0}

\newcounter{comment}

\usepackage{amsthm}
\usepackage[framemethod=TikZ]{mdframed}

\mdfdefinestyle{theoremstyle}{%
	linewidth = 1pt,%
	roundcorner = 10pt,%
	leftmargin = 0,%
	rightmargin = 0,%
	backgroundcolor = cyan!3,%
	outerlinecolor = magenta!70!black,%
	splittopskip = \topskip,%
	ntheorem = true,%
}
\mdtheorem[style=theoremstyle]{claim}{Claim}

\newmdtheoremenv[%
linewidth = 1pt,%
roundcorner = 10pt,%
leftmargin = 0,%
rightmargin = 0,%
backgroundcolor = green!3,%
outerlinecolor = blue!70!black,%
splittopskip = \topskip,%
ntheorem = true,%
]{theorem}{Theorem}

\newmdtheoremenv[%
linewidth = 1pt,%
roundcorner = 10pt,%
leftmargin = 0,%
rightmargin = 0,%
backgroundcolor = green!3,%
outerlinecolor = blue!70!black,%
splittopskip = \topskip,%
ntheorem = true,%
]{corollary}{Corollary}

\newmdtheoremenv[%
linewidth = 1pt,%
roundcorner = 10pt,%
leftmargin = 0,%
rightmargin = 0,%
backgroundcolor = green!3,%
outerlinecolor = blue!70!black,%
splittopskip = \topskip,%
ntheorem = true,%
]{lemma}{Lemma}

\newmdtheoremenv[%
linewidth = 1pt,%
roundcorner = 10pt,%
leftmargin = 0,%
rightmargin = 0,%
backgroundcolor = yellow!3,%
outerlinecolor = blue!70!black,%
splittopskip = \topskip,%
ntheorem = true,%
]{definition}{Definition}

\newmdtheoremenv[%
linewidth = 1pt,%
roundcorner = 10pt,%
leftmargin = 0,%
rightmargin = 0,%
backgroundcolor = green!3,%
outerlinecolor = blue!70!black,%
splittopskip = \topskip,%
ntheorem = true,%
]{proposition}{Proposition}

\newmdtheoremenv[%
linewidth = 1pt,%
roundcorner = 10pt,%
leftmargin = 0,%
rightmargin = 0,%
backgroundcolor = green!3,%
outerlinecolor = blue!70!black,%
splittopskip = \topskip,%
ntheorem = true,%
]{condition}{Condition}

\newmdtheoremenv[%
linewidth = 1pt,%
roundcorner = 10pt,%
leftmargin = 0,%
rightmargin = 0,%
backgroundcolor = green!3,%
outerlinecolor = blue!70!black,%
splittopskip = \topskip,%
ntheorem = true,%
]{assumption}{Assumption}

\theoremstyle{definition}
\newmdtheoremenv[%
linewidth = 1pt,%
roundcorner = 10pt,%
leftmargin = 0,%
rightmargin = 0,%
backgroundcolor = blue!3,%
outerlinecolor = blue!70!black,%
splittopskip = \topskip,%
ntheorem = true,%
]{example}{Example}

\theoremstyle{definition}
\newmdtheoremenv[%
linewidth = 1pt,%
roundcorner = 10pt,%
leftmargin = 0,%
rightmargin = 0,%
backgroundcolor = red!3,%
outerlinecolor = blue!70!black,%
splittopskip = \topskip,%
ntheorem = true,%
]{remark}{Remark}

\usepackage{tikz}
\usepackage{xparse}% So that we can have two optional parameters

\NewDocumentCommand\DownArrow{O{2.0ex} O{black}}{%
	\mathrel{\tikz[baseline] \draw [<-, line width=0.5pt, #2] (0,0) -- ++(0,#1);}
}

\usepackage{listings} % to inser code

\definecolor{mygreen}{rgb}{0,0.6,0}
\definecolor{mygray}{rgb}{0.5,0.5,0.5}
\definecolor{mymauve}{rgb}{0.58,0,0.82}

\newcommand*\dotprod[1]{\left\langle #1\right\rangle}

\newcommand {\Ex} { {\mathbb E} }

\newcommand*\bigO[1]{\mathcal O\left( #1\right)}

\usepackage{dsfont}

\usepackage{amsmath}
\usepackage{amsfonts}
\usepackage{amssymb}
\usepackage{makeidx}
\usepackage{graphicx}
\usepackage{caption}
\usepackage{subcaption}
\usepackage{framed}
\usepackage{booktabs,array}
\usepackage{xcolor}

\usepackage{algorithmic}

\newcommand {\innd}  { {\in \R^{n \times d}} }
\DeclareDocumentCommand{\inn}{mgO{\R}}
  {%
  \IfNoValueTF{#2}
    {\in #3^{#1}}
    {\in #3^{#1 \times #2}}%
}

%% file: Abstract.tex
\begin{abstract}
	In time series analysis, when fitting an autoregressive model, one must solve a Toeplitz ordinary least squares problem numerous times to find an appropriate model, which can severely affect computational times with large data sets. Two recent algorithms (\emph{LSAR} and \emph{Repeated Halving}) have applied randomized numerical linear algebra (RandNLA) techniques to fitting an autoregressive model to big time-series data. We investigate and compare the quality of these two approximation algorithms on large-scale synthetic and real-world data. While both algorithms display comparable results for synthetic datasets, the LSAR algorithm appears to be more robust when applied to real-world time series data. %Overall this paper displays the effectiveness of RandNLA in a big-data regime for a time-series context. 
    We conclude that RandNLA is effective in the context of big-data time series.
\end{abstract}

%% file: 1Introduction.tex
\section{\bf Introduction}

Advancements in technology and computation have led to enormous data sets being generated from various fields of research including science, internet datasets and business. These data sets, commonly described as Big Data, are stored in the form of vectors and matrices, allowing us to draw on our knowledge of linear algebra to analyze them. The enormity of Big Data matrices  has mandated the search for large-scale matrix algorithms with improved run times and stability \cite{drineas2016randnla}.

Randomised Numerical Linear Algebra (RandNLA) is a new tool to deal with big data \cite{mahoney2011randomized}. RandNLA utilises random sampling of elements, rows, or columns of a matrix to produce a second smaller matrix that is similar to the first matrix in some way, yet computationally easier to deal with (e.g., \cite{drineas2016randnla,drineas2012fast,drineas2017lecture,clarkson2017low,woodruff2014sketching}). An application of RandNLA is in finding fast solutions for Toeplitz least squares problems (e.g., \cite{eshragh2019lsar,shi2019sublinear}).

Toeplitz matrices and Toeplitz least squares problems occur in many practical large-scale matrix problems such as time series analysis and signal/image processing \cite{drineas2016randnla, eshragh2019lsar}. In practice they can become a computational bottleneck. In the context of stochastic dynamic systems, autoregressive models require the solutions of many ordinary least squares problems with Toeplitz structure \cite{eshragh2021rollage}. Such stochastic dynamic models have a wide range of applications, from supply chains \cite{fahimnia2018cor,fahimnia2015ijpe,fahimnia2015omega,abolghasemi2020ijpe} and energy systems (e.g., \cite{eshragh2021enmo,eshragh2011es}) to epidemiology \cite{bean2015ajp,bean2016comminstat,eshragh2020plosone,eshragh2015modsim}) and computational complexity (e.g., \cite{eshragh2011mor,avrachenkov2016aor,eshragh2011aor}). 

Recently, by utilising the particular structure of Toeplitz matrices and methods from RandNLA, some superfast algorithms have been developed for approximating Toeplitz linear least squares solutions (e.g., \cite{eshragh2019lsar,shi2019sublinear}). %In this paper, 
We aim to compare the efficacy of these new algorithms on large-scale synthetic as well as real-world data.

\paragraph{Notation.} Vectors and matrices are denoted by bold lower-case and bold upper-case letters respectively (e.g., $\vv$ and $\VV$). Vectors are assumed to be column vectors. We use lower-case letters or Greek letters to denote scalar constants (e.g., $d$, $\epsilon$). Random variables are denoted by upper-case letters (e.g., $Y$).
For a real vector $\vv$, its transpose is $\vv^{\transpose}$. For two vectors $\vv,\ww$, their inner-product is $\dotprod{\vv, \ww}  = \vv^{\transpose} \ww$. For a vector $\vv$ and matrix $\VV$, $\|\vv\|$ and $\|\VV\|$ denote the vector $\ell_{2}$ norm and matrix spectral norm. %The condition number of a matrix $ \AA $, which is the ratio of its largest and smallest singular values, is denoted by $ \kappa(\AA) $. The range of a matrix $ \AA \in \reals^{n \times d}$, denoted by $ \range(\AA) $, is a subspace of $ \reals^{n} $ consisting of all the vectors $ \left\{ \AA \xx \mid \xx \in \reals^{d} \right\}$. 
Adopting \texttt{Matlab} notation, we use $ \AA(i,:) $ to mean the $ i\th $ row of $ \AA $, but we consider it as a column vector. Finally, $ \ee_{i} $ denotes a vector whose $ i\th $ component is one, and zero elsewhere.

%% file: 2OLS_Problems.tex
\section{\bf Ordinary Least Squares Problems}

%Consider the classic regression problem, 

Suppose we have a system of linear equations $\AA \xx = \bb$ such that $\AA \in \R ^{n \times d}$, $\xx \in \R ^d$, $\bb \in \R ^n$, and the system is strongly overdetermined ($n \gg d$), %. In other words, there are more equations than unknowns, 
so that $\AA$ is a tall and thin matrix. In general, $\AA \xx = \bb$ will be infeasible, meaning we may not be able to find an $\xx$ that satisfies the equation. However, in many applications, it is of interest to find an$\xx^{\star}$ that minimizes the difference between $\AA \xx^{\star} $ and $\bb$. The method of Ordinary Least Squares achieves this by minimising the sum of squares of the residual vector $\rr = \bb - \AA\xx$. % who is it attributed to ?)%  
To formalise this, we define Ordinary Least Squares problems as follows.

\newpage
\begin{definition}[Ordinary Least Squares Problem] \label{def:ols}
An Ordinary Least Squares (OLS) problem with inputs $\AA \innd$ and $\bb \inn{d}$ solves the minimisation problem 
\begin{align*} 
	\min_{\xx \inn{d}} \|\AA\xx-\bb\|^2.
\end{align*} 
\end{definition}

The solution to this minimisation problem is well known and shown in \cref{thm:OLSsoln} \cite{eshragh2019lsar}.

\begin{theorem}[Solution to OLS problem] \label{thm:OLSsoln}
    The optimal solution of the OLS minimisation problem in \cref{def:ols} satisfies
    the normal equation
    \begin{align*} 
	%\xx^{\star} = (\AA^{\transpose} \AA)^{-1}  
    \AA^{\transpose} \AA \xx^{\star} = 
	\AA^{\transpose} \bb,
	\end{align*}
	which always has a solution. If $\AA^{\transpose}\AA$ is nonsingular, $\xx^{\star}$ is unique.
	%the OLS solution is given by
	%\begin{align*} 
	%	\xx^{\star} = (\AA^{\transpose} \AA)^{-1}  %\AA^{\transpose} \bb.
	%\end{align*} 
	Otherwise, %that is if $\AA^{\transpose} \AA$ is singular, 
	the unique solution of minimum norm $\|\xx^{\star}\|$
	%one can find a solution using an approach involving the Moore-Penrose pseudo-inverse of $\AA$ and singular value decomposition.
	can be found via the singular value decomposition of~$\AA$.
\end{theorem}

%
%
%\begin{definition}[Condition Number] \label{cono}
%    The condition number of a matrix $\AA$, denoted by $\kappa(\AA)$, is the ratio of its largest and smallest singular values.
%\end{definition}
%If the condition number is very large, we say that the matrix is \emph{ill-conditioned}, and if the condition number is not much larger then 1 it is \emph{well-conditioned}. In broad terms, the condition number represents the rate at which $\xx$ will change with respect to change in $\bb$ when solving equation $\AA \xx = \bb$. If the matrix is ill-conditioned a numerical approximation of $\xx$ may have large error even with small error in $\bb$.
%wiki^
%perhaps one of the most common uses of this is...
%Include its relation to time series data...
%complexity of finding a solution. 

%---------------------------------------------
\subsection{Solving Large OLS Problems via RandNLA} \label{sec:OLSvRand}
%---------------------------------------------

Randomised Numerical Linear Algebra (RandNLA) is a new tool to deal with big data. It utilises random sampling of the elements or rows or columns of a matrix to produce a second smaller (compressed) matrix that is similar to the first matrix in some way, yet computationally easier to deal with \cite{drineas2017lecture}.

In this section we look at the application of RandNLA to OLS regression. We consider again the system of linear equations $\AA \xx = \bb$ with $\AA \in \R ^{n \times d}, \xx \in \R ^d$, $\bb \in \R ^n$ and  $n > d$. In essence, RandNLA methods for OLS problems involve the appropriate choice of matrix $\bSS \in \R^{c \times n}$ to perform some form of sampling (according to a chosen distribution) and/or pre-processing operation. We are able to compress our data matrix $\AA \in \R ^{n \times d}$ into a smaller matrix $\bSS\AA \in \R^{c \times d}$ that will ideally lead to similar results. We replace the OLS problem (\cref{def:ols}) by a compressed least squares approximation problem \cite{woodruff2014sketching}
%need dim of S above%
\begin{align}
	\min_{\xx \in \R ^d} \|(\bSS\AA\xx)-(\bSS\bb)\|^2.  %different to prev Z dropped or something 
	\label{eq:OLSc}
\end{align}

A solution $\xx^{\star}_s$ to the smaller problem \cref{eq:OLSc} can be found using a direct method such as %that in \cref{thm:OLSsoln}.
QR factorization of the matrix $\bSS\AA$,
%Instead of finding an exact solution to \cref{def:ols}, we find 
giving an approximation to $\xx^{\star}$ such that 
\begin{align} \label{eq:rel}
    \|\AA\xx^{\star}_s-\bb\| \leq (1+\varepsilon)\|\AA\xx^{\star}-\bb\|,
\end{align}
%should there be a lower and upper? nah?
where  $\varepsilon>0$.
%and $\xx^{\star}$ is the optimal solution to our full scale problem %\cref{def:ols}, with solution described in \cref{thm:OLSsoln} %\cite{woodruff2014sketching}. 
Since we have a randomised algorithm, there is some probability $\delta <1$ (depending on $\varepsilon>0$) with which the algorithm will fail, i.e., \cref{eq:rel} will not be satisfied. 
%eq 9 ali's paper.

As mentioned, RandNLA methods for OLS problems depend on the choice of $\bSS$. One may argue that the simplest choice for $\bSS$ performs uniform random sampling on the rows of $\AA$\cite{eshragh2019lsar}. Unfortunately, while this can be achieved quite easily and quickly, uniform sampling strategies perform poorly because of nonuniformity in the rows of $\AA$. 

%Routines involving RandNLA for OLS problems
There are two ways to address this problem \cite{woodruff2014sketching}. A data-independent (or data-oblivious) approach, such as \cref{RandLS}, involves some kind of preprocessing (or preconditioning) of matrix $\AA$ that transforms it in order to make it more uniform. Random sampling can then be applied to this uniform, transformed $\AA$. A second data-aware approach (such as using leverage scores) involves weighting the rows of $\AA$ so that rows with more information, in some sense, are randomly sampled with higher probability.

%---------------------------------------------
\subsubsection{Data-oblivious Approach: Sampled Randomised Hadamard Transform}
%---------------------------------------------

Drineas and Mahoney \cite{drineas2017lecture} present a data-oblivious method for OLS %via RandNLA 
called Sampled Randomised Hadamard Transform (SRHT). As discussed, a data-oblivious method overcomes %the problem of 
nonuniformity in the rows of $\AA$ by preprocessing $\AA$ in some way.
The Randomised Hadamard Transform (RHT) $\HH_m\DD$ performs this role. This is the product of $\HH_m \in \R^{n\times n}$ (defined in \cref{def:Hadamard}) and the diagonal matrix $\DD \in \R^{n \times n }$ with $\DD_{ii}$ equal to 1 or $-1$ with probability $\frac{1}{2}$. Using the RHT to \emph{uniformise} $\AA$ has the advantage of being quite fast:  $\bigO{n \log_2 n}$ time to compute a vector $\HH_m \DD \xx$, or $\bigO{n \log_2 c}$ if we only want to access $c$ elements of vector $\HH_m \DD \xx$ (as we do when sampling).

%preprocseseccing spreads it out
%then it is sampled 

\begin{definition}[Hadamard Transform] \label{def:Hadamard}
    For some $m>0$, the Hadamard transform (normalised), denoted $\HH_m \inn{n}{n}$, $n = 2^{m+1}$, is defined recursively with $\HH_0 = 1$ and 
    \[
    \HH_m = \frac{1}{\sqrt 2}
    \begin{pmatrix} 
       \HH_{m-1} & \HH_{m-1}
    \\ \HH_{m-1} & -\HH_{m-1}
    \end{pmatrix}.
    \]
\end{definition}
%the definition exists for n = a power of 2 but can be generalised 
%wiki^
%discuss subspace embeddings?
Following preprocessing, a uniform sampling matrix $\bSS \in \R^{c \times n}$ is applied. This matrix is given in the sampling-and-rescaling form, but it may be implemented implicitly in practice through simply sampling the rows. Thus, when sampling and preprocessing are applied, we arrive at a smaller OLS problem
\begin{align}\label{SRHT}
    \min_{\xx \in \R ^d} \|(\bSS\HH_m\DD\AA\xx)-(\bSS\HH_m\DD\bb)\|^2.
\end{align}
\begin{theorem}[Number of Rows to Sample in SRHT Algorithm \cite{drineas2017lecture}] \label{thm:cstate}
    Let $x_s^*$ be an optimal solution to \cref{SRHT}. If the ideal number of sampled rows is given by
    \begin{align}
        c = \max{\Big( 48^2 d \ln{(40nd)} \ln{\big(100^2d \ln{(40nd) }\big)}, 
        40d \ln{\big(40nd\big) / \varepsilon} \Big)},
        \label{eq:c}
    \end{align}
    then for some $0 < \varepsilon < 1$, 
    %\begin{align*}
        $\Pr(\|\AA\xx^{\star}_s-\bb\| \leq (1+\varepsilon)\|\AA\xx^{\star}-\bb\|) \geq 0.8$.
    %\end{align*}
\end{theorem}
\cref{thm:cstate} ensures that $\xx^{\star}_s$ satisfies \cref{eq:rel} with probability at least 0.8. The SRHT algorithm is described in \cref{RandLS}.

\begin{algorithm}[ht]
\begin{algorithmic}[1]
    \REQUIRE $\AA \in \R ^{n \times d}$, $\bb \in \R ^n$, error parameter $\varepsilon \in (0,1)$
    \STATE Let $c$ be given by \cref{eq:c};
    \STATE Let $\bSS$ be an empty matrix;
    \FOR{$t = 1, \dots, c$ (i.i.d.\ trials with replacement)}
      %  \STATE Select uniformly at random an integer from ${1, 2, \dots, n}$;
      %  \IF{integer $i$ is selected;}
      %      \STATE{Append the column vector $(\sqrt{n/c})\ee_i$ to $\bSS$;}
      %  \ENDIF
        \STATE Select uniformly $i$ at random an integer from ${1, 2, \dots, n}$;
        \STATE{Append the row vector $(\sqrt{n/c})\ee_i^\transpose$ to $\bSS$.}%, where $\ee_i$ is the $i^{th}$ column of an $n\times n$ identity matrix;}
    \ENDFOR
    \STATE Let $\HH_m \in \R^{n \times n}$ be the normalised Hadamard transform matrix;
    \STATE Let $\DD \in \R^{n \times n}$ be a diagonal matrix with
    \[
    	\DD_{ii} = 
    	\begin{cases} 
    	    +1, & \text{with probability} \frac{1}{2}; \\
    	    -1, & \text{with probability} \frac{1}{2}; \\
        \end{cases}
    \]
    \ENSURE $\xx^{\star}_s$, the solution of the OLS problem 
    %with inputs $\big(\bSS\HH_m\DD \AA\big) \xx^{\star}_s$ and $(\bSS\HH_m\DD) \bb$, as given in \cref{def:ols,thm:OLSsoln}.
    $\min_{\xx_s \inn{d}} \|\big(\bSS\HH_m\DD \AA\big) \xx_s-(\bSS\HH_m\DD) \bb\|^2$.
\caption{SRHT Algorithm \cite{drineas2017lecture}}
\label{RandLS}
\end{algorithmic}
\end{algorithm}

%---------------------------------------------
\subsubsection{Data-aware Approach: Leverage Scores-based Random Sampling} \label{sec:LSS}
%---------------------------------------------

As discussed in \cref{sec:OLSvRand}, an alternative to the data-oblivious approach is data-aware approaches, in which information from the data matrix is assessed before sampling to determine which rows are deemed (in some sense) more important and thus more ideal to be selected in the sampling procedure. 
%are sampled with higher probability. In particular, we will examine
%non uniformity of the distribution.
%we will examine?
In particular, \emph{leverage score sampling} is a common way to assess the \emph{importance} of a row. 
%in which rows are sampled with probability proportional to their leverage scores.
%Given matrix $\AA \in \R^{n \times d}$ with $n\geq p$ 
%The $i$\th leverage score corresponding to the $i$\th row of $\AA$ is defined by $\ell (i) := \|\QQ(i,:)\|^2$, where $\QQ$ is any orthogonal matrix such that $\range(\QQ) = \range(\AA)$.
In general terms, a statistical leverage score measures how far the values of an observation are from other observations. \cref{thm:lev} presents a more precise definition of leverage scores. 
%Since Q can be any matrix we find leverage scores as follows
\begin{definition}[Leverage Score]\label{thm:lev}
    Given matrix $\AA \in \R^{n \times d}$ with $n\geq d$ and $\text{rank}(\AA)=d$, the $i$\th leverage score corresponding to the $i$\th row of $\AA$ is given by the $i$\th diagonal entry of 
    %the hat matrix given by
    $\AA(\AA^{\transpose}\AA)^{-1}\AA^{\transpose}$; that is,
    \begin{align}
        \ell(i) &= \ee_i^{\transpose}\AA(\AA^{\transpose}\AA)^{-1}\AA^{\transpose}\ee_i ~~ \text{for} ~~ i = 1, \dots, n.
    \end{align}
\end{definition}
It can be shown that $\ell(i) \geq 0$ for all $i$ and $\sum_{i=1}^m \ell(i) = d$, and so we can construct a probability distribution $\pi$ over the rows of $\AA$ by 
\begin{align}\label{ldist}
    \pi(i) := \frac{\ell(i)}{d} ~~\text{for}~~ i = 1, \dots, m.
\end{align}

Sampling according to leverage scores thus involves randomly selecting and rescaling rows of $\AA$ proportional to their leverage scores. In terms of our sampling and rescaling formalism \cref{eq:OLSc}, $\bSS \in \R^{c \times n}$ is constructed by randomly choosing each row (with replacement) from the $n \times n$ identity matrix according to the nonuniform distribution \cref{ldist}. If row $i$ is selected, it is rescaled by multiplying by $1/\sqrt{c \pi (i)}$.
A Leverage-Score-Based Random Sampling algorithm is given in \cref{LSRSA}.

\begin{algorithm}[h!]
\begin{algorithmic}[1]
    \REQUIRE $\AA \in \R ^{n \times d}$, $\bb \in \R ^n$, error parameter $\varepsilon \in (0,1)$;
	\STATE Compute leverage scores $\ell(i)$ for $i = 1, \dots, n$ as in \cref{thm:lev};
	\STATE Compute the sampling distribution $\pi(i)$ for $i = 1, \dots, n$ as in \cref{ldist};
	\STATE Set $c$ as in \cref{LSARc};
	\STATE Form $\bSS \in \R^{c \times n}$ by randomly choosing $c$ rows of the corresponding identity matrix according to the probability distribution $\pi $ with replacement and rescaling factor $\frac{1}{\sqrt{c\pi(i)}}$;
	\STATE Construct the sampled data matrix $\widehat \AA = \bSS \AA$ and response vector $\widehat \bb = \bSS \bb$;
	\STATE Solve the associated compressed OLS problems as in \cref{eq:OLSc} using a conventional method;
    \ENSURE $\xx^{\star}_s$, the solution of $\min_{\xx \in \R ^d} \|\widehat\AA \xx - \widehat\bb\|$.
\caption{Leverage Score Based Random Sampling}
\label{LSRSA}
\end{algorithmic}
\end{algorithm}
% In this manner, similar to our compressed problem \cref{eq:OLSc}, the \texttt{LSAR} algorithm replaces a Toeplitz OLS problem into a compressed least squares approximation problem
% \begin{align}
% 	\min_{x\in\R^d} \|(\bSS\XX_{n,p}\mathbf{\phi})-(\bSS\yy)\|^2.  %different to prev Z dropped or something
% \end{align}
Computing leverage scores as in \cref{thm:lev} %O(nd^2) flops alis paper pg 9 
is more costly than solving the original OLS problem. However, as we see in the coming section, one can find approximate leverage scores cheaply. The \texttt{LSAR} algorithm \cite{eshragh2019lsar} utilises \emph{approximate leverage scores}, and the \texttt{REPEATEDHALVING} \cite{shi2019sublinear} algorithm utilises \emph{generalised leverage scores} with respect to a smaller approximate matrix. The downside of approximate leverage scores is that they mis-estimate the true leverage scores by some factor $0 < \beta \leq 1$, that is $\widehat \ell (i) \geq \beta \ell (i), \text{     for } i = 1, \dots, m$. 
This leads to a trade-off between speed and accuracy.

% Template for Algorithms

%\begin{algorithm}[h!]
%\begin{algorithmic}[1]
%\REQUIRE $n, x_1,x_2,\cdots,x_n$
%\ENSURE $Norm$
%\STATE Set $sum=0$.
%\FOR {$i=1,2,\cdots,n$} 
%\STATE  $ sum = sum + x_i^2$
%\ENDFOR
%\STATE $Norm = sum^{1/2}$
%\caption{Compute the Euclidean norm of a vector ${\bf x}$}
%\label{algintro}
%\end{algorithmic}
%\end{algorithm}

%% file: 3Toeplitz_OLS_Problems.tex
\section{\bf Toeplitz OLS Problems for Time-series Data}

%\section{Introduction}
\label{sec:TOLS:intro}
A Toeplitz Ordinary Least Squares (TOLS) problem is an OLS problem 	
\begin{align}\label{TOLS}
\min_{x \in \R^d} \zz (\xx) = \|\bm{T} \xx-\bb\|^2
\end{align}
in which $\bm{T}$ is a Toeplitz matrix as given in \cref{def:Toeplitz}.
As we can see, there are only $n+p-1$ distinct numbers. This is useful for computation and storage of Toeplitz matrices.

\paragraph{Motivation.} Toeplitz matrices arise in a wide range of problems in both pure mathematics (such as algebra, combinatorics,  differential geometry, etc.)\ and applied mathematics (approximation theory, image processing, time series analysis, etc.)\  \cite{ye2016Toeplitz}. In particular, fitting an \texttt{AR}$(p)$ model (see \cref{def:AR}) to time series data requires solving a TOLS problem for multiple possible orders of $p$. Given the present ubiquity of data, one is often required to fit an \texttt{AR} model to very large time series data sets, referred to as \emph{big time-series data}. This TOLS problem quickly becomes a computational bottleneck because of the need to %calculate
%%or should say A-1A?
%$(\TT^{\transpose} \TT)^{-1}  \TT^{\transpose} \bb$
solve \cref{TOLS} repeatedly
for different values of $p$. However, by utilising the unique structure of Toeplitz matrices and methods from RandNLA, some superfast (i.e., faster than $O(np)$) algorithms have been developed recently for approximating solutions to TOLS problems. In the following sections we examine some of these algorithms.

\begin{definition}[Toeplitz Matrix] \label{def:Toeplitz}
A Toeplitz matrix has the form
\begin{align*} 
    \bm{T} & = \begin{pmatrix}
        a_p & a_{p-1} & \dots & a_{1}\\
        a_{p+1} & a_p & \ddots & a_{2}\\
        \vdots & \ddots & \ddots & \vdots\\
        a_{2p-1} & a_{2p-2} & \dots & a_{p}\\
        \vdots & \vdots & \vdots & \vdots\\
        a_{n+p-1} & a_{n+p-2} & \dots & a_{n}\\
    \end{pmatrix},
\end{align*}
where $a_i \in \R$ for all $i$. It has constant descending diagonals.
\end{definition}

A time series can be defined as a collection of random variables $\{Y_t;~~t=0, \pm 1, \pm 2, \dots\}$ indexed according to time $t$. A time series is \emph{stationary} (weakly stationary) if it has a constant mean $\mu$ that does not depend on $t$, and the autocovariance function $\text{Cov}(Y_t, T_{t+h})$ depends only on lag $h = |s-t|$, the difference between two time points. 

If the current value $Y_t$ can be explained with a function of $p$ past values, we can model the time series with an $\texttt{AR}(p)$ model.

%\pagebreak
\begin{definition}[Autoregressive Model of Order $p$]\label{def:AR}
    An autoregressive model of order $p$, denoted \texttt{AR}$(p)$, has the form
    \begin{align*}
        Y_t = \phi_1 Y_{t-1} + \phi_2 Y_{t-2} +\dots + \phi_p Y_{t-p} + \W_t,
    \end{align*}
    where $Y_t$ is a stationary time series with mean zero, $\phi_1, \phi_2, \dots, \phi_p$ are the regression parameters with $\phi_p \neq 0$, and $\W_t$ is a Gaussian white noise series, i.e., each $\W_t$ is an independent and identically distributed normal random with mean 0 and variance $\sigma_{\W}^2$. 
\end{definition}

Given a set of time series observations $y_1, \dots, y_n$, if we wish to fit an $\texttt{AR}(p)$ model, we need to find auto-regression parameters $\phi_1, \phi_2, \dots, \phi_p$. Finding these parameters exactly by the method of \emph{maximum likelihood estimates} (i.e., by maximising the log-likelihood function) can be shown to be intractable and must be solved numerically \cite{Hamilton1989}. However, finding the log-likelihood function of the parameters conditional (CMLE) on the first $p$ observations is an alternative approach for large samples, equivalent to obtaining the parameters from an OLS problem that regresses $y_t$ on $p$ of its own lagged values \cite{Hamilton1989}.

Parameters $\phi_1, \phi_2, \dots, \phi_p$ in the $\texttt{AR}(p)$ model $Y_t = \phi_1 Y_{t-1} + \phi_2 Y_{t-2} +\dots + \phi_p Y_{t-p} + \W_t$ are found by solving the OLS problem
\begin{subequations}\label{xphi}
\begin{align} \label{eq:lsprob}
    \begin{pmatrix}
        y_{p+1}\\ y_{p+2}\\ \vdots\\ y_{n}
    \end{pmatrix} &\approx \begin{pmatrix}
        y_p & y_{p-1} & \dots & y_{1}\\
        y_{p+1} & y_p & \dots & y_{2}\\
        \vdots & \vdots & \dots & \vdots\\
        y_{n-1} & y_{n-2} & \dots & y_{n-p}\\
    \end{pmatrix} %\times
    \begin{pmatrix}
        \phi_{1}\\ \phi_{2}\\ \vdots\\\phi_p
    \end{pmatrix}, \\
    \yy_{n-p,1} &\approx \XX_{n-p,p}\bm{\phi}_{p,1},
\end{align}
\end{subequations}
where $\XX_{n,p}$ is a Toeplitz matrix, often referred to as the data matrix. Thus we arrive at a TOLS problem,
%When fitting an $\texttt{AR}(p)$ auto-regression model,
where the order $p$ is an unknown parameter that must be estimated, typically using the partial autocorrelation function (PACF).
\begin{definition}[Partial Autocorrelation Function]
    \label{PACF}
    The PACF of a stationary time series $\{Y_t; ~t = 0, \pm 1, \pm 2, ...\}$ at lag h is defined by
    \begin{align}
    \begin{cases}
        \rho(Y_t, Y_{t+1}) & \text{for } h = 1, \\
        \rho(Y_t- \widehat{Y}_{t}, Y_t-\widehat Y_{t,h}) & \text{for } h \geq 2,
    \end{cases}
    \end{align}
    where $\rho$ denotes the \emph{correlation function} and $\widehat Y_t$ and $\widehat Y_{t+h}$ are the linear regression of $Y_t$ and $Y_{t+h}$ on $\{X_{t+1}, \dots ,  X_{t+h-1}\}$.
\end{definition}

Order $p$ is estimated by selecting the last lag at which the PACF is nonzero. In the algorithm, $\tau_h$ denotes the PACF value estimated at lag $h$ using the CMLE of the parameters $\phi_1, \phi_2, \dots, \phi_p$ (i.e., through solving the associated TOLS problem). Also, $\widehat\tau_h$ denotes the PACF value using the CMLE of the parameters when based on the compressed (sampled and re-scaled) OLS problem.

\begin{remark}
Estimating the order $p$ requires solving the TOLS regression problem \cref{TOLS} multiple times with different $p$. In the context of big time-series data, these TOLS problems are the computational bottleneck. 
\end{remark}

%-----------------------------------
\subsection{\texttt{LSAR} Algorithm}
%-----------------------------------

Eshragh et al.\ \cite{eshragh2019lsar} develop a fast algorithm called \texttt{LSAR} for estimating the leverage scores of an autoregressive model in big time-series data. To mitigate the computational complexity of solving numerous OLS problems, \texttt{LSAR} utilises a data-aware RandNLA sampling routine based on leverage scores.
%in order to speed up the OLS problem solving computation.
In this section we discuss how LSAR relates to Toeplitz OLS problems and then how to perform the algorithm.

%%
%Overcoming this is the core motivation of the \texttt{LSAR} algorithm.

% In, general terms, a \emph{statistical leverage score} measures how far the values of an observation are from other observations. Sampling according to leverage scores, thus, involves randomly selecting and re-scaling rows of a matrix $\XX_{n,p}$ proportional to their leverage scores. In this manner, similar to \cref{eq:OLSc}, the \texttt{LSAR} algorithm replaces a Toeplitz OLS problem into a compressed least squares approximation problem
% \begin{align}
% 	\min_x \|(\bSS\XX_{n,p}\mathbf{\phi})-(\bSS\yy)\|^2.  %different to prev Z dropped or something 
% \end{align}

As stated in \cref{sec:LSS}, calculating leverage scores exactly could be computationally costly. However, Eshragh et al.\ \cite{eshragh2019lsar} developed an efficient approximation to estimate the leverage scores recursively, as presented in \cref{alev}.

\begin{subequations}
\begin{definition}[Approximate Leverage Scores] \label{alev}
    For an \texttt{AR}(p) model with $p \geq 1$, the fully-approximate leverage scores are given by the recursion
    \begin{align}
    \widehat{\ell}_{n,p}(i):= \label{eq:FALS}
        \begin{cases} 
          {\ell}_{n,1}(i) & \text{for } p=1, \\
          \tilde{\ell}_{n,2}(i), & \text{for } p=2, \\ \widehat{\ell}_{n-1,p-1}(i)+\frac{(\widehat\rr_{n-1,p-1}(i))^2}{\|\widehat\rr_{n-1,p-1}\|^2}, & \text{for } p\geq3, 
       \end{cases}
    \end{align}
    where
    \begin{align}\label{eq:FAR}
        \widehat\rr_{n-1,p-1}&:=\XX_{n-1,p-1}\widehat{\bm{\phi}} _{n-1,p-1}-\yy_{n-1,p-1},
    \end{align}
    and $\widehat{\bm{\phi}}_{n-1,p-1}$ is the solution of the OLS problem with inputs $\widehat\XX_{n-1,p-1}\in \R^{c\times (p-1)}$ and $\widehat\yy_{n-1,p-1} \in \R^{c}$. Here, $\widehat\XX_{n-1,p-1}$ and $\widehat\yy_{n-1,p-1}$ are compressed data from \cref{eq:lsprob} sampled according to the leverage score distribution
    \begin{align}\label{eq:FASD}
        \widehat \pi_{n-1,p-1}(i) = \frac{\widehat \ell _{n-1,p-1}(i)}{p-1} \text{ \ for } i = 1,\dots,n-p.
    \end{align}
\end{definition}
\end{subequations}

%Here the approximate leverage scores are found recursively and 
The first and second cases are given respectively by the \emph{exact leverage score},
\begin{align*}
    \ell_{n,1}(i) = \frac{y_i^2}{\sum_{t=1}^{n-1}y_t^2}, \text{     for } i = 1, \dots, n-1,
\end{align*}
and the \emph{quasi-approximate leverage scores}, 
\begin{align*}
    \tilde{\ell}_{n,2}(i) = \ell_{n-1,1}(i)+\frac{(\tilde\rr_{n-1,1}(i))^2}{\|\tilde\rr_{n-1,1}\|^2}, \text{     for } i = 1, \dots, n-1,
\end{align*}
and the remainder are evaluated recursively.
Note that $\tilde\rr_{n,p}:= \XX_{n,p} \bm{\tilde\phi}_{n,p} - \yy_{n,p}$, where $\bm{\tilde\phi}_{n,p}$ is the vector of OLS parameters computed from the sampled and re-scaled problem.

It can be shown that the approximate leverage scores misestimate the true leverage scores by some factor $0 < \beta \leq 1$, that is, $\widehat \ell (i) \geq \beta \ell (i), \text{     for } i = 1, \dots, m$. Hence, the choice of $c$ (the number of rows to sample from $\XX_{n-1,p-1}$) is given by 
\begin{align} \label{LSARc}
    c \in \bigO{p \log (p/\delta)/(\beta\varepsilon^2)},
\end{align}
where $\beta$ can be shown to be $1-\bigO{p\sqrt{\varepsilon} }$, and $\varepsilon$ and $\delta$ are given by \cref{eq:rel}.

The \texttt{LSAR} algorithm is given in \cref{RandLSA}.

\begin{algorithm}[t] %[H]
\begin{algorithmic}[1]
 	\REQUIRE Time series data $\{ y_1, \ldots, y_n\}$ and large enough $\bar p \ll n$.
    \STATE Set $h=0$ and $m=n - \bar p$;
    \WHILE {$p<\bar p$}
		\STATE $p \leftarrow p+1$ and $m \leftarrow m+1$;
		\STATE Estimate PACF at lag $p$, i.e., $\widehat \tau_p $;
		\STATE Compute the approximate leverage scores $\widehat \ell_{m,p}(i)$ for $i = 1, \dots, m- p$ as in \cref{eq:FALS};
		\STATE Compute the sampling distribution $\widehat \pi_{m,p}(i)$ for $i = 1, \dots, m- p$ as in \cref{eq:FASD};
		\STATE Set $c$ as in \cref{LSARc};
		\STATE Form $\bSS \in \R^{c \times m}$ by randomly choosing $c$ rows of the corresponding identity matrix according to the probability distribution $\widehat \pi $ with replacement and rescaling factor
		$1/\sqrt{c\pi_i}$;
		\STATE Construct the sampled data matrix $\widehat \XX_{m,p} = \bSS \XX_{m,p}$ and response vector $\widehat \yy_{m,p} = \bSS \yy_{m,p}$;
		\STATE Solve the associated compressed OLS problems to estimate parameters $\widehat{\bm{\phi}}_{m,p}$ and residuals $\widehat \rr_{m,p}$ as in \cref{alev};
    \ENDWHILE
	\STATE Estimate $p^*$ as the largest $p$ such that $|\widehat \tau| \geq 1.96/\sqrt c$.
	\ENSURE	Estimate of $p^*$ and parameters $\widehat{\bm{\phi}}_{n-\bar{p}+p^*,p^*}$.
\caption{\texttt{LSAR} Algorithm \cite{eshragh2019lsar}}
\label{RandLSA}
\end{algorithmic}
\end{algorithm}

%I dont want all the numbers
%Too copied?: 
%eqn 15c could provide more detail.

%-----------------------------------
\subsection{Repeated Halving Algorithm}
%-----------------------------------

The Repeated Halving (RH) algorithm was first given in \cite{cohen2015uniform}. It is a data-aware, sampling-based procedure that returns $\tilde \CC \in \R^{c\times (d+1)}$, a spectral approximation of $\XX = [\TT,\bb]$ where $\TT$ is a Toeplitz matrix. A spectral approximation preserves the magnitude of matrix-vector multiplication, and also preserves the singular values of the matrix \cite{cohen2015uniform}.

\begin{definition}[$\lambda$-Spectral Approximation]
For any $\lambda \geq 1$, $\tilde \AA \in \R^{n' \times d}$ is a $\lambda$-spectral approximation of $\AA \in \R^{n \times d}$ if, for all $\xx \in \R^d$,
\begin{subequations}
\begin{align}
\frac{1}{\lambda} \|\AA \xx \|^2 \leq \|\tilde \AA \xx \|^2 \leq \|\AA \xx \|^2, \\
\frac{1}{\lambda} \xx^{\top} \AA^{\top} \AA \xx 
\leq \xx^{\top} \tilde\AA^{\top} \tilde\AA \xx 
\leq \xx^{\top} \AA^{\top} \AA \xx.
\end{align}
\end{subequations}
\end{definition}

The RH algorithm recursively computes a spectral approximation $\tilde \CC'$ of $\CC'$, using the steps outlined in \cite{cohen2015uniform} and shown in \cref{RHA}. The algorithm utilises generalised leverage scores with respect to a spectral approximation as a way mitigate the cost of calculating leverage scores in full.

\begin{definition}[Generalised Leverage Score \cite{cohen2015uniform}]
    Let $\CC$ and $\BB$ be two matrices with the same number of columns, where $\BB$ has full column rank. The $i$\th generalised leverage score corresponding to the $i$\th row of $\CC$ with respect to $\BB$ is defined to be 
    \[\ell^B(i)=\CC(i,:)^\top (\BB^\top \BB)^{-1}\CC(i,:) = \|\BB(\BB^\top \BB)^{-1}\CC(i,:) \|^2.\]
Furthermore, the approximate generalised leverage score is given by
        \begin{align} \label{eq:glev}
        \tilde \ell^B(i) = \|\GG\BB(\BB^{\top}\BB)^{-1}\CC(i,:)\|^2,
    \end{align}
where $\GG$ is a random Gaussian matrix with $\bigO{\log n}$ rows.
\end{definition}

Note that the $i\th$ generalised leverage score of $\AA$ with respect to $\AA$ is just the leverage score as defined in \cref{thm:lev}.

Uniform sampling to approximate a matrix leads to approximate leverage scores that are good enough for sampling \cite{cohen2015uniform}. Following recursive computation of the spectral approximation, a standard leverage score sampling procedure using approximate generalised leverage scores $\{\tilde \ell^B(i)\}$ for $i = 1, \dots, n$ in \cref{eq:glev} samples $c = \bigO{(d \log d )/\varepsilon^2}$ rows of $\CC$ with probability proportional to its leverage score to form $\tilde \CC$. We utilise the \emph{sampling and rescaling} formulation of the sampling procedure to form $ \tilde \CC = \bSS \CC $, where the $t$\th row of $\bSS \inn{c}{n}$ is $ \ee_i^\transpose/\sqrt{p_{i(t)}}$ if the $i$\th row of $\CC$ is sampled in the $t$\th trial. Rows of $\CC$ are sampled with probability 
\[p_{i(t)} = \frac{\tilde \ell^B(t)}{\sum_{k=1,\dots,n}\tilde \ell^B(k)}.\]

% Using the \emph{sampling and rescaling} formulation of the sampling procedure $ \tilde \CC = \bSS \CC $, where the $i$\th row $\bSS$ has a $ 1/\sqrt{p_{j(i)}}$ in the $j$\th element where $j$ is the row of $\CC$ sampled in the $i$\th trial, with sampling probability 
% \[p_{j(i)} = \frac{\tilde \ell^B(i)}{\sum_{k=1,\dots,n}\tilde \ell^B(k)}.
% \text{ The preceding par is hard to parse.}
% \]

\begin{theorem}\label{thm:cohen1}
\emph{\textbf{(Leverage Score Approximation via Uniform Sampling).}} For any m, let $\AA_u$ be obtained by selecting $\bigO{m}$ rows uniformly at random from $\AA$. Let, ${\ell^{\AA_u}(i)}$ be a set of generalised leverage scores for $\AA$ w.r.t.\ $\AA_u$. Then
\[
\forall{i}, ~~ \ell^{\AA_u}(i) \geq \ell(i),
\]
where  $\ell(i)$ are the true leverage scores given in \cref{thm:lev}, and
\(
\Ex \Big[ \sum_{i=1}^n \ell^{\AA_u}(i) \Big] \leq \frac{nd}{m}.
\)
\end{theorem}

The validity of \cref{RHA} follows from \cref{thm:cohen1}. If we set $m=n/2$ we achieve a uniformly sampled matrix $\AA_u$ with $n/2$ rows, which if used to calculate approximate leverage scores $\ell^{\AA_u}(i)$ will lead to a spectral approximation $\tilde \AA$ with $\bigO{d \log d}$ rows. As $\AA_u$ may still be quite large, in the same manner we can recursively sample $n/2$ rows of $\AA_u$ to produce spectral approximations in an iterative manner. The RH algorithm is given in \cref{RHA}. 

%This paper by shi and woodruff mainly just reports on a pre existing algorithm by cohen et al. but shi and woodruss explain how it can be used in an autoregression context
%%Sampling
%making this more like the original sampling and re-scaling or the original more like this.
%%speeding up 

\begin{algorithm}[!ht]
\begin{algorithmic}[1]
 	\REQUIRE $\XX = [\TT,\bb] \in \R^{n\times (d+1)}$, where $\TT$ is a Toeplitz matrix; %is the [] notation a thing?
 	\STATE Uniformly sample $n/2$ rows of $\XX$ to form $\XX_1$;
 	\STATE Set $i = 1$;
 	\WHILE{$\XX_i$ has greater than $\bigO{d \log d}$ rows}
 	    \STATE Set $i = i+1$;
 	    \STATE Uniformly sample $n/2$ rows of $\XX_{i-1}$ to form $\XX_i$;
    \ENDWHILE
    \WHILE{$i \geq 1$}
        \STATE Set $i = i-1$;
        \STATE Approximate generalised leverage scores of $\XX_i$ w.r.t.\ $\tilde\XX_{i+1} $ by replacing $\BB$ with $\tilde\XX_{i+1}$ in \cref{eq:glev};
        \STATE Use these estimates to sample rows of $\XX_i$ to form $\tilde \XX_i$;
    \ENDWHILE
  	\ENSURE $\tilde \XX \in \R^{c \times (d+1)}$, a spectral approximation consisting of $c = \bigO{d \log d}$ re-scaled rows of $\XX$.
\caption{Repeated Halving Algorithm \cite{cohen2015uniform}}
\label{RHA}
\end{algorithmic}
\end{algorithm}

% \STATE Uniformly sample $n$/2 rows of $\CC$ to form $\CC'$;
% \STATE If $\CC'$ has more that $\bigO{(d \log d)/\varepsilon^2}$ rows, recursively compute a spectral approximation $\tilde\CC'$ of $\CC'$;
% \STATE Approximate generalised leverage scores of $\CC$ w.r.t. $\tilde \CC '$ as in \cref{eq:glev};
% \STATE Use these estimates to sample rows of $\CC$ to form $\tilde \CC$;

\begin{theorem} [Time Complexity of RH Algorithm \cite{shi2019sublinear}]
Given $\TT \in \R^{n\times d}$, $\bb \in \R^n$, accuracy $0<\varepsilon<1$, and probability of failure $0<\delta <1$, $x_s^*$ satisfying \cref{eq:rel} with probability of at least $1-\delta$ can be found in total time
\(
    \bigO{(n \log^2 (n) + \mathtt{poly}(d \log (n/\varepsilon))) \log(1/\delta)},
\)
where $\mathtt{poly}$ is a polynomial function.
\end{theorem}
Note that to fit an $\mathtt{AR}(p)$ model, we need to solve a TOLS problem repeatedly $\bigO {p}$ times. Using the \texttt{RH} algorithm \cref{RHA} to fit an $\mathtt{AR}(p)$ model is considered by \cite{shi2019sublinear} and is achieved by first running \cref{RHA} on the data matrix $\XX_{n-\bar p, \bar p}$ (as in \cref{xphi}) to obtain leverage scores to form a spectral approximation of $\XX$. Then we run the \texttt{LSAR} algorithm (\cref{RandLSA}) except we replace the leverage scores obtained in step 5 with leverage scores obtained by \cref{RHA}.

%% file: 4numerical_results.tex
\section{\bf Numerical Results}

In this section, we implement the \texttt{LSAR} and \texttt{RH} algorithms (often referred to as compressed algorithms) on some time series data, both synthetically generated and real, to investigate the quality and run time of the algorithms. Calculations utilising the full data matrix (referred to as the exact computation or algorithm) are also used to compare run time and error. The algorithms are implemented in MATLAB R2020b on a 64-bit Windows operating system with a 1.8GHz processor and 16GB of RAM. All numerical experiments are performed with double precision. Code is included to measure the computation time (using the \texttt{tic} and \texttt{toc} MATLAB functions) and accuracy. 
%computer
% matlab version
% MATLAB Version: 9.9.0.1538559 (R2020b) Update 3
% 64-bit operating system: Microsoft Windows 10 Home Version 10.0 (Build 19042)
% 16 gb installed ram 
% Intel core i7 10th gen, @1.8GHZ, 4 cores.

All numerical results show the potential of both algorithms, which use compressed data, to provide comparable accuracy and utility to that of the existing alternative using the entire data set. Further, by using compressed data matrices, the algorithms are able to produce these results in considerably less time then the alternative, exact method.

The numerical results are discussed in three subsections where we report the computation time of each algorithm and the accuracy of the estimated parameters,  and compare the PACF generated by each algorithm. In \cref{sec:NRwithout} we present numerical analysis on synthetic data generated \textit{without} outliers from a range of sizes of \texttt{AR}$(p)$ time series models. In \cref{sec:NRwith} we present numerical analysis on synthetic data generated \textit{with} outliers. Finally, in \cref{sec:NRreal} we examine the performance of these algorithms on a real data set.
%each sentence is the same bit boring See lsar
%little mention of the real data set

%-----------------------------------
\subsection{Synthetic Data Without Outliers}
\label{sec:NRwithout}
%-----------------------------------

Two million realisations from six \texttt{AR}$(p)$ time series models were randomly generated for \[
p = \{5, 10, 20, 50, 100, 150\}.\] For each \texttt{AR}$(p)$ model, coefficients corresponding to a stationary time series model for each order were obtained randomly. Synthetic data was generated with a zero constant and variance of one, using the \texttt{simulate} function of MATLAB's econometrics toolbox. When fitting an \texttt{AR} model we run the algorithms over a number of lags up to some maximum value $\bar p$, which we choose to be large enough to detect order $p$. For the synthetically generated data sets we choose $\bar p = \{50, 50, 50, 100, 200, 250\}$ to correspond respectively to the \texttt{AR}$(p)$ models with $p = \{5, 10, 20, 50, 100, 150\}$.

%-----------------------------------
\subsubsection{Computational Time of the PACF} 
\label{sec:Time}
%-----------------------------------

We compare the time it takes for each algorithm to find the PACF (\cref{PACF}) for a range of lag values, $h$.
%discuss max pbar

For each lag, finding the PACF involves solving a TOLS problem. To compare the computational time of the algorithms, the associated TOLS problem is solved
at each lag using a compressed data matrix based the approximate leverage scores of the \texttt{LSAR} algorithm, using a compressed data matrix based on leverage scores generated by the \texttt{RH} algorithm and using the entire data matrix to calculate solve the TOLS problem exactly. We choose $c=2,000$ as the number of sampled rows for each of the compressed algorithms.

\begin{figure}[p]  % Fig 1
	\centering
	\begin{subfigure}{.49\textwidth}
		\includegraphics[width=\textwidth]{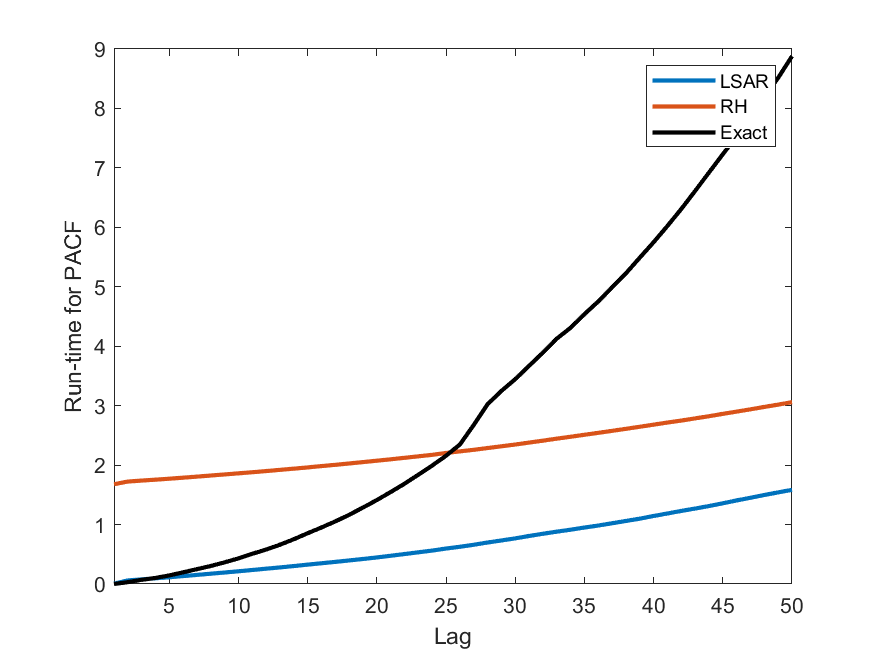}
		\caption{$\mathtt{AR(5)}$ }
		\label{fig:AR5_Time}
	\end{subfigure}
	\begin{subfigure}{.49\textwidth}
		\includegraphics[width=\textwidth]{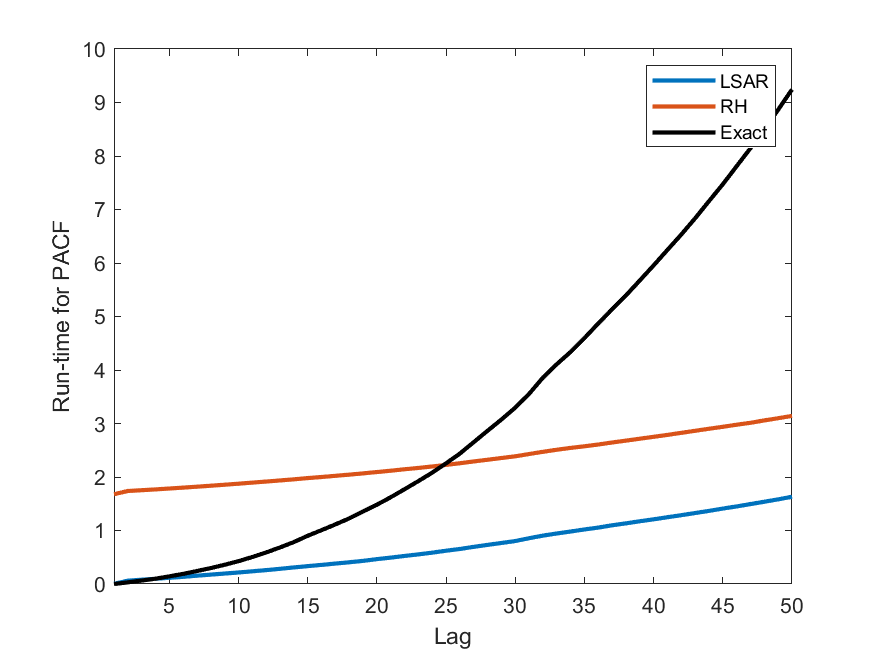}
		\caption{$\mathtt{AR(10)}$ }
		\label{fig:AR10_Time}
	\end{subfigure}
    \begin{subfigure}{.49\textwidth}
		\includegraphics[width=\textwidth]{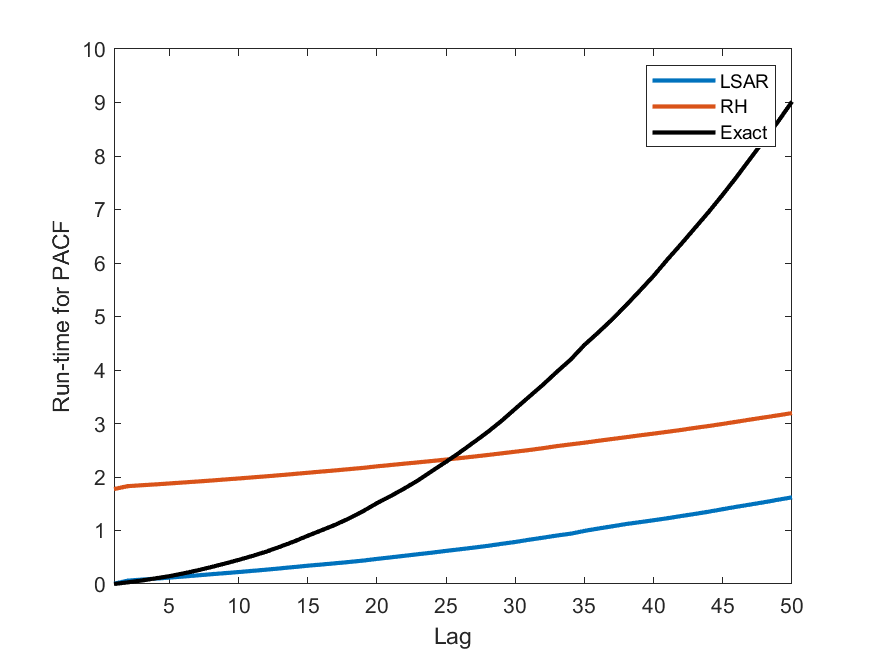}
		\caption{$\mathtt{AR(20)}$ }
		\label{fig:AR20_Time}
	\end{subfigure}
	\begin{subfigure}{.49\textwidth}
		\includegraphics[width=\textwidth]{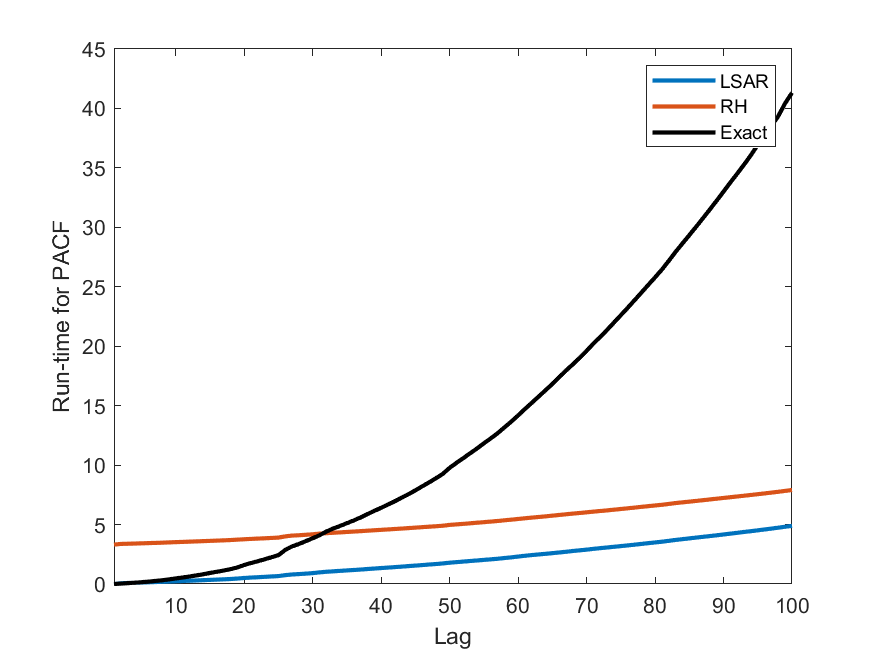}
		\caption{$\mathtt{AR(50)}$ }
		\label{fig:AR50_Time}
	\end{subfigure}
	\begin{subfigure}{.49\textwidth}
		\includegraphics[width=\textwidth]{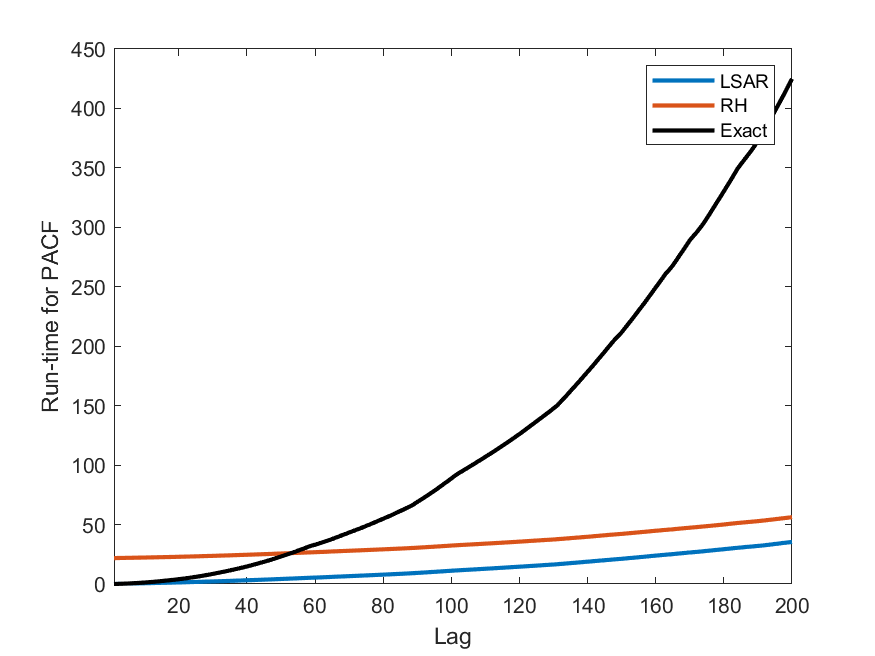}
		\caption{$\mathtt{AR(100)}$ }
		\label{fig:AR100_Time}
	\end{subfigure}
	\begin{subfigure}{.49\textwidth}
		\includegraphics[width=\textwidth]{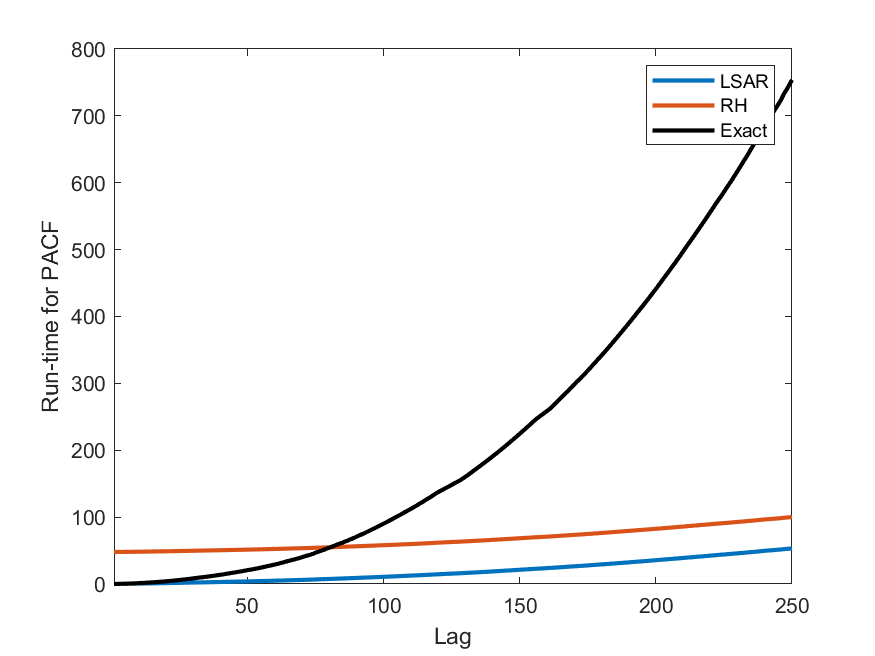}
		\caption{$\mathtt{AR(150)}$ }
		\label{fig:AR150_Time}
	\end{subfigure}
	\caption[Computation Time to Compute the PACF of \texttt{LSAR}, \texttt{RH} and Exact Algorithms on Synthetic Data Without Outliers]{Figures (a) to (f) corresponding to the labeled \texttt{AR}$(p)$ models, show the comparison between the computation time (in seconds) to generate the PACF for the \texttt{LSAR} algorithm (in blue), the Repeated Halving algorithm (in red) and the exact computation of PACF (in black).}
	\label{fig:Time}
\end{figure}

\cref{fig:Time} compares the run time to compute the PACF by the \texttt{LSAR} algorithm, the \texttt{RH} algorithm and the exact calculation, for each \texttt{AR}$(p)$ model. Time is plotted cumulatively over each lag. We can clearly verify the speed of both compressed algorithms when compared to exact computation. In particular, the difference in computation time is exemplified by \cref{fig:AR150_Time}, which presents a 700-second difference between the exact method and the two compressd methods.

For the \texttt{RH}  %\cref{RHA} and 
and \texttt{LSAR} % \cref{RandLSA} we note that
algorithms, we see that they have similar computation times, separated by a constant that is due to the \texttt{RH} algorithm computing leverage scores prior to the algorithms' iteration over the lag.
%
% 3rd order vs log order add a reference line or discuss shape
%

\begin{figure}[p]  % Fig 2
	\centering
	\begin{subfigure}{.49\textwidth}
		\includegraphics[width=\textwidth]{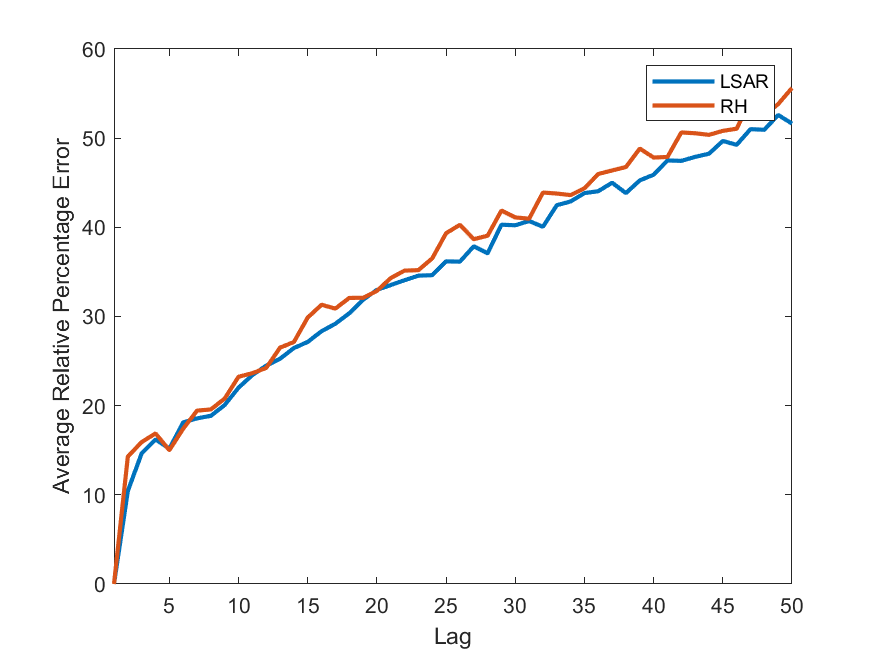}
		\caption{$\mathtt{AR(5)}$ }
		\label{fig:AR5_Error}
	\end{subfigure}
	\begin{subfigure}{.49\textwidth}
		\includegraphics[width=\textwidth]{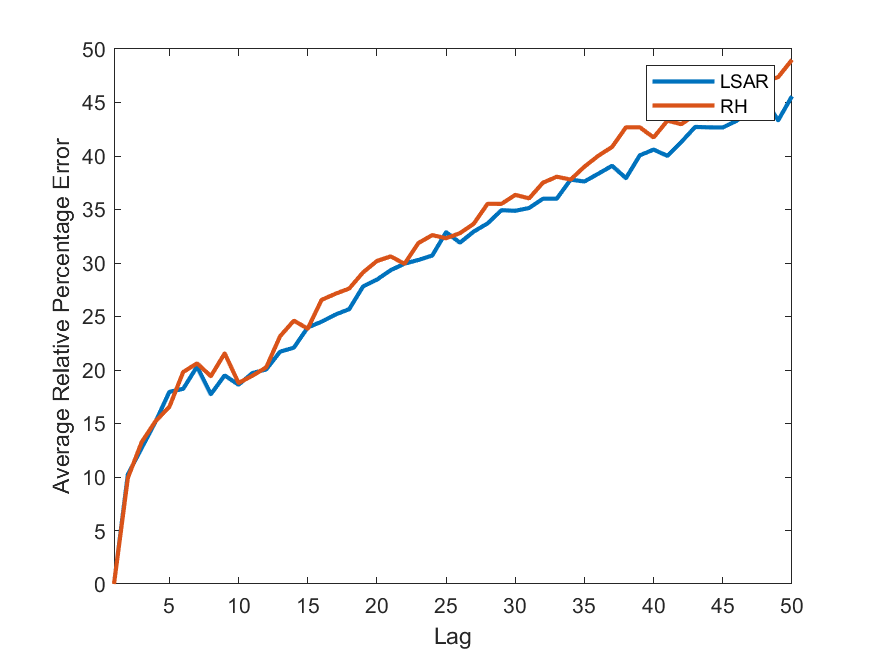}
		\caption{$\mathtt{AR(10)}$ }
		\label{fig:AR10_Error}
	\end{subfigure}
	\begin{subfigure}{.49\textwidth}
		\includegraphics[width=\textwidth]{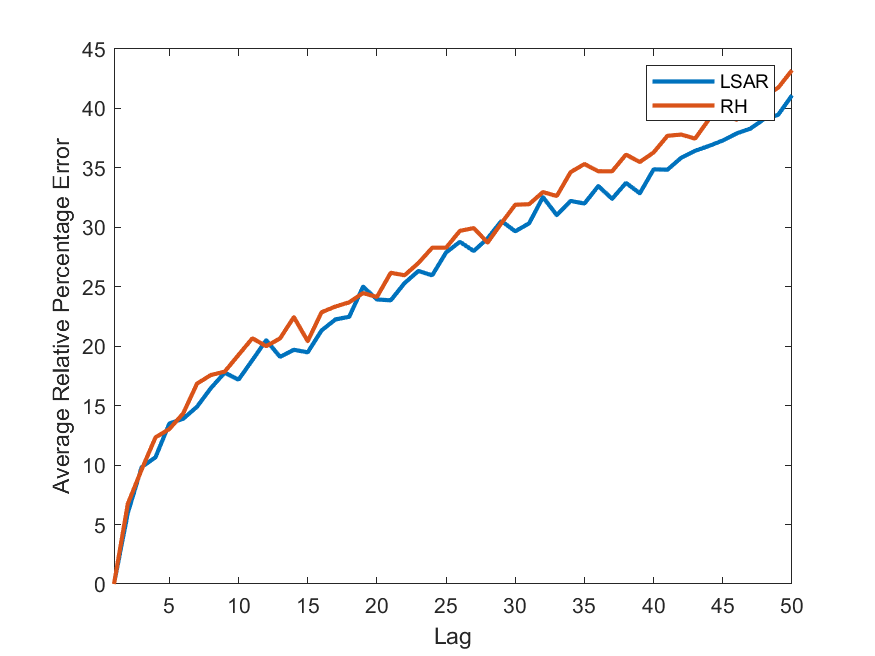}
		\caption{$\mathtt{AR(20)}$ }
		\label{fig:AR20_Error}
	\end{subfigure}
	\begin{subfigure}{.49\textwidth}
		\includegraphics[width=\textwidth]{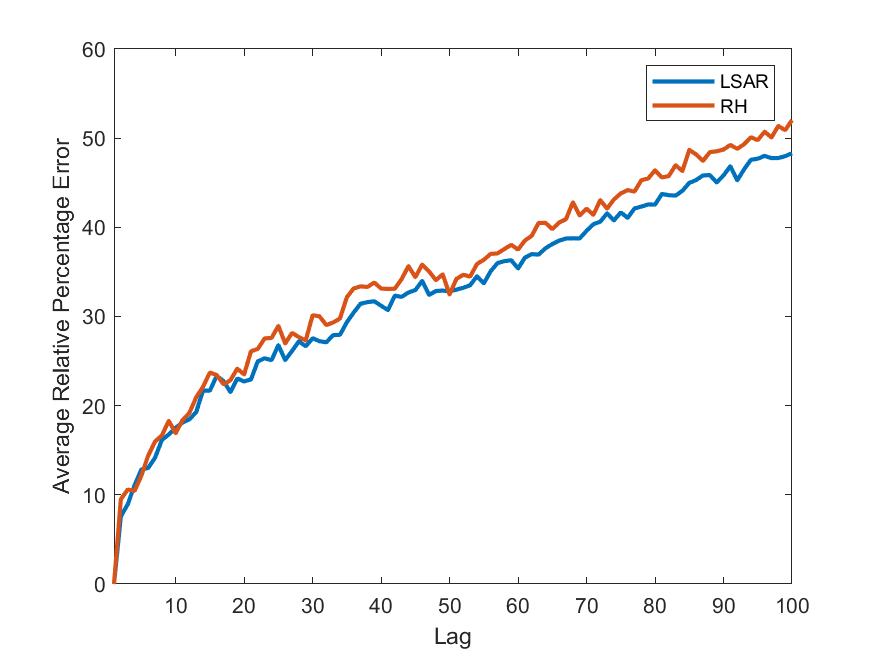}
		\caption{$\mathtt{AR(50)}$ }
		\label{fig:AR50_Error}
	\end{subfigure}
	\begin{subfigure}{.49\textwidth}
		\includegraphics[width=\textwidth]{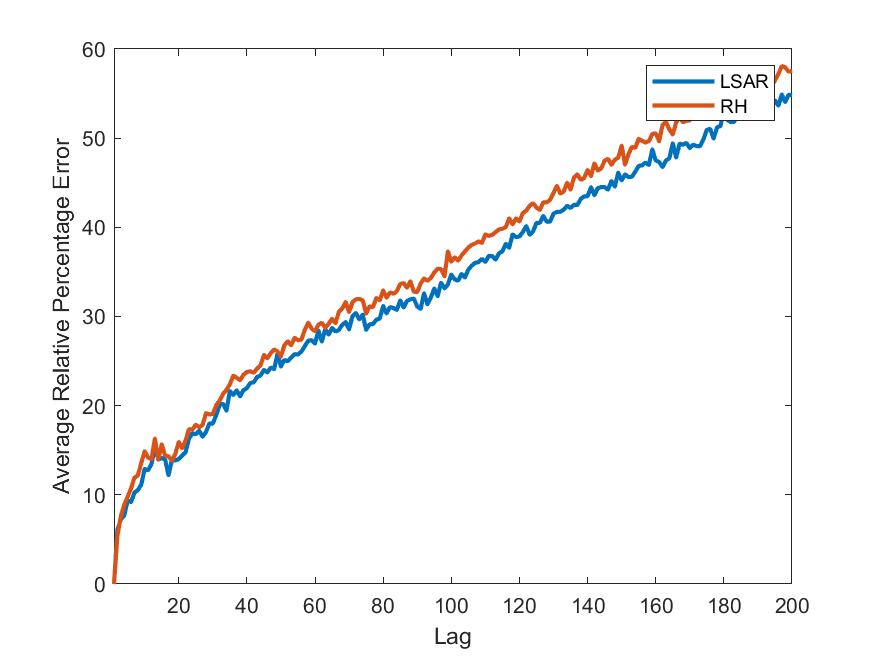}
		\caption{$\mathtt{AR(100)}$ }
		\label{fig:AR100_Error}
	\end{subfigure}
	\begin{subfigure}{.49\textwidth}
		\includegraphics[width=\textwidth]{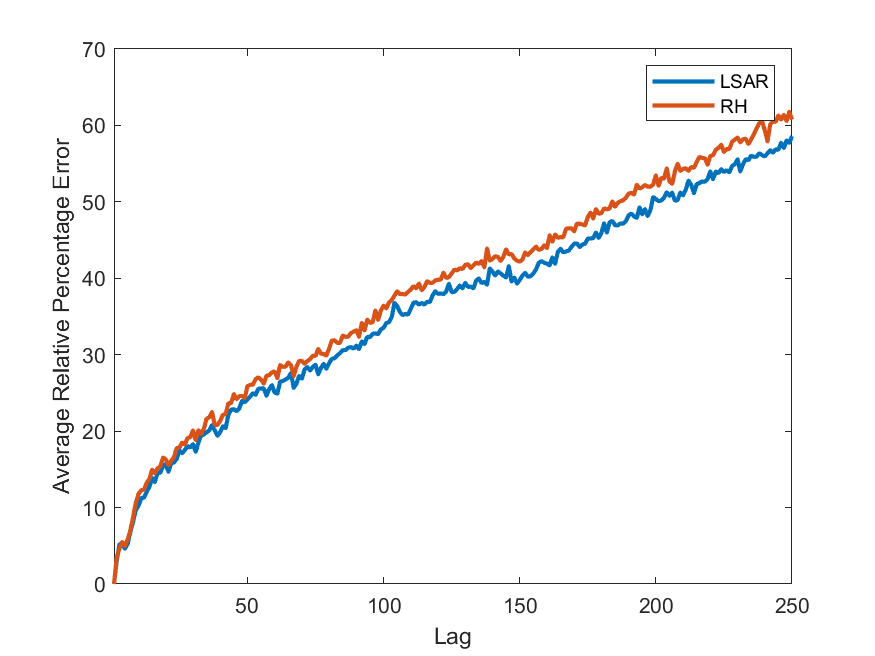}
		\caption{$\mathtt{AR(150)}$ }
		\label{fig:AR150_Error}
	\end{subfigure}
	\caption[Average Percentage Relative Error of \texttt{LSAR}, \texttt{RH} and Exact Algorithms on Synthetic Data Without Outliers]{Figures (a) to (f) corresponding to the labeled \texttt{AR}$(p)$ models, show the percentage relative error in $\bm \phi$ given by \cref{phierr} at each lag, for values of $\bm \phi$ determined by the \texttt{LSAR} algorithm (in blue) and the Repeated Halving algorithm (in red). The average error was computed after running the algorithms 50 times.}
	\label{fig:ErrorNoOuts}
\end{figure}

%-----------------------------------
\subsubsection{Estimation Quality}
\label{sec:ErrorNoOuts}
%-----------------------------------

To look at the estimation quality of each algorithm we compare how well they find the maximum likelihood estimates $\bm \phi$ of the models' parameters at each lag.

When fitting an \texttt{AR}$(p)$ model, we find the maximum likelihood estimates of  $\bm \phi$ at each lag. Each algorithm derives estimates of $\bm{\widehat \phi}^s_{p, h}$ based on the compressed data matrices. We can also calculate the estimate of $\bm {\widehat\phi}_{p, h}$ exactly using the full data matrix. To examine the quality of the algorithms, we wish to look at the relative difference between each algorithm's maximum likelihood estimates of the parameters (based on the compressed data matrices) and the estimate based on the full data matrix (the exact algorithm). For this 
purpose we define the relative percentage error as
\begin{align}
    \label{phierr}
    \frac{\|\bm{\widehat \phi}^s_{p, h} - \bm{\widehat \phi}_{p, h}\|}{\|\bm{\widehat \phi}_{p, h}\|} \times 100.
\end{align}

\cref{fig:ErrorNoOuts} compares the average relative percentage error, at each lag, between the \texttt{LSAR} algorithm and the \texttt{RH} algorithm for each \texttt{AR}$(p)$ model. Once again we have used 2 million synthetically generated data points, and the hyper-parameter $c$ (the number of sampled rows) was $2,000$ ($0.1\%$ of the data). To smooth out the error curves, the algorithms were repeated 50 times and the mean of error at each lag was computed after excluding 5\% of the data values at each end of the data set. This was done to remove outliers.

As we can see in \cref{fig:ErrorNoOuts}, despite the \texttt{LSAR} and \texttt{RH} algorithms taking very different approaches to obtaining leverage scores for sampling the data, the difference in the resultant estimated parameters in negligible.

%-----------------------------------
\subsubsection{PACF Plots} \label{sec:PACFNoOuts}
%-----------------------------------

In this section, we compare the PACF plots for each \texttt{AR}$(p)$ synthetic data set. 
The PACF plot is of primary importance in the time series analysis process. We discussed in \cref{sec:TOLS:intro} that in order to estimate the order $p$ of a time series, we can use the PACF (\cref{PACF}). The PACF plot displays the PACF at each lag as a bar graph. The order $p$ is estimated by choosing the largest lag in the PACF plot where the corresponding PACF is outside the 95\% zero confidence boundary. 

\begin{figure}[p]  % Fig 3
	\centering
	%AR5
	\begin{subfigure}{.32\textwidth}
		\includegraphics[width=\textwidth]{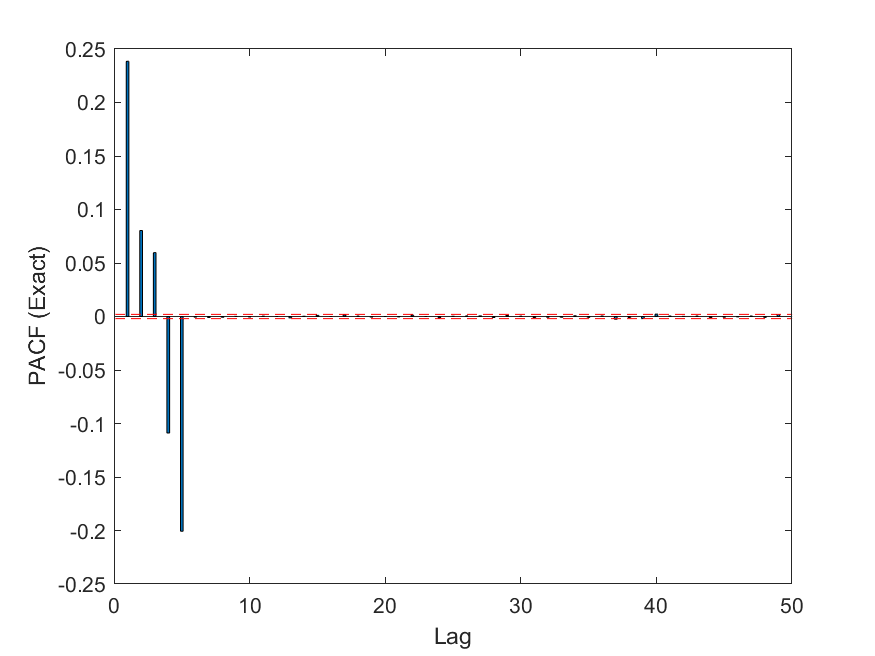}
		\caption{$\mathtt{AR(5)}$ }
		\label{fig:AR5_PACF_n}
	\end{subfigure}
	\begin{subfigure}{.32\textwidth}
		\includegraphics[width=\textwidth]{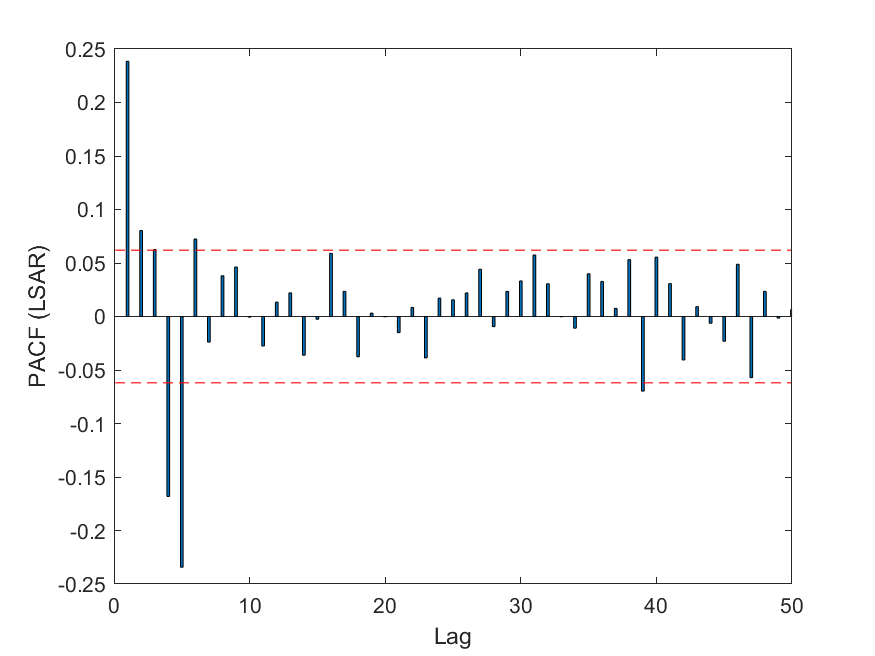}
		\caption{$\mathtt{AR(5)}$ }
		\label{fig:AR5_PACF_s}
	\end{subfigure}
	\begin{subfigure}{.32\textwidth}
		\includegraphics[width=\textwidth]{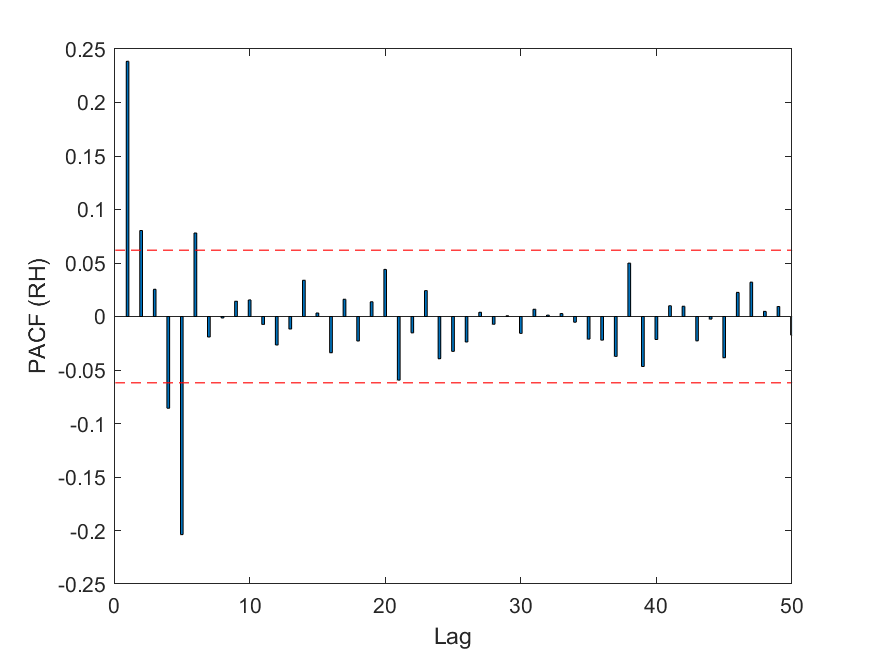}
		\caption{$\mathtt{AR(5)}$ }
		\label{fig:AR5_PACF_d}
	\end{subfigure}

	%AR10
	\begin{subfigure}{.32\textwidth}
		\includegraphics[width=\textwidth]{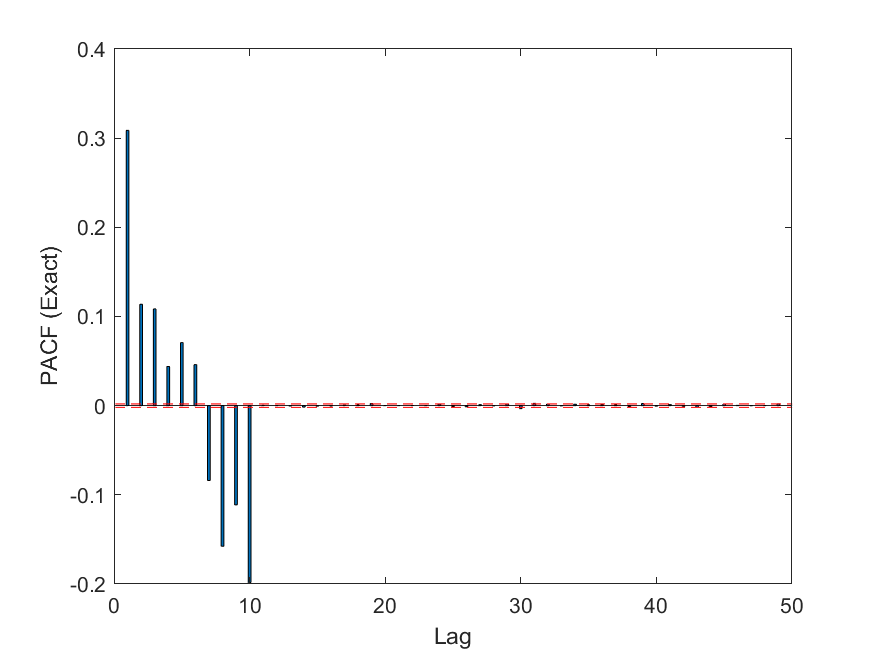}
		\caption{$\mathtt{AR(10)}$ }
		\label{fig:AR10_PACF_n}
	\end{subfigure}
	\begin{subfigure}{.32\textwidth}
		\includegraphics[width=\textwidth]{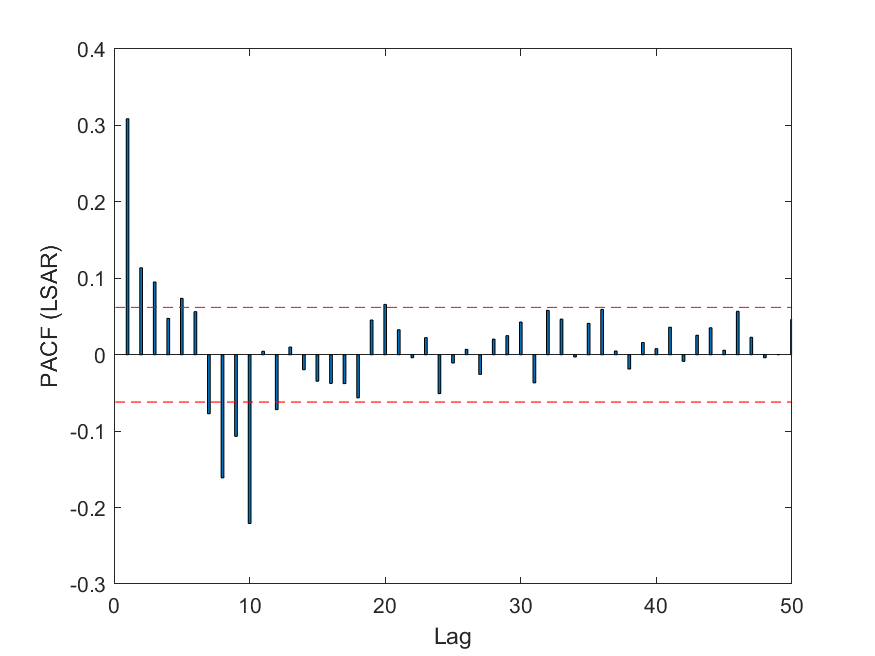}
		\caption{$\mathtt{AR(10)}$ }
		\label{fig:AR10_PACF_s}
	\end{subfigure}
	\begin{subfigure}{.32\textwidth}
		\includegraphics[width=\textwidth]{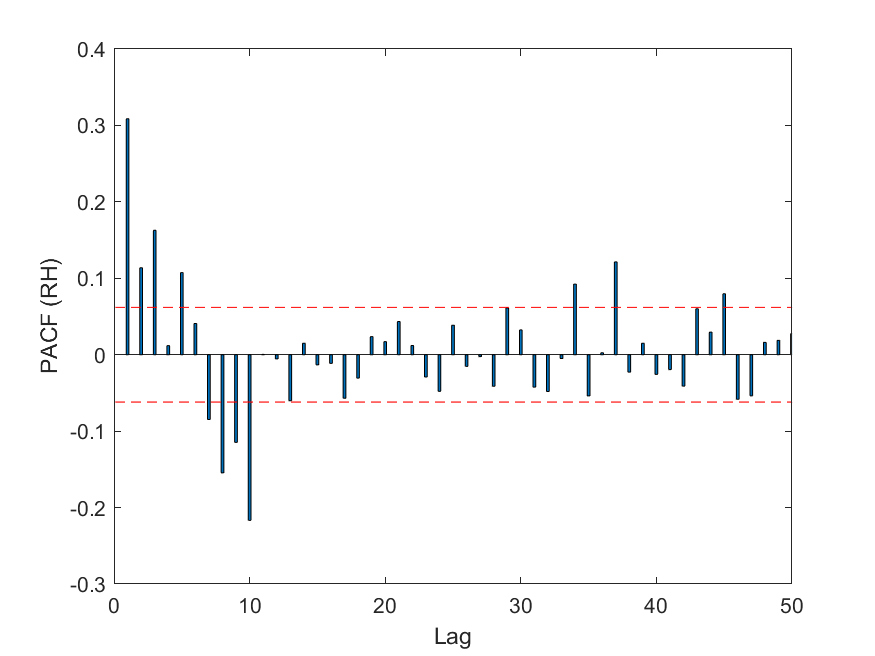}
		\caption{$\mathtt{AR(10)}$ }
		\label{fig:AR10_PACF_d}
    \end{subfigure}
	%AR20

	\begin{subfigure}{.32\textwidth}
		\includegraphics[width=\textwidth]{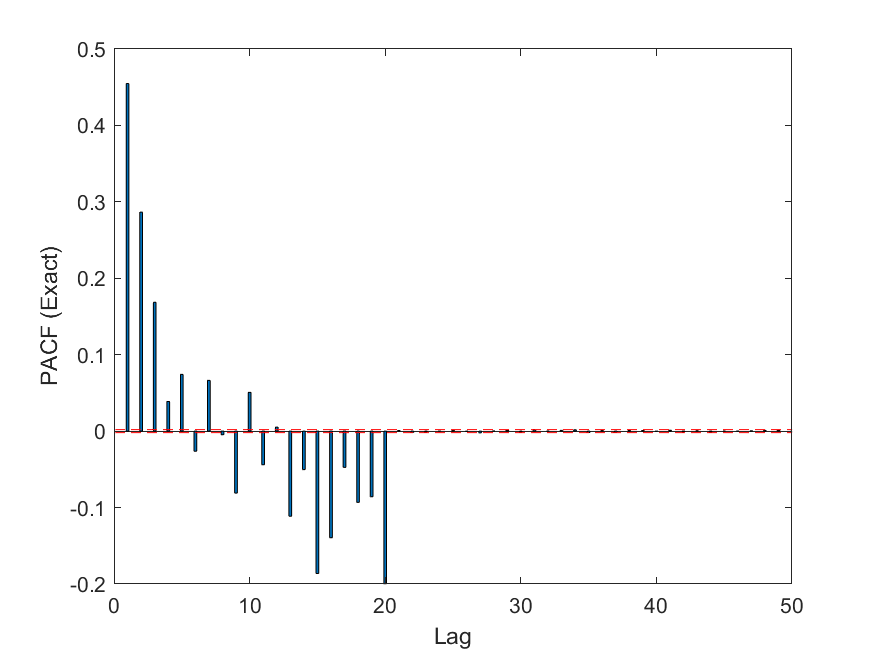}
		\caption{$\mathtt{AR(20)}$ }
		\label{fig:AR20_PACF_n}
	\end{subfigure}
	\begin{subfigure}{.32\textwidth}
		\includegraphics[width=\textwidth]{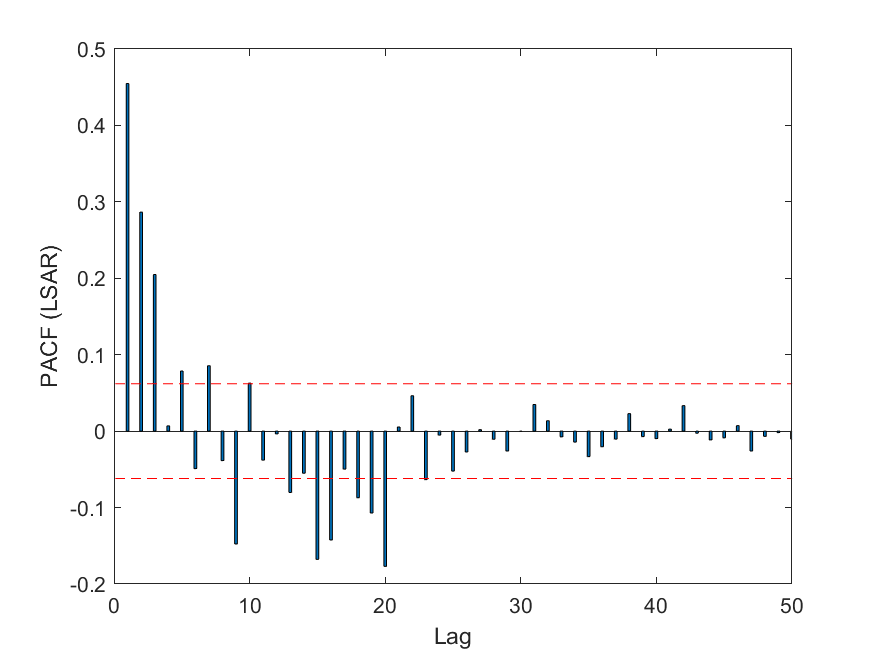}
		\caption{$\mathtt{AR(20)}$ }
		\label{fig:AR20_PACF_s}
	\end{subfigure}
	\begin{subfigure}{.32\textwidth}
		\includegraphics[width=\textwidth]{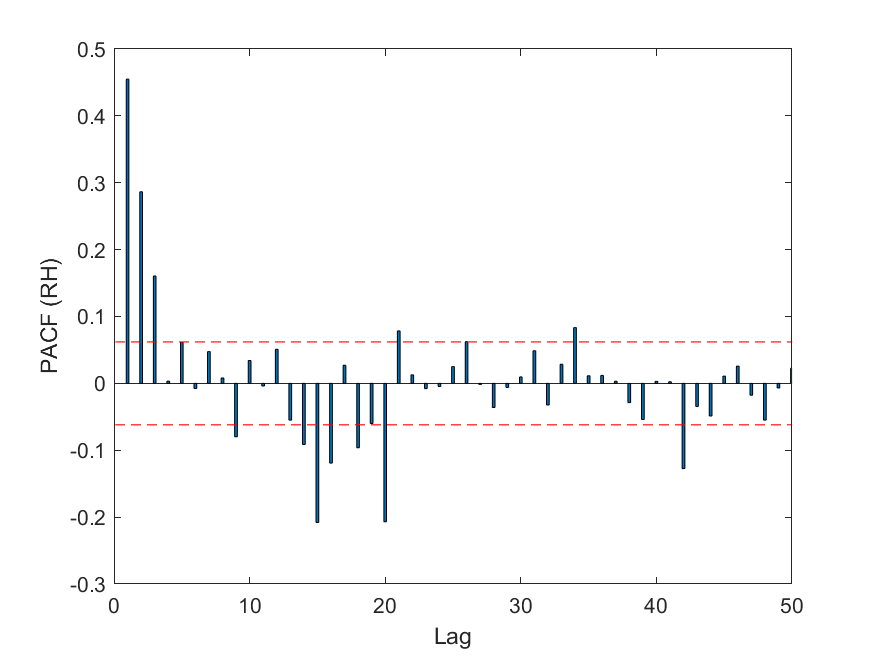}
		\caption{$\mathtt{AR(20)}$ }
		\label{fig:AR20_PACF_d}
	\end{subfigure}
	\caption[PACF Plot Generated by \texttt{LSAR}, \texttt{RH} and Exact Algorithms on Synthetic Data Without Outliers for \texttt{AR}$(5)$, \texttt{AR}$(10)$, \texttt{AR}$(20)$]{Figures (a) to (c), (d) to (f) and (g) to (i) correspond to randomly generated data from \texttt{AR}$(5)$, \texttt{AR}$(10)$ and \texttt{AR}$(20)$ models respectively. For each model we show the PACF plot computed exactly by the \texttt{LSAR} algorithm and by the \texttt{RH} Algorithm. These are displayed from left to right, respectively.  Excluding some noise, we are able use the PACF plots to correctly identify the order $p$ of the data sets, even though the sampled algorithms use only $0.1\%$ of the data.}
	\label{PACFNoOuts5-20}
\end{figure}

\begin{figure}[p]  % Fig 4
    \centering
	\begin{subfigure}{.32\textwidth}
		\includegraphics[width=\textwidth]{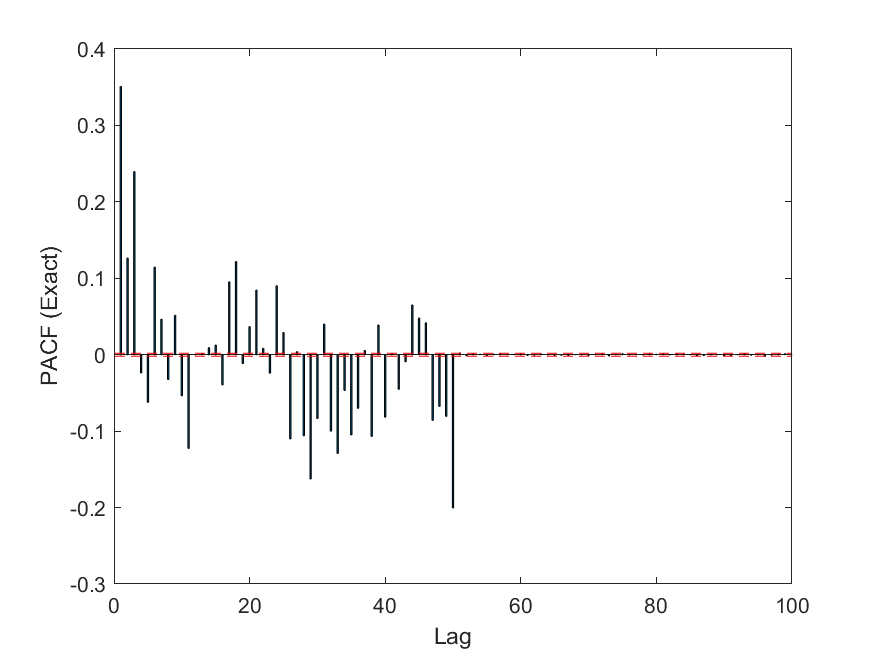}
		\caption{$\mathtt{AR(50)}$ }
		\label{fig:AR50_PACF_n}
	\end{subfigure}
	\begin{subfigure}{.32\textwidth}
		\includegraphics[width=\textwidth]{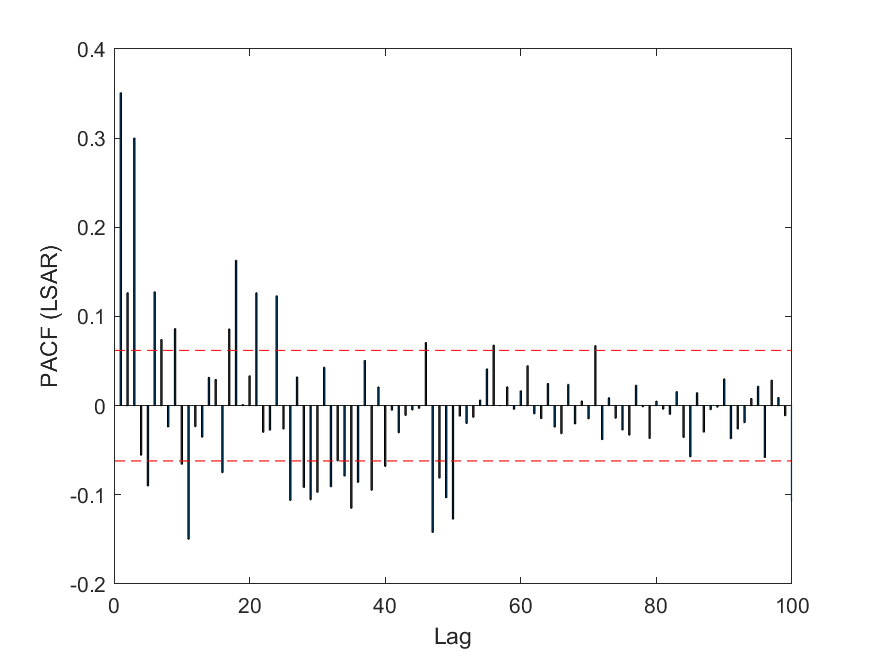}
		\caption{$\mathtt{AR(50)}$ }
		\label{fig:AR50_PACF_s}
	\end{subfigure}
	\begin{subfigure}{.32\textwidth}
		\includegraphics[width=\textwidth]{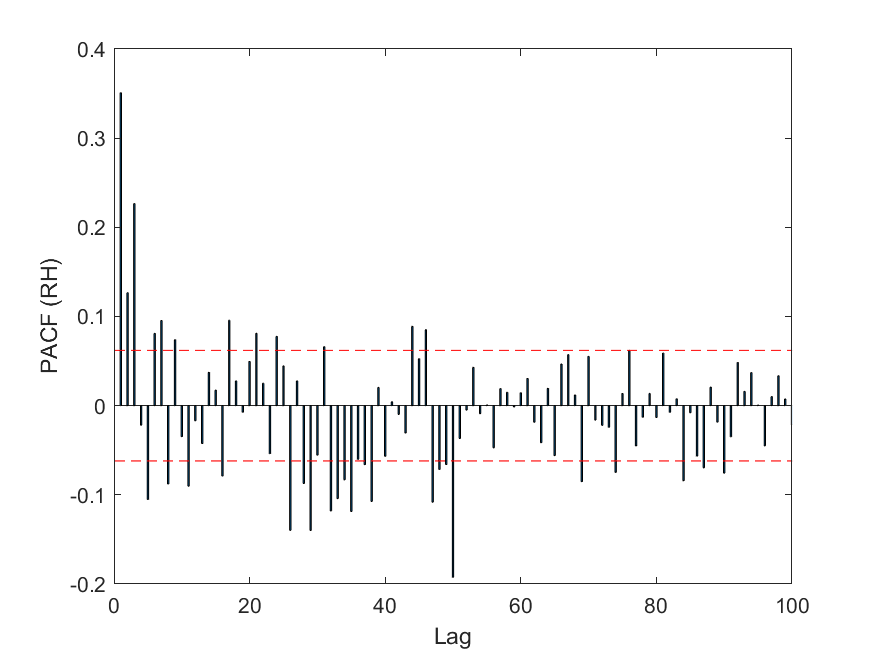}
		\caption{$\mathtt{AR(50)}$ }
		\label{fig:AR50_PACF_d}
	\end{subfigure}

	%AR100

	\begin{subfigure}{.32\textwidth}
		\includegraphics[width=\textwidth]{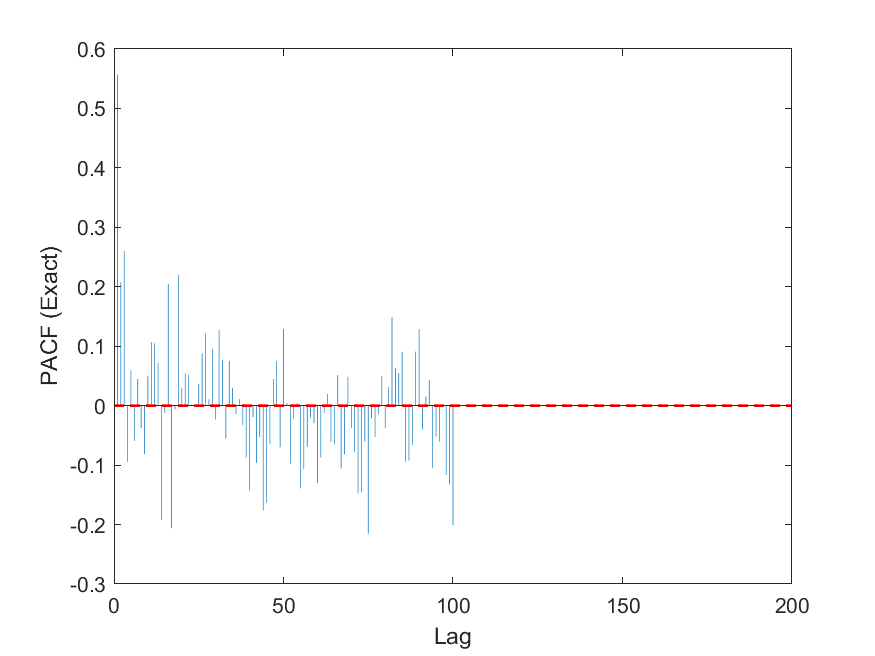}
		\caption{$\mathtt{AR(100)}$ }
		\label{fig:AR100_PACF_n}
	\end{subfigure}
	\begin{subfigure}{.32\textwidth}
		\includegraphics[width=\textwidth]{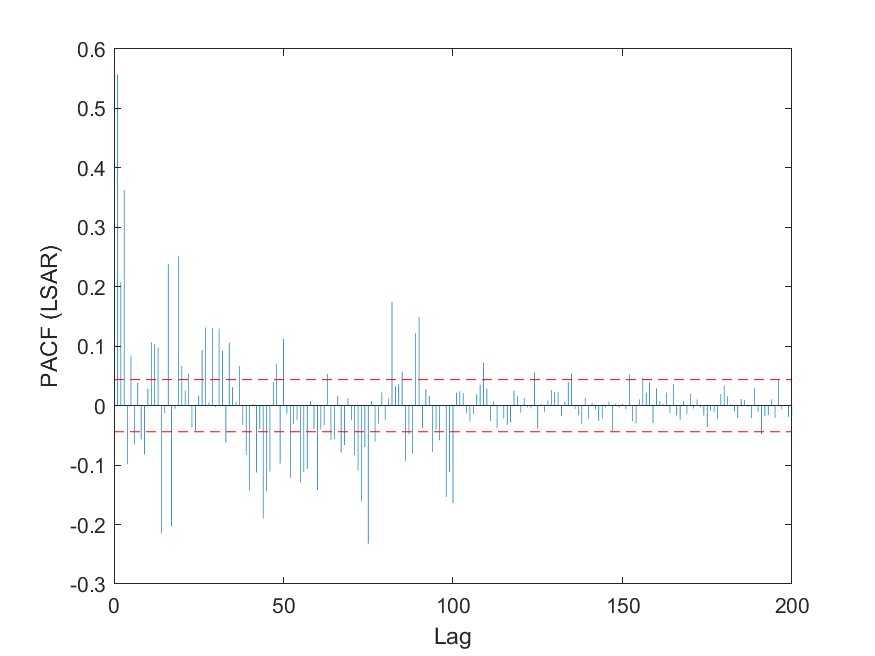}
		\caption{$\mathtt{AR(100)}$ }
		\label{fig:AR100_PACF_s}
	\end{subfigure}
	\begin{subfigure}{.32\textwidth}
		\includegraphics[width=\textwidth]{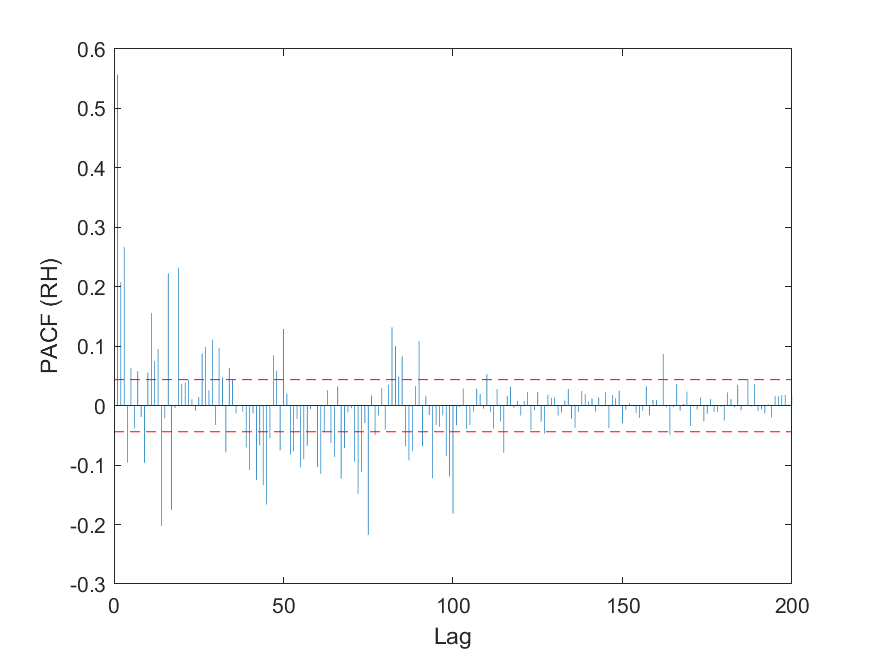}
		\caption{$\mathtt{AR(100)}$ }
		\label{fig:AR100_PACF_d}
	\end{subfigure}

	%AR150

    \begin{subfigure}{.32\textwidth}
		\includegraphics[width=\textwidth]{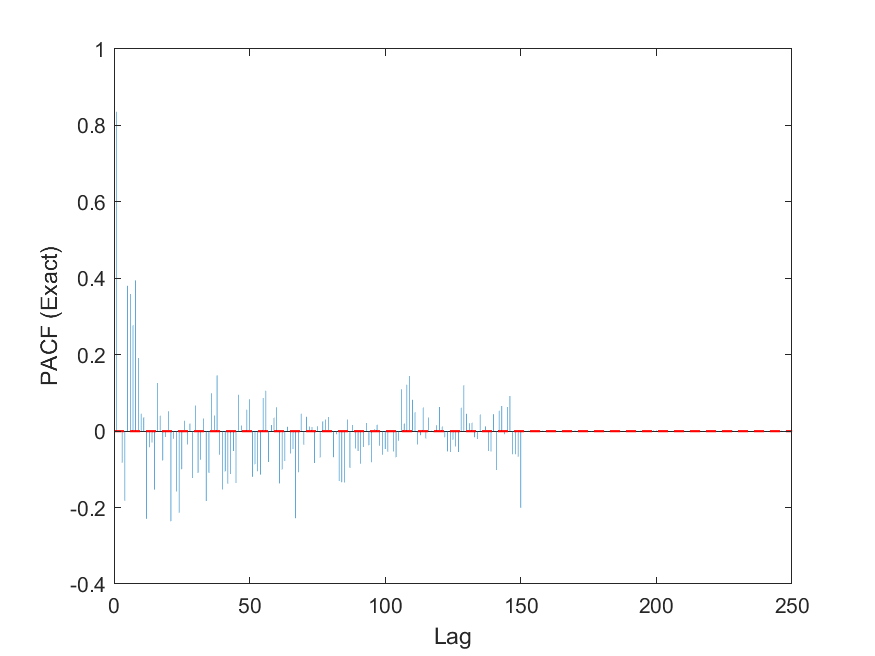}
		\caption{$\mathtt{AR(150)}$ }
		\label{fig:AR150_PACF_n}
	\end{subfigure}
	\begin{subfigure}{.32\textwidth}
		\includegraphics[width=\textwidth]{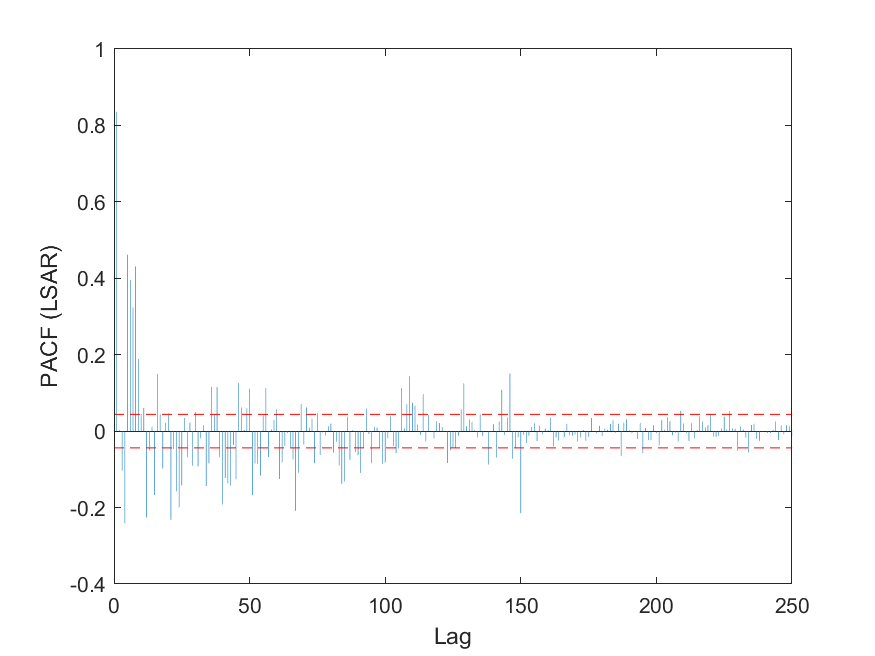}
		\caption{$\mathtt{AR(150)}$ }
		\label{fig:AR150_PACF_s}
	\end{subfigure}
	\begin{subfigure}{.32\textwidth}
		\includegraphics[width=\textwidth]{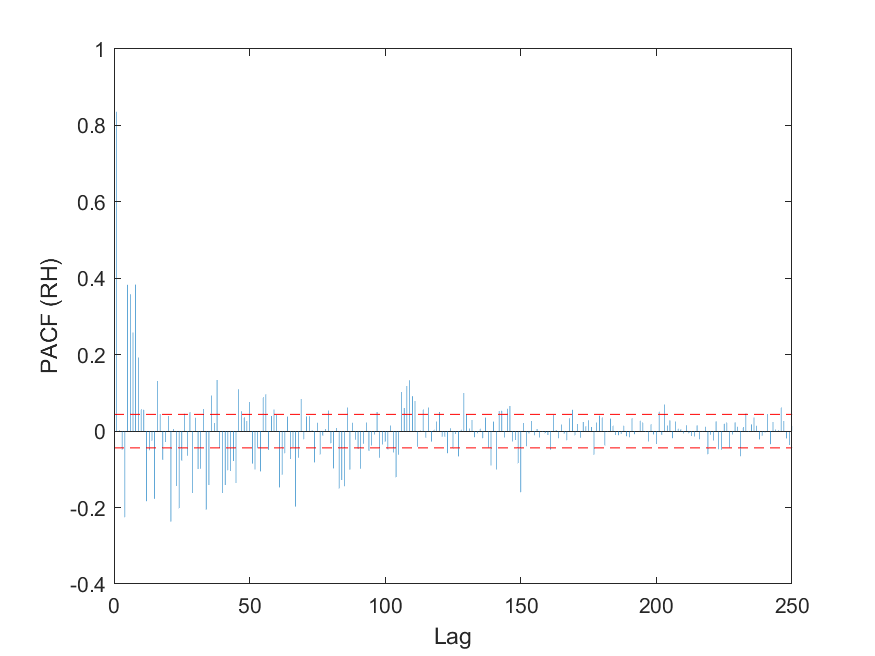}
		\caption{$\mathtt{AR(150)}$ }
		\label{fig:AR150_PACF_d}
	\end{subfigure}
	\caption[PACF Plot Generated by \texttt{LSAR}, \texttt{RH} and Exact Algorithms on Synthetic Data Without Outliers for \texttt{AR}$(50)$, \texttt{AR}$(100)$, \texttt{AR}$(150)$]{Figures (a) to (c), (d) to (f) and (g) to (i) correspond to randomly generated data from \texttt{AR}$(50)$, \texttt{AR}$(100)$ and \texttt{AR}$(150)$ models respectively. For each model we show the PACF plot computed exactly by the \texttt{LSAR} algorithm and by the \texttt{RH} Algorithm. These are displayed from left to right, respectively.  Excluding some noise, we are able to use the PACF plots to correctly identify the order $p$ of the data sets, even though the sampled algorithms use only $0.1\%$ of the data.}
	\label{PACFNoOuts50-150}
\end{figure}

\cref{PACFNoOuts5-20,,PACFNoOuts50-150} display the PACF plots generated by each algorithm, for all synthetic data sets. We estimate the PACF for each lag up to $\bar p$, first using the full data matrix, then twice more using the PACF obtained by each of the compressed data matrices of the \texttt{LSAR} and \texttt{RH} algorithms respectively. For each \texttt{AR}$(p)$ model we use the same data sets from \cref{sec:Time}, with $ n = 2,000,000$ and number of sampled rows $c = 2,000$. The dashed red error lines indicate the 95\% zero confidence boundary.

In \cref{PACFNoOuts5-20,,PACFNoOuts50-150} we are able to obtain a correct estimate of the order $p$ for the generated synthetic data from each of the exact, \texttt{LSAR} and \texttt{RH} algorithms. All PACF plots generated by the compressed data matrices appear to be quite similar to the corresponding exact PACF plots. This is by sampling only 0.1\% of the rows of the data matrix. 

There is clearly some error at each lag of the compressed algorithms. This is particularly evident at lags greater than the order of the model, which should be closer to zero. However, while we should be aware of this error, it must be noted that it does not affect the estimation of the order in the synthetic data examples that we present. Furthermore, we are able to obtain these reasonable approximations of the PACF in a significantly reduced time when compared to the exact alternative. 

	%AR50

\begin{figure}[t]  % Fig 5
    \begin{subfigure}{.49\textwidth}
		\includegraphics[width=\textwidth]{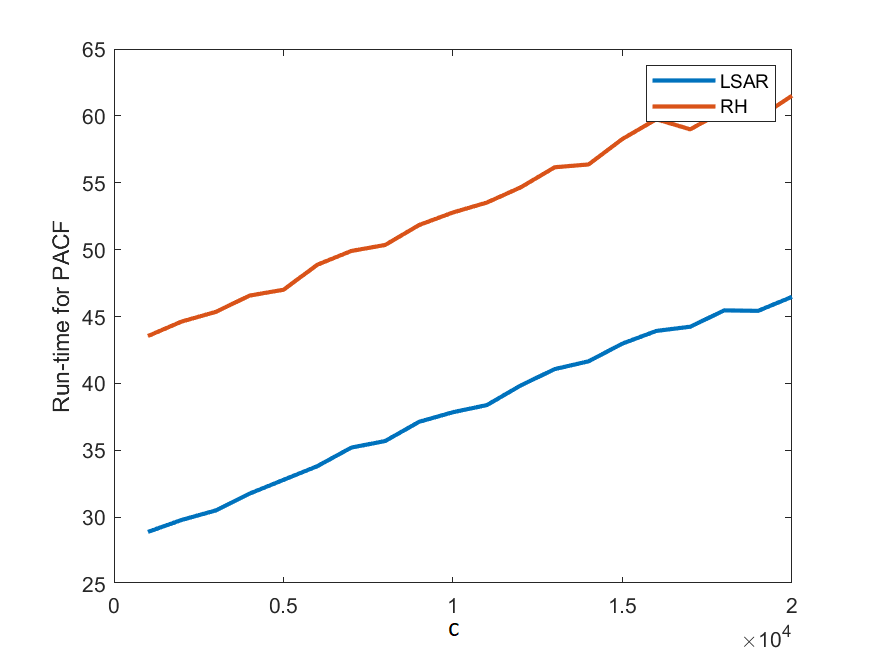}
		\caption{Time}
		\label{PER1000_time}
	\end{subfigure}
	\begin{subfigure}{.49\textwidth}
		\includegraphics[width=\textwidth]{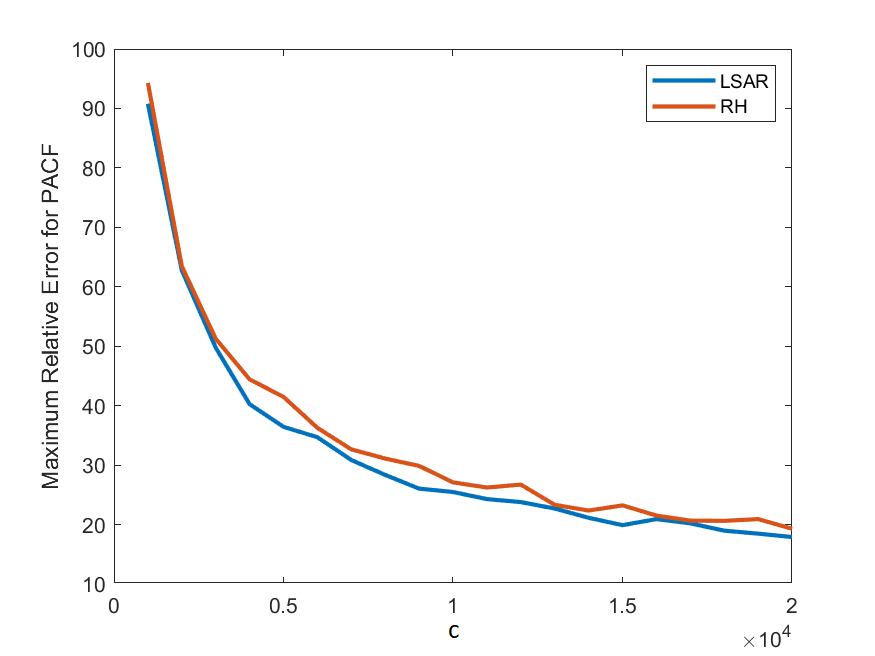}
		\caption{Error}
		\label{PER1000_Error}
	\end{subfigure}
	\caption[The Effect of Sample Size on Computation Time and Error]{Figure (a) displays the run time of the \texttt{LSAR} algorithm (in blue) and the \texttt{RH} algorithm (in red) when run multiple time over different values of $c$. Figure (b) displays the maximum point-wise value of relative percentage error for each of the algorithms.}
	\label{overc}
\end{figure}

%-----------------------------------
\subsubsection{The Effect of Sample Size}
%-----------------------------------

We recall that the \texttt{LSAR} and \texttt{RH} algorithms have hyper-parameter $c$, the number of rows sampled from the full data matrix to construct the compressed data matrices used by these algorithms. Choosing different values of $c$ leads to a trade-off between run time and accuracy of the compressed algorithms. For example, if we used a larger value of $c$ we would expect our accuracy to increase (or our error to decrease) at the expense of computation time.

\cref{overc} displays the computation time and relative percentage error (given by \cref{phierr}) as $c$ changes for a synthetically generated \texttt{AR}$(100)$ data set with $n = 2,000,000$. Maximum point-wise time and error over the lags was taken for each instance of the algorithm running for $c = \{1000, 2000, \dots, 20,000\}$ and is displayed in \cref{PER1000_time,,PER1000_Error} respectively.

\cref{overc} confirms the discussed trade-off between time and error. Computation time appears to increase linearly as a function of $c$, while the error decreases.

%-----------------------------------
\subsection{Synthetic Data with Outliers}
\label{sec:NRwith}
%-----------------------------------

This section follows a similar pattern to \cref{sec:NRwithout}, except this time with the addition of outliers in the data. Once again, two million realisations from six \texttt{AR}$(p)$ time series models were randomly generated for 
\(p = 5, 10, 20, 50, 100, 150\) 
with coefficients corresponding to a stationary time series model for each order  obtained randomly. Synthetic data was generated with constant 0 and variance 1. One thousand data points were randomly selected and replaced with the sum of the data point, a randomly generated number from a uniform distribution over $[-3, 3]$, and a normally distributed variable with a mean of 0 and variance of 100.

\begin{figure}[p]  % Fig 6
	\centering
	\begin{subfigure}{.49\textwidth}
		\includegraphics[width=\textwidth]{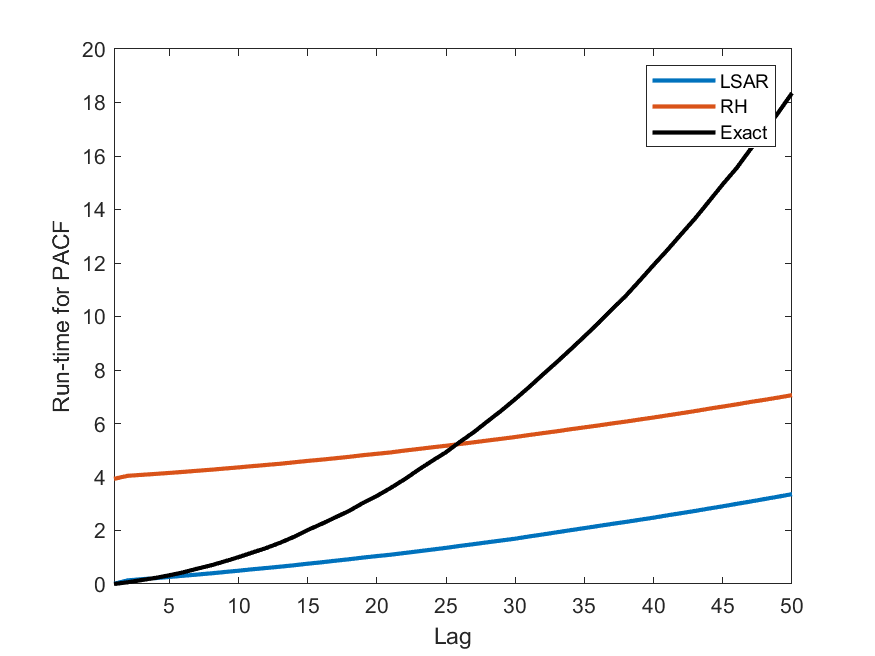}
		\caption{$\mathtt{AR(5)}$ }
		\label{figsout:AR5_Time}
	\end{subfigure}
	\begin{subfigure}{.49\textwidth}
		\includegraphics[width=\textwidth]{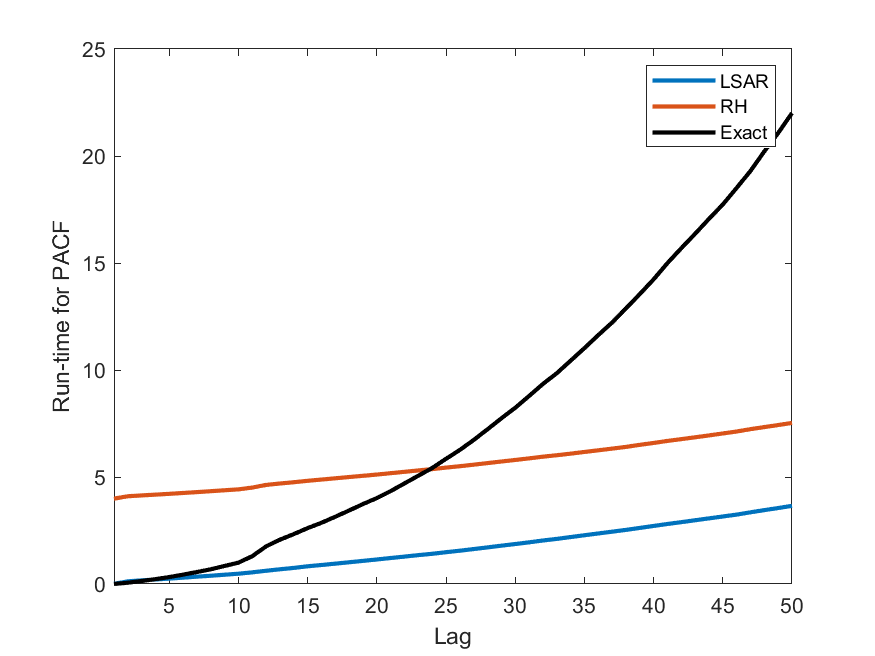}
		\caption{$\mathtt{AR(10)}$ }
		\label{figsout:AR10_Time}
	\end{subfigure}
	\begin{subfigure}{.49\textwidth}
		\includegraphics[width=\textwidth]{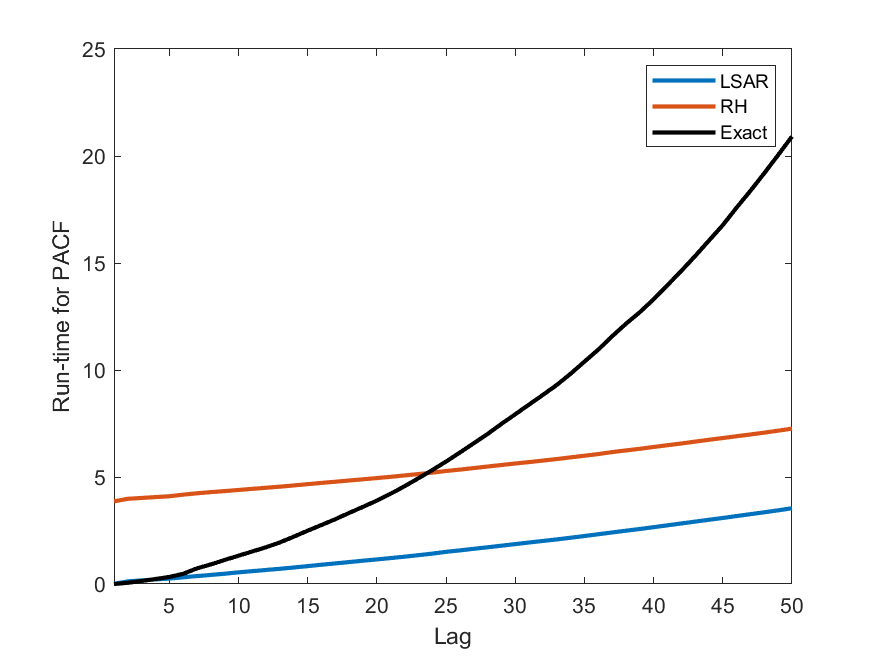}
		\caption{$\mathtt{AR(20)}$ }
		\label{figsout:AR20_Time}
	\end{subfigure}
	\begin{subfigure}{.49\textwidth}
		\includegraphics[width=\textwidth]{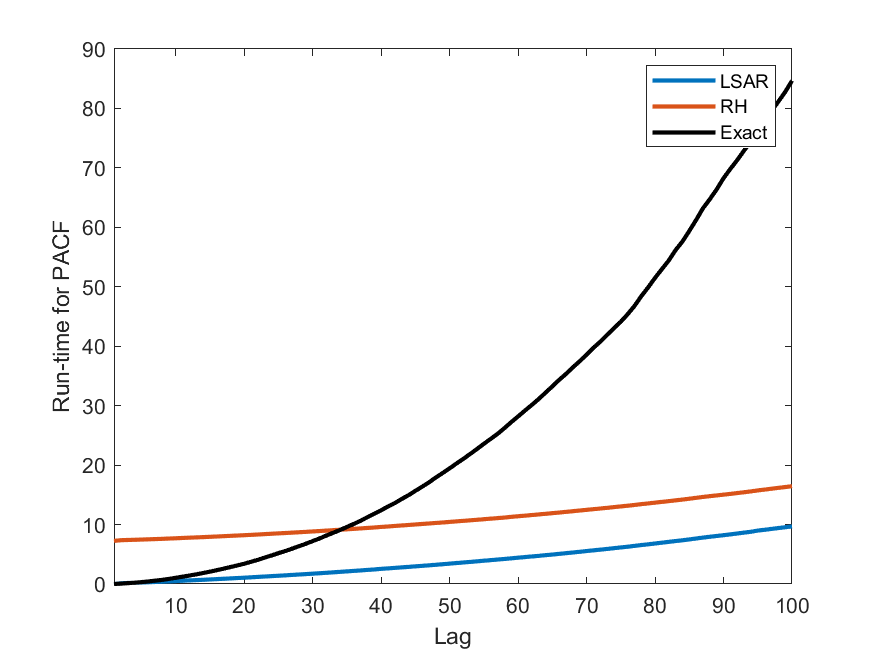}
		\caption{$\mathtt{AR(50)}$ }
		\label{figsout:AR50_Time}
	\end{subfigure}
	\begin{subfigure}{.49\textwidth}
		\includegraphics[width=\textwidth]{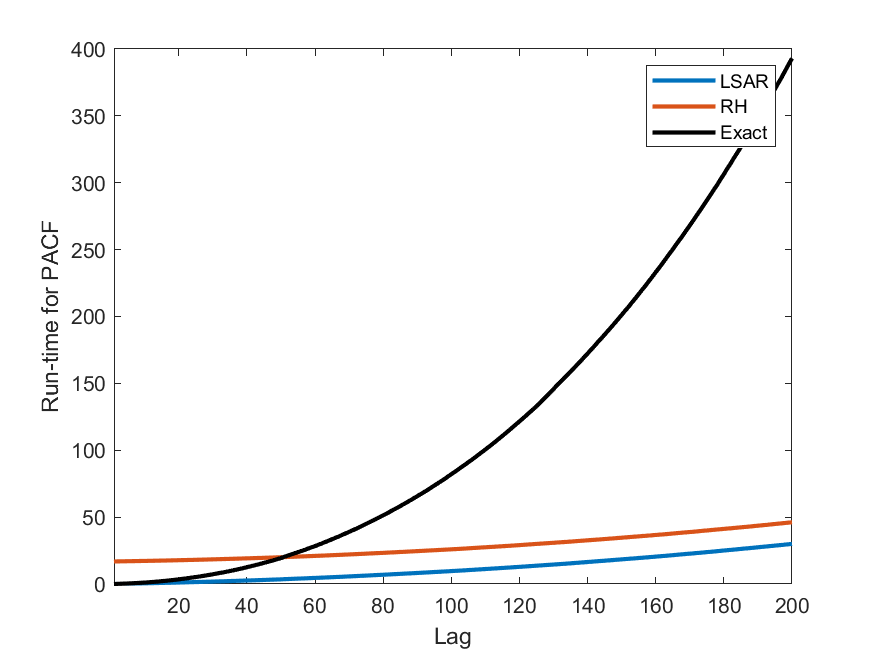}
		\caption{$\mathtt{AR(100)}$ }
		\label{figsout:AR100_Time}
	\end{subfigure}
	\begin{subfigure}{.49\textwidth}
		\includegraphics[width=\textwidth]{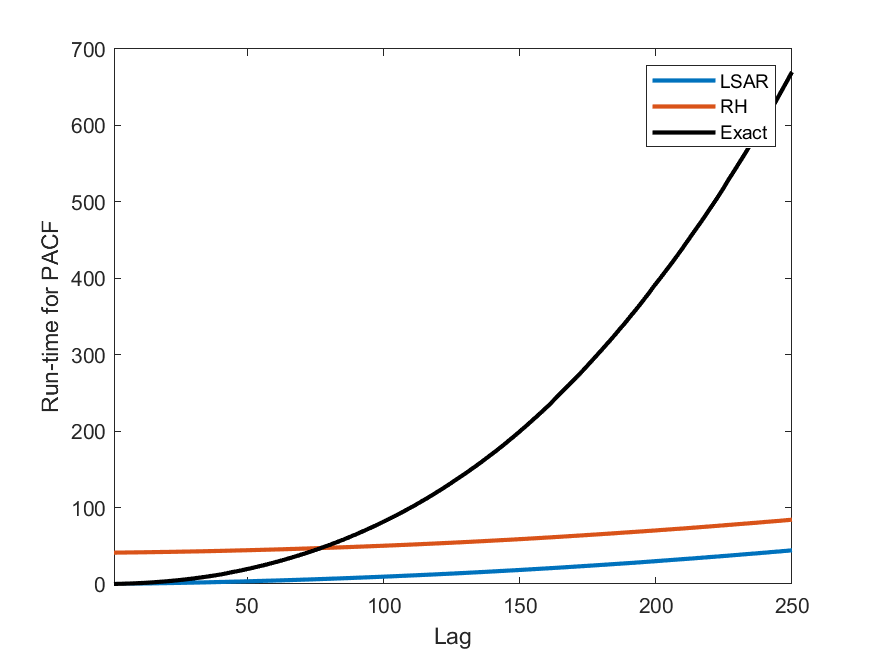}
		\caption{$\mathtt{AR(150)}$ }
		\label{figsout:AR150_Time}
	\end{subfigure}
	\caption[Computation Time to Compute the PACF of \texttt{LSAR}, \texttt{RH} and Exact Algorithms on Synthetic Data With Outliers]{Figures (a) to (f) corresponding to the labeled \texttt{AR}$(p)$ models, compare the computation time (in seconds) to generate the PACF for the \texttt{LSAR} algorithm (in blue), the Repeated Halving algorithm (in red) and the exact computation of PACF (in black).}
	\label{figsout:Time}
\end{figure}

%-----------------------------------
\subsubsection{Computational Time of the PACF} 
\label{sec:TimeYesOuts}
%-----------------------------------

Similarly to \cref{sec:Time}, \cref{figsout:Time} compares the run time to compute the PACF by the \texttt{LSAR} algorithm, the \texttt{RH} algorithm, and the exact calculation, for each \texttt{AR}$(p)$ model. Time is plotted cumulatively over each lag. 

The computation times of the algorithms, including the exact computations, are unaffected by the presence of outliers. We note  that computation times are approximately the same as those in \cref{sec:Time} because we are performing calculations on matrices of the same size. The \texttt{RH}  %\cref{RHA} and 
and \texttt{LSAR} % \cref{RandLSA} we note that
algorithms appear to have similar computation times, separated by a constant, which is due to the \texttt{RH} algorithm computing leverage scores prior to the algorithm's iteration over the lag.
In particular, the difference in computation time is exemplified by \cref{figsout:AR150_Time}, which presents a 700-second difference between the exact method and the two compressed methods.

%-----------------------------------
\subsubsection{Estimation Quality}
%-----------------------------------

We examine the estimation quality of each algorithm, as in \cref{sec:ErrorNoOuts}, by comparing how well they find the maximum likelihood estimates $\bm \phi$ of the models' parameters at each lag. This time, the models include outliers.

\begin{figure}[p]  % Fig 7
	\centering
	\begin{subfigure}{.49\textwidth}
		\includegraphics[width=\textwidth]{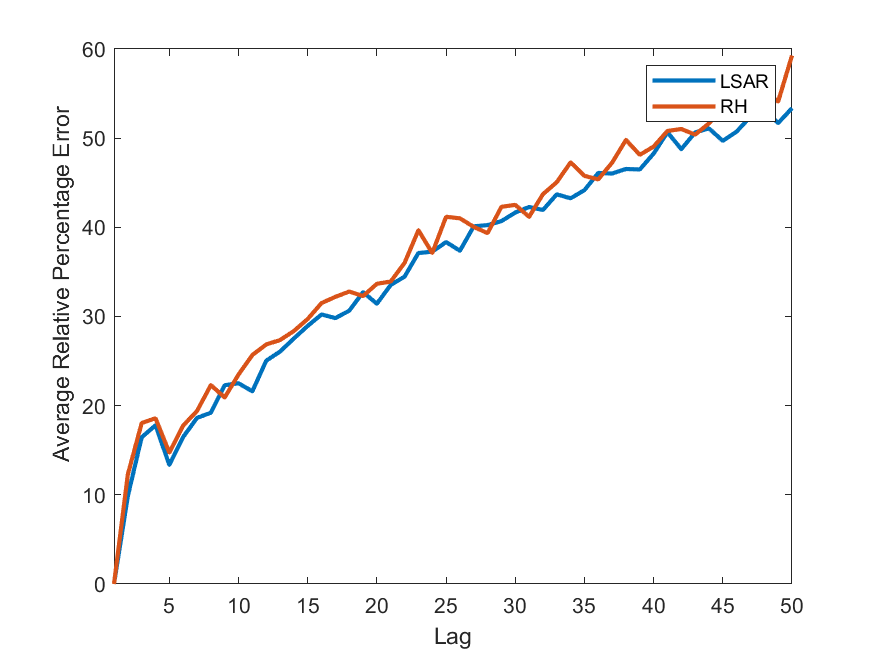}
		\caption{$\mathtt{AR(5)}$ }
		\label{figsout:AR5_Error}
	\end{subfigure}
	\begin{subfigure}{.49\textwidth}
		\includegraphics[width=\textwidth]{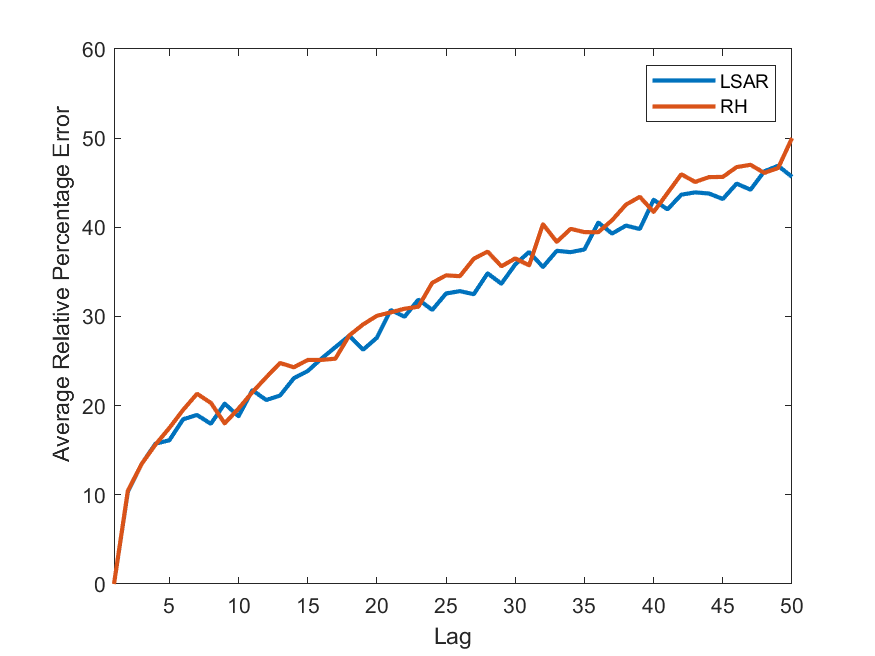}
		\caption{$\mathtt{AR(10)}$ }
		\label{figsout:AR10_Error}
	\end{subfigure}
    \begin{subfigure}{.49\textwidth}
		\includegraphics[width=\textwidth]{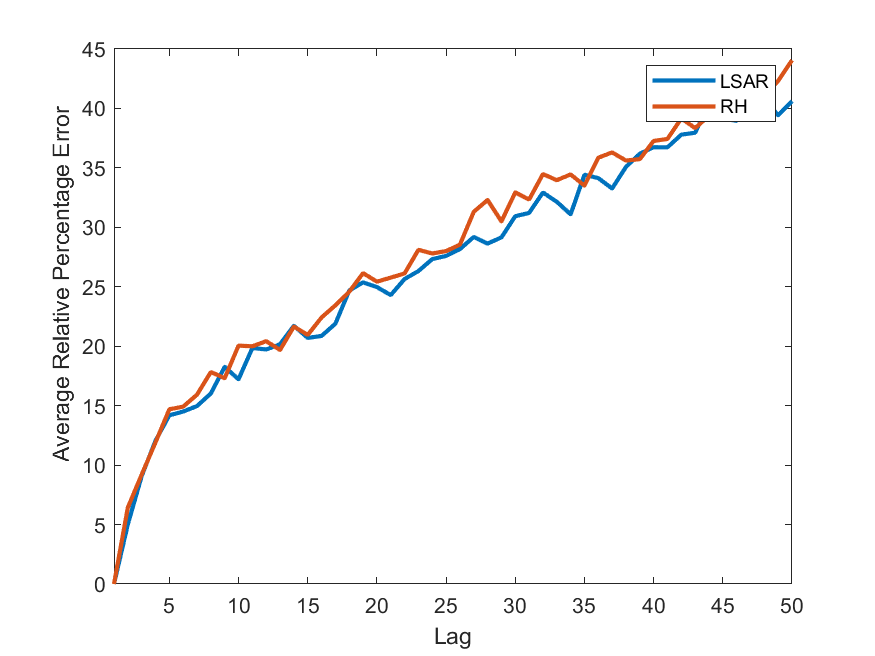}
		\caption{$\mathtt{AR(20)}$ }
		\label{figsout:AR20_Error}
	\end{subfigure}
	\begin{subfigure}{.49\textwidth}
		\includegraphics[width=\textwidth]{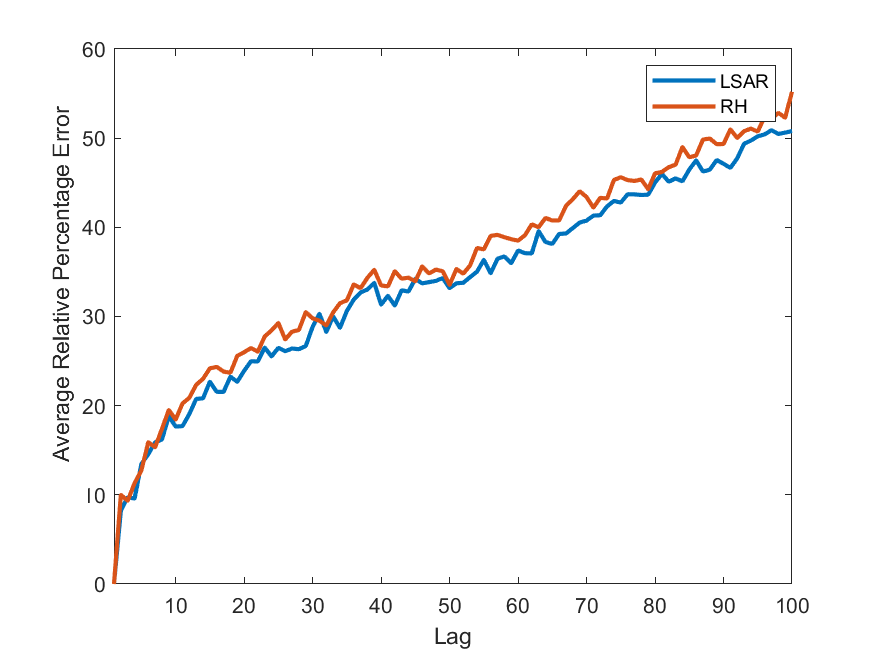}
		\caption{$\mathtt{AR(50)}$ }
		\label{figsout:AR50_Error}
	\end{subfigure}
	\begin{subfigure}{.49\textwidth}
		\includegraphics[width=\textwidth]{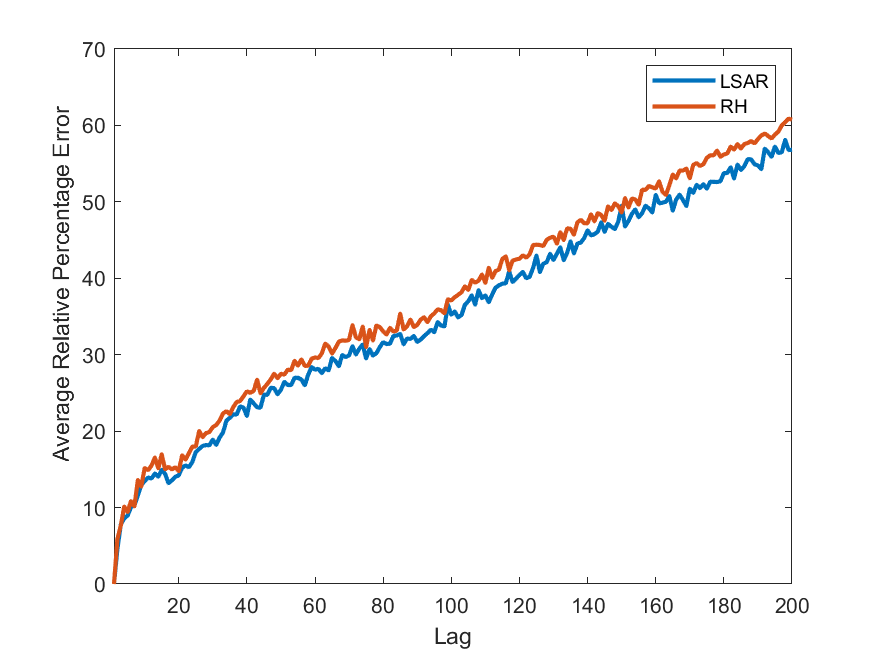}
		\caption{$\mathtt{AR(100)}$ }
		\label{figsout:AR100_Error}
	\end{subfigure}
	\begin{subfigure}{.49\textwidth}
		\includegraphics[width=\textwidth]{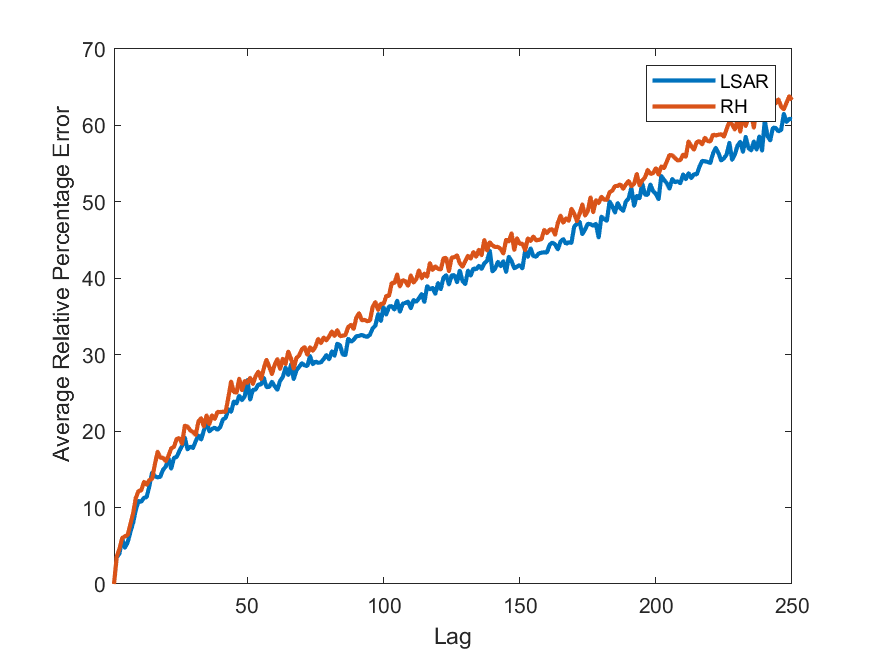}
		\caption{$\mathtt{AR(150)}$ }
		\label{figsout:AR150_Error}
	\end{subfigure}
	\caption[Relative Percentage Error of \texttt{LSAR}, \texttt{RH} and Exact Algorithms on Synthetic Data With Outliers]{Figures (a) to (f) corresponding to the labeled \texttt{AR}$(p)$ models show the percentage relative error in $\bm \phi$ given by \cref{phierr} at each lag, for values of $\bm \phi$ determined by the \texttt{LSAR} algorithm (in blue) and the Repeated Halving algorithm (in red). The average error was computed after running the algorithms 50 times.}
	\label{fig:ErrorYesOuts}
\end{figure}

\cref{fig:ErrorYesOuts} compares the relative percentage error according to \cref{phierr}, at each lag, between the \texttt{LSAR} algorithm and the \texttt{RH} algorithm for each \texttt{AR}$(p)$ model. We have used 2 million synthetically generated data points with 1000 replaced by outliers as discussed in \cref{sec:NRwith}. The number of sampled rows $c$ was $2,000$ ($0.1\%$ of the data). To smooth out the error curves, the algorithms were repeated 50 times and the mean of error at each lag was computed after excluding 5\% of the data values at each end of the data set. This was done to remove outliers. Thus each graph pertains to the average relative percentage error.

In \cref{fig:ErrorYesOuts}, we can again observe that despite \texttt{LSAR} and \texttt{RH} taking very different approaches to obtaining leverage scores for sampling the data, the difference between the resultant estimated parameters is negligible. There also appears to be negligible difference between the errors of the algorithms for the data sets with and without outliers. This would suggest that the presence of outliers has very little effect on the estimated parameters for each algorithm.

%-----------------------------------
\subsubsection{PACF Plots}
%-----------------------------------

\begin{figure}[p]  % Fig 8
	\centering
	%AR5
	\begin{subfigure}{.32\textwidth}
		\includegraphics[width=\textwidth]{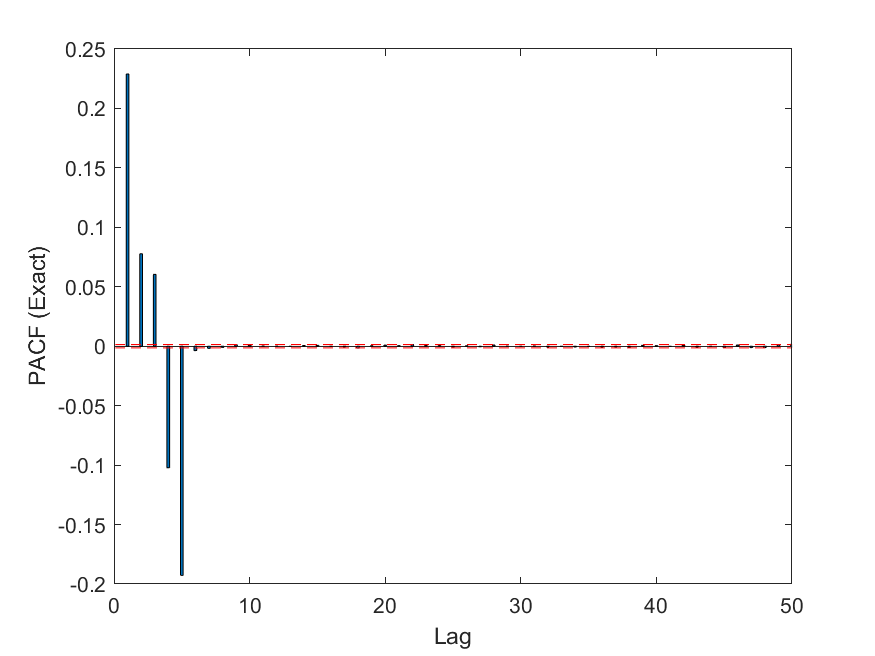}
		\caption{$\mathtt{AR(5)}$ }
		\label{figsout:AR5_PACF_n}
	\end{subfigure}
	\begin{subfigure}{.32\textwidth}
		\includegraphics[width=\textwidth]{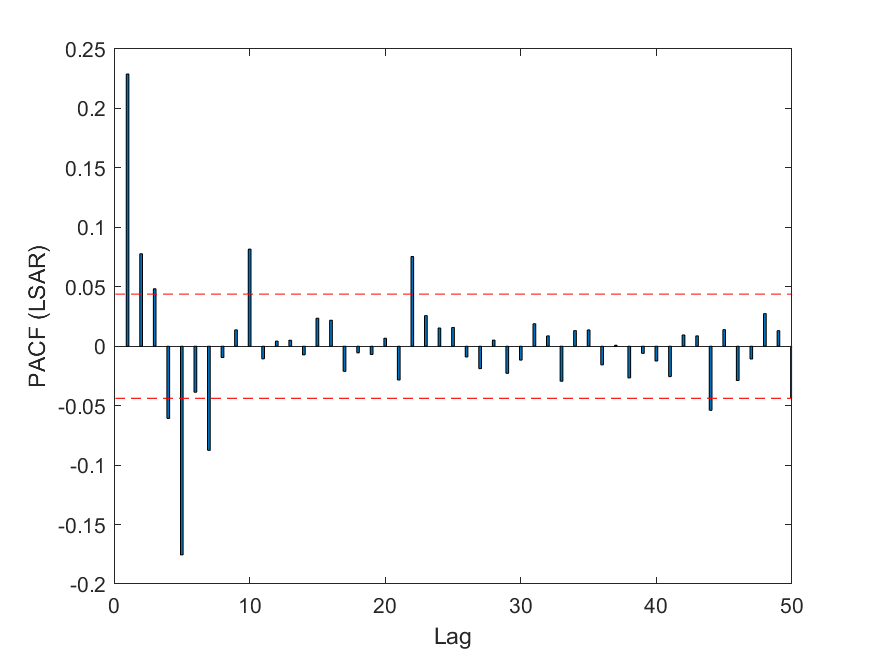}
		\caption{$\mathtt{AR(5)}$ }
		\label{figsout:AR5_PACF_s}
	\end{subfigure}
	\begin{subfigure}{.32\textwidth}
		\includegraphics[width=\textwidth]{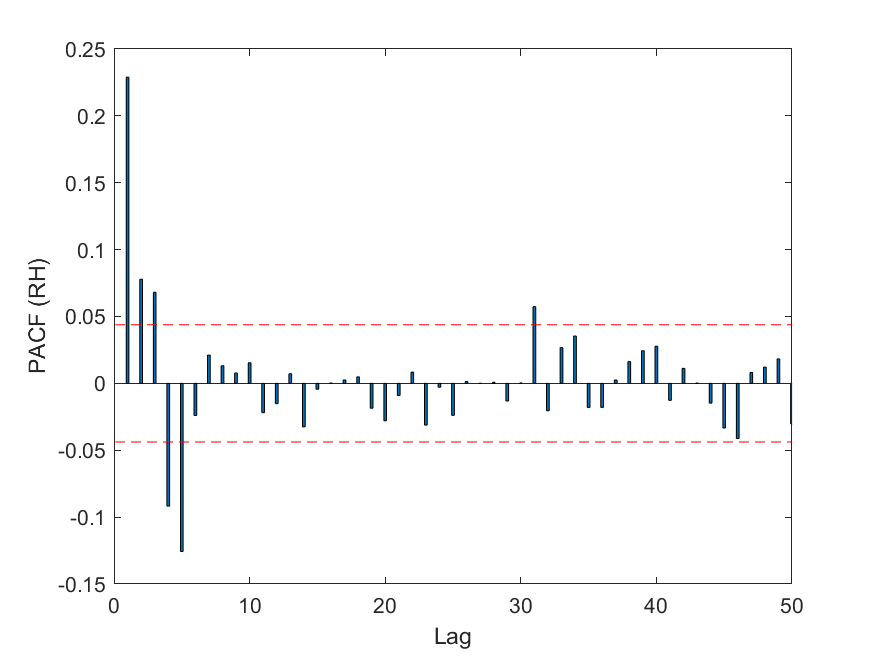}
		\caption{$\mathtt{AR(5)}$ }
		\label{figsout:AR5_PACF_d}
	\end{subfigure}

	%AR10

	\begin{subfigure}{.32\textwidth}
		\includegraphics[width=\textwidth]{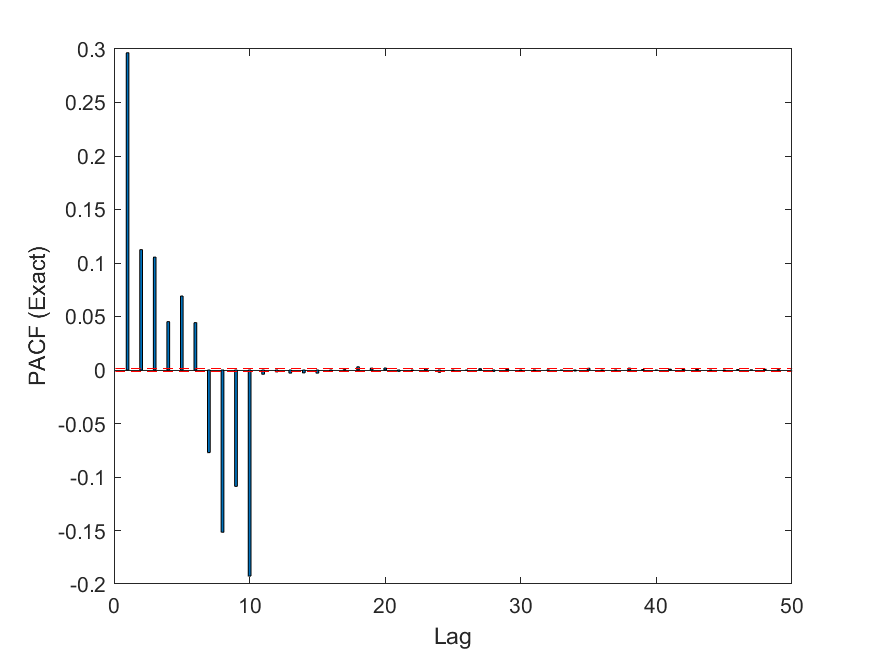}
		\caption{$\mathtt{AR(10)}$ }
		\label{figsout:AR10_PACF_n}
	\end{subfigure}
	\begin{subfigure}{.32\textwidth}
		\includegraphics[width=\textwidth]{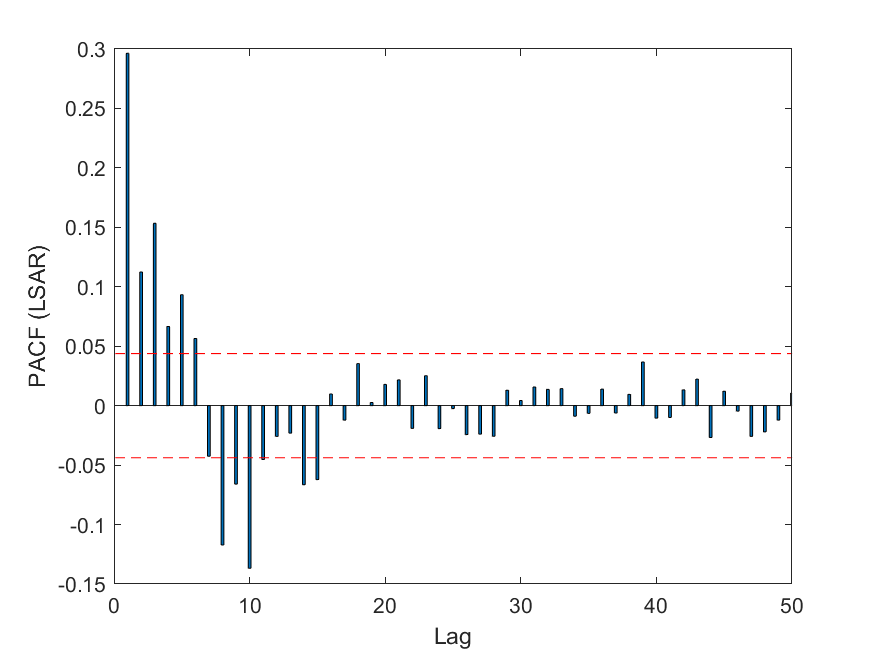}
		\caption{$\mathtt{AR(10)}$ }
		\label{figsout:AR10_PACF_s}
	\end{subfigure}
	\begin{subfigure}{.32\textwidth}
		\includegraphics[width=\textwidth]{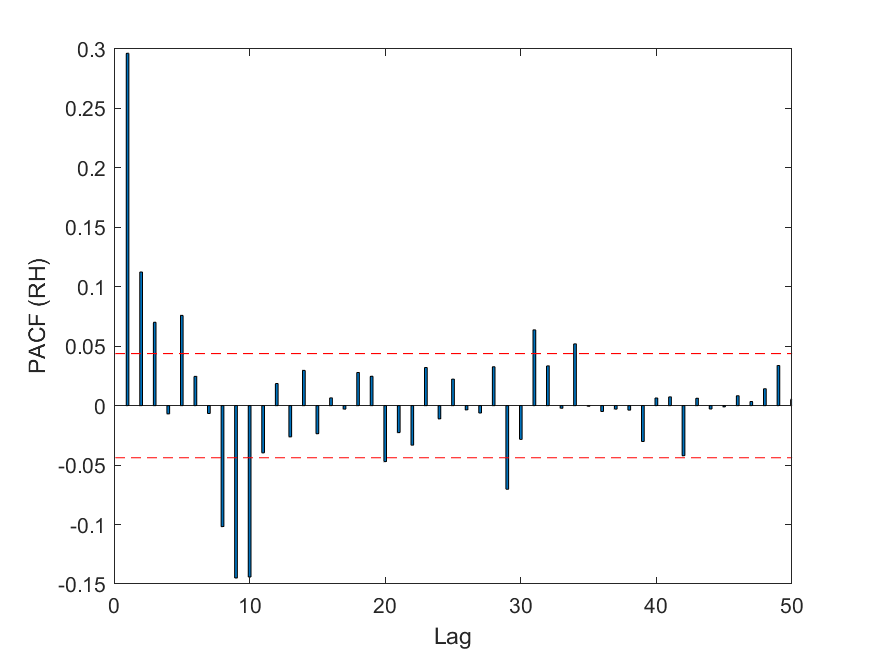}
		\caption{$\mathtt{AR(10)}$ }
		\label{figsout:AR10_PACF_d}
	\end{subfigure}

	%AR20

	\begin{subfigure}{.32\textwidth}
		\includegraphics[width=\textwidth]{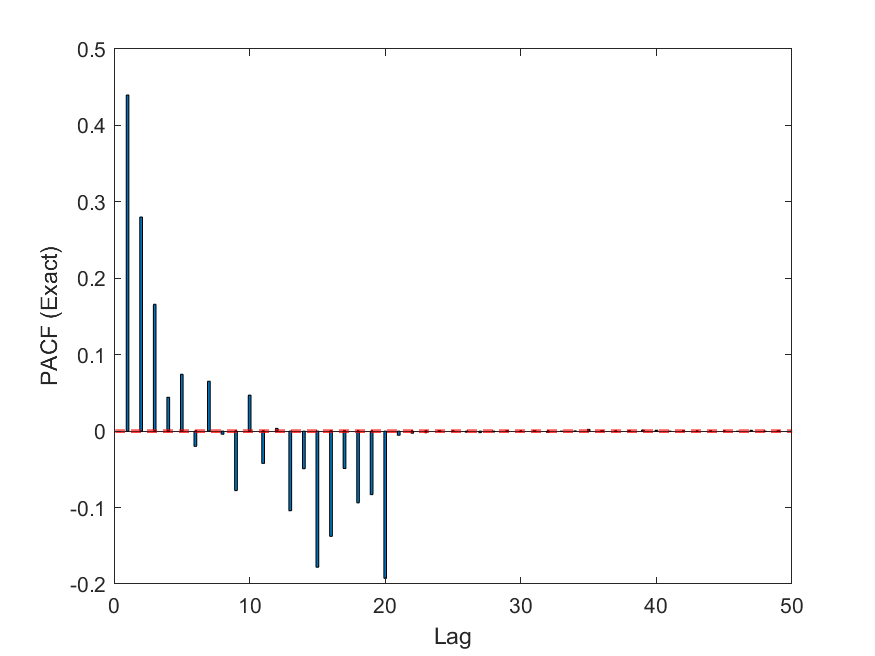}
		\caption{$\mathtt{AR(20)}$ }
		\label{figsout:AR20_PACF_n}
	\end{subfigure}
	\begin{subfigure}{.32\textwidth}
		\includegraphics[width=\textwidth]{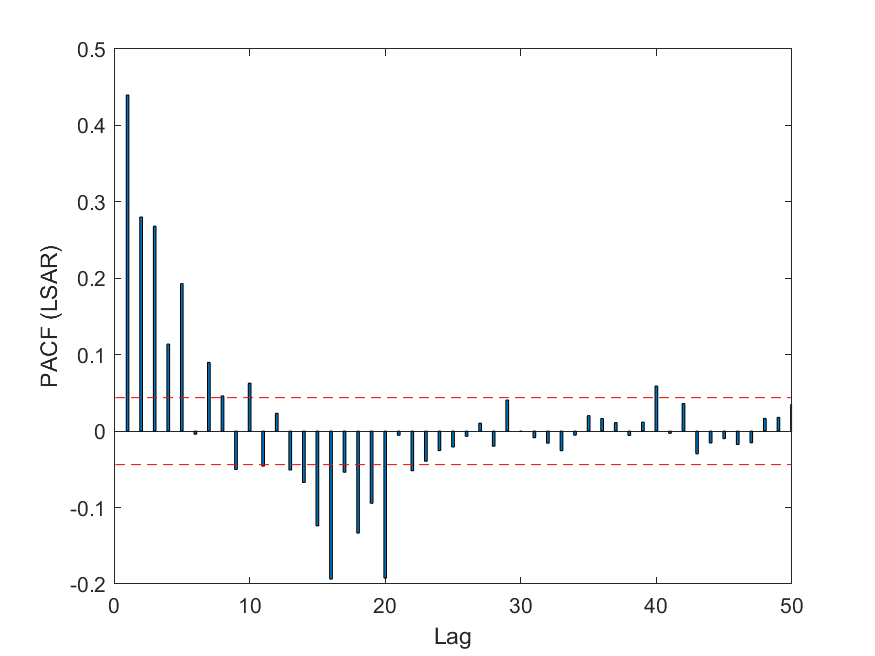}
		\caption{$\mathtt{AR(20)}$ }
		\label{figsout:AR20_PACF_s}
	\end{subfigure}
	\begin{subfigure}{.32\textwidth}
		\includegraphics[width=\textwidth]{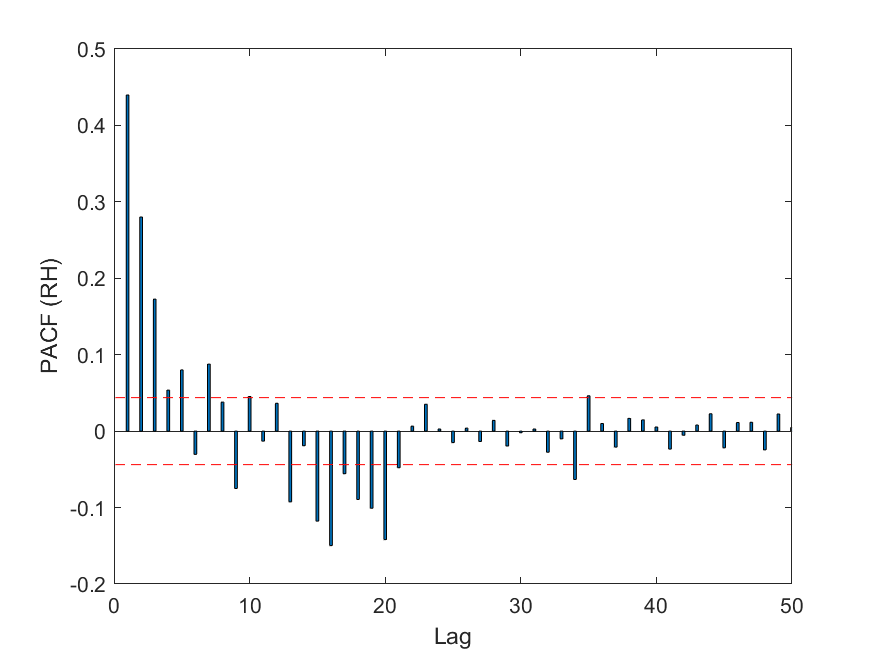}
		\caption{$\mathtt{AR(20)}$ }
		\label{figsout:AR20_PACF_d}
	\end{subfigure}
	\caption[PACF Plot Generated by \texttt{LSAR}, \texttt{RH} and Exact Algorithms on Synthetic Data With Outliers for \texttt{AR}$(5)$, \texttt{AR}$(10)$, \texttt{AR}$(20)$]{Figures (a) to (c), (d) to (f) and (g) to (i) correspond to randomly generated data from \texttt{AR}$(5)$, \texttt{AR}$(10)$ and \texttt{AR}$(20)$ models (with outliers) respectively. For each model we show the PACF plot computed exactly, by the \texttt{LSAR} algorithm and by the \texttt{RH} Algorithm. These are displayed from left to right, respectively.}
	\label{PACFYesOuts5-20}
\end{figure}

\begin{figure}[p]  % Fig 9
    \centering
	\begin{subfigure}{.32\textwidth}
		\includegraphics[width=\textwidth]{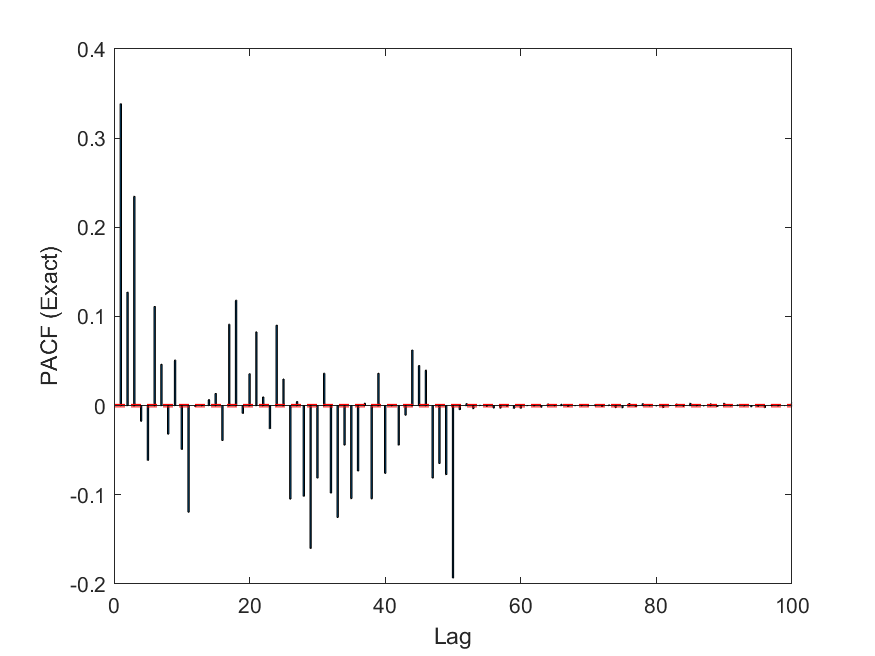}
		\caption{$\mathtt{AR(50)}$ }
		\label{figsout:AR50_PACF_n}
	\end{subfigure}
	\begin{subfigure}{.32\textwidth}
		\includegraphics[width=\textwidth]{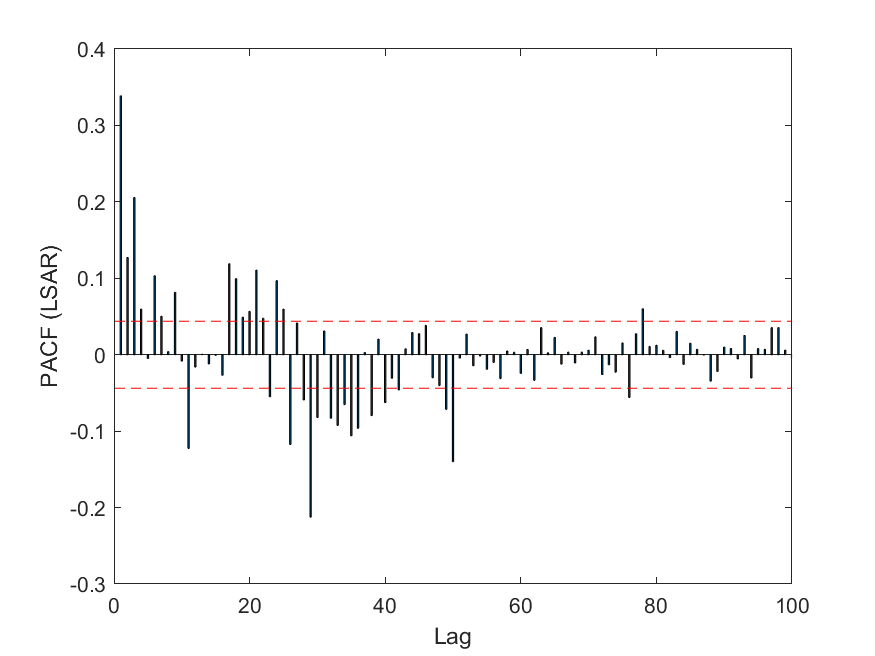}
		\caption{$\mathtt{AR(50)}$ }
		\label{figsout:AR50_PACF_s}
	\end{subfigure}
	\begin{subfigure}{.32\textwidth}
		\includegraphics[width=\textwidth]{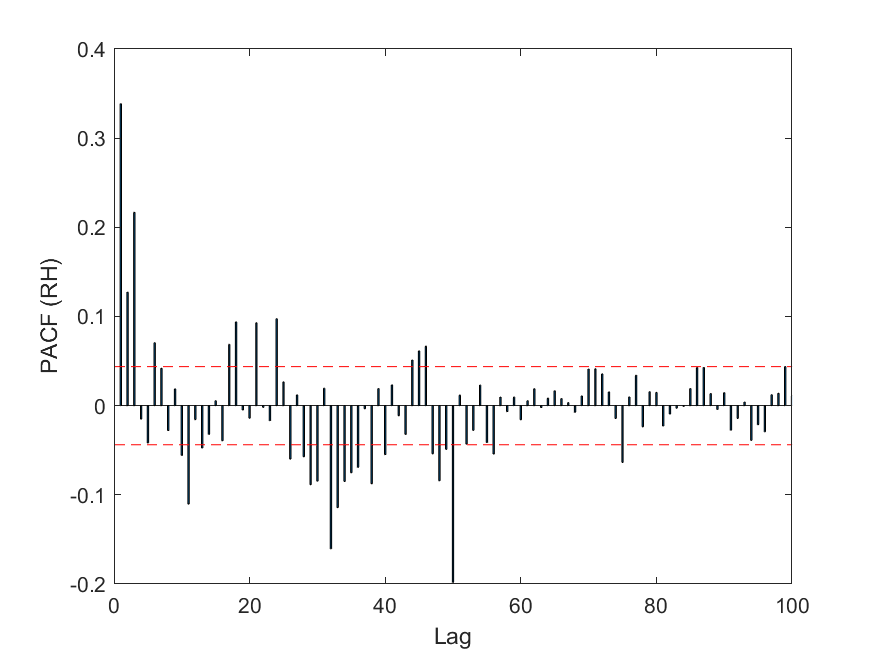}
		\caption{$\mathtt{AR(50)}$ }
		\label{figsout:AR50_PACF_d}
	\end{subfigure}

	%AR100

	\begin{subfigure}{.32\textwidth}
		\includegraphics[width=\textwidth]{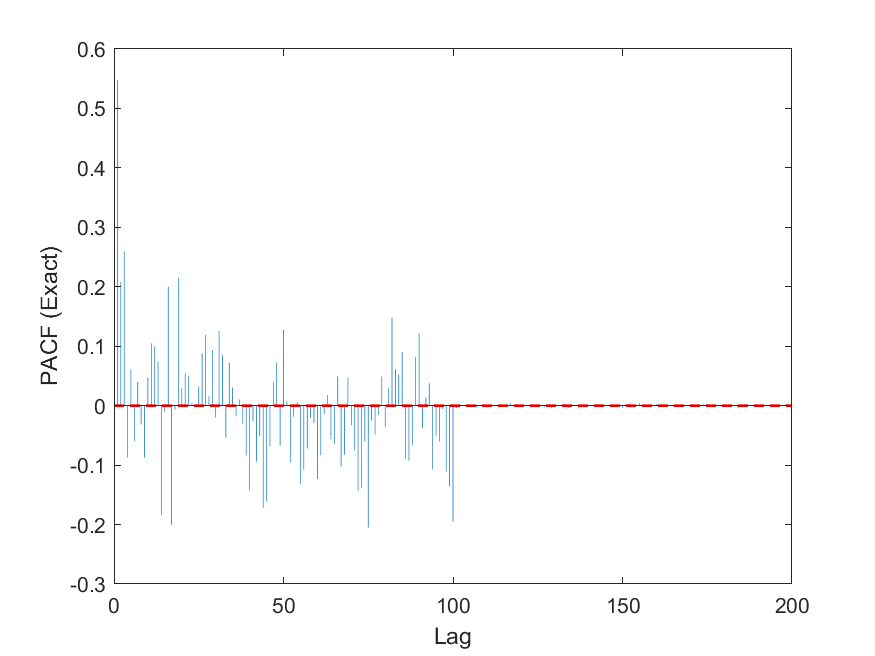}
		\caption{$\mathtt{AR(100)}$ }
		\label{figsout:AR100_PACF_n}
	\end{subfigure}
	\begin{subfigure}{.32\textwidth}
		\includegraphics[width=\textwidth]{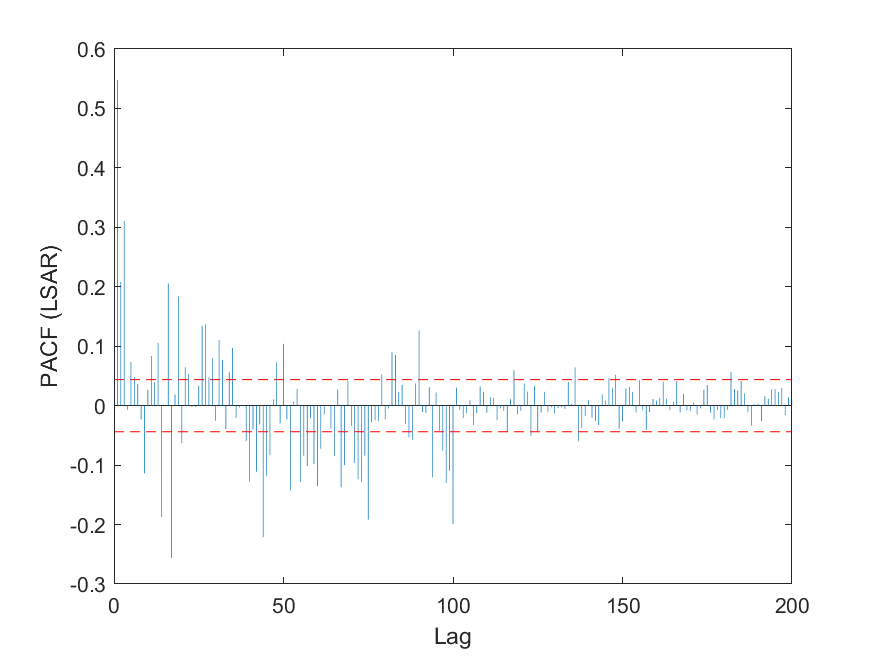}
		\caption{$\mathtt{AR(100)}$ }
		\label{figsout:AR100_PACF_s}
	\end{subfigure}
	\begin{subfigure}{.32\textwidth}
		\includegraphics[width=\textwidth]{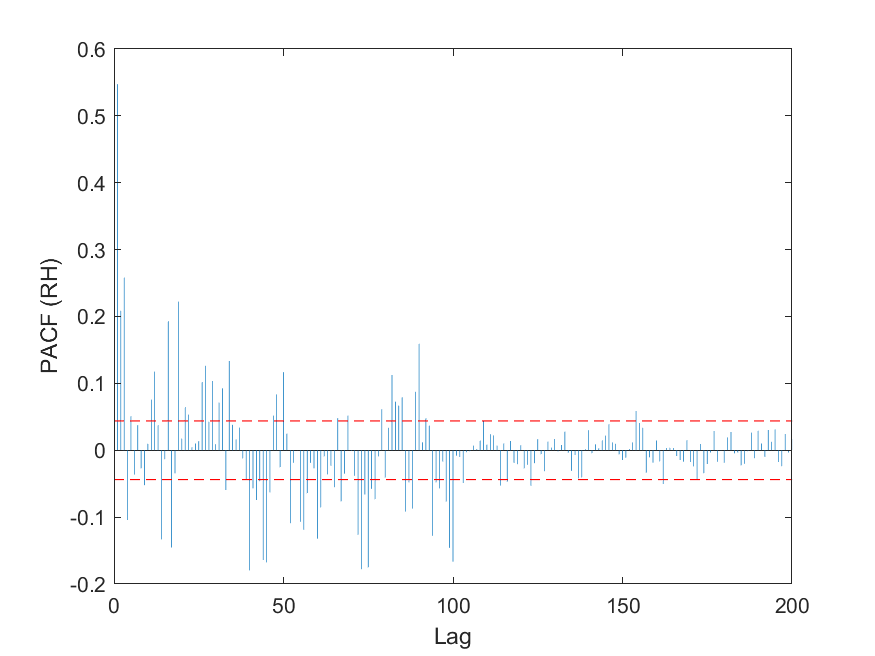}
		\caption{$\mathtt{AR(100)}$ }
		\label{figsout:AR100_PACF_d}
	\end{subfigure}

	%AR150

    \begin{subfigure}{.32\textwidth}
		\includegraphics[width=\textwidth]{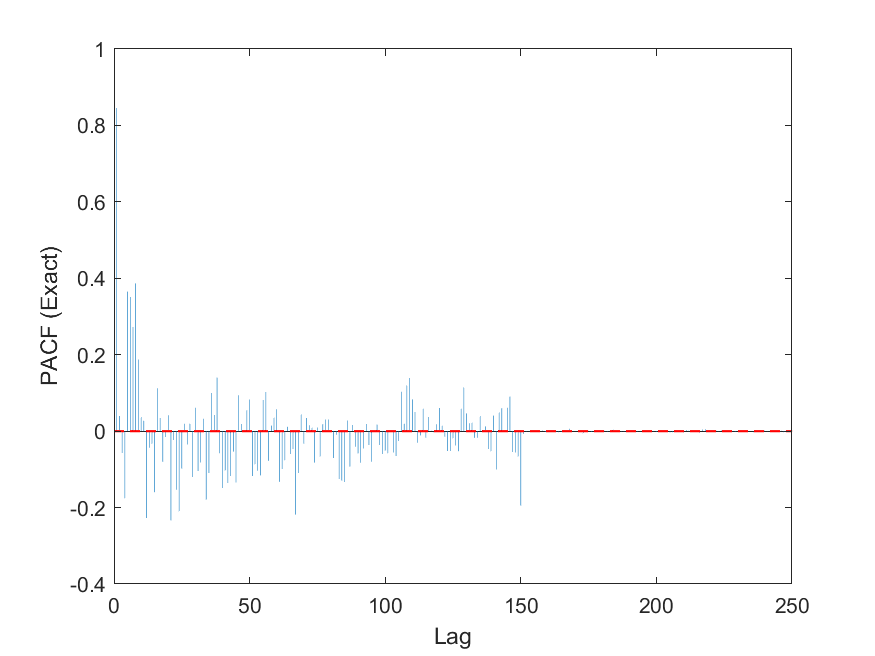}
		\caption{$\mathtt{AR(150)}$ }
		\label{figsout:AR150_PACF_n}
	\end{subfigure}
	\begin{subfigure}{.32\textwidth}
		\includegraphics[width=\textwidth]{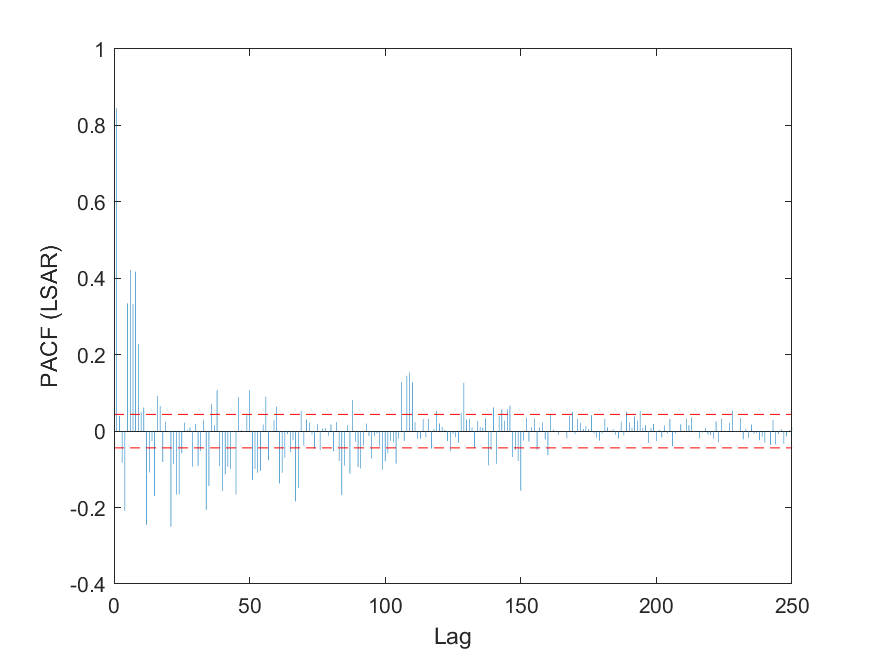}
		\caption{$\mathtt{AR(150)}$ }
		\label{figsout:AR150_PACF_s}
	\end{subfigure}
	\begin{subfigure}{.32\textwidth}
		\includegraphics[width=\textwidth]{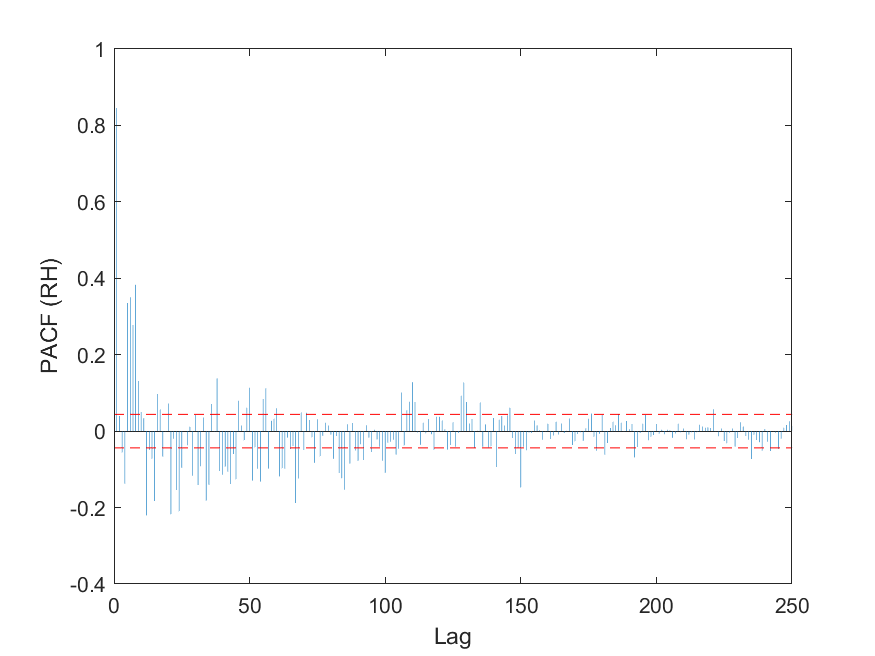}
		\caption{$\mathtt{AR(150)}$ }
		\label{figsout:AR150_PACF_d}
	\end{subfigure}
	\caption[PACF Plot Generated by \texttt{LSAR}, \texttt{RH} and Exact Algorithms on Synthetic Data With Outliers for \texttt{AR}$(50)$, \texttt{AR}$(100)$, \texttt{AR}$(150)$]{Figures (a) to (c), (d) to (f) and (g) to (i) correspond to randomly generated data from \texttt{AR}$(50)$, \texttt{AR}$(100)$ and \texttt{AR}$(150)$ models (with outliers) respectively. For each model we show the PACF plot computed exactly, by the \texttt{LSAR} algorithm and by the \texttt{RH} Algorithm. These are displayed from left to right, respectively. We are able use the PACF plots to correctly identify the order $p$ of the data sets.}
	\label{PACFYesOuts50-150}
\end{figure}

\cref{PACFYesOuts5-20,,PACFYesOuts50-150} display the PACF plots generated by each algorithm, for all synthetic data sets with included outliers. We estimate the PACF using the full data matrix and each of the compressed data matrices as we did in \cref{sec:PACFNoOuts}. For each \texttt{AR}$(p)$ model we use the same data sets from \cref{sec:TimeYesOuts}, with $ n = 2,000,000$ and number of sampled rows $c = 2,000$.
   
We are able use the PACF plots of the compressed algorithms to correctly identify the order $p$ of the data sets, notwithstanding some error. The plots were obtained using only $0.1\%$ of the data. As we saw in \cref{sec:TimeYesOuts}, the algorithms took significantly less time than the exact method, and were unaffected by outliers.

	%AR50

%-----------------------------------
\subsection{Real-world Data}
\label{sec:NRreal}
%-----------------------------------

We now test the quality and run time of the algorithms on some real-world data. We turn to data collected by Huerta et al. \cite{Huerta2016OnlineHA} studying the accuracy of electronic nose measurements. An electronic nose is an array of metal-oxide sensors capable of detecting chemicals in the air as a way of mimicking how a human or animal nose works. In this study the nose was constructed from eight different metal-oxide sensors, as well as humidity and temperature sensors. Measurements from each of these sensors were taken simultaneously at a rate of one observation per second for a period of almost two years in one of the author's home. Huerta et al.\ were able to use a statistical model utilising measurements from the nose to discriminate between different gasses with an R-squared close to 1.

The data was obtained from the UCI machine learning repository \cite{huerta2016UCI}. We look specifically at measurements of the eighth metal-oxide sensor (column R8 in the data set). The data set has $n = 919,438$ observations, and we must transform the data by taking the logarithm and difference in one lag to obtain a stationary data set.

As we have for our synthetic data sets, we compare the run time, error and PACF plots for each of the \texttt{LSAR} algorithm, the \texttt{RH} algorithm and the exact calculation. The algorithms are run with $\bar p = 100$, and the number of rows sampled for each of the compressed algorithms was $s = 0.01n = 9194$.

\begin{figure}[t]  % Fig 10
	\centering
	\begin{subfigure}{.49\textwidth}
		\includegraphics[width=\textwidth]{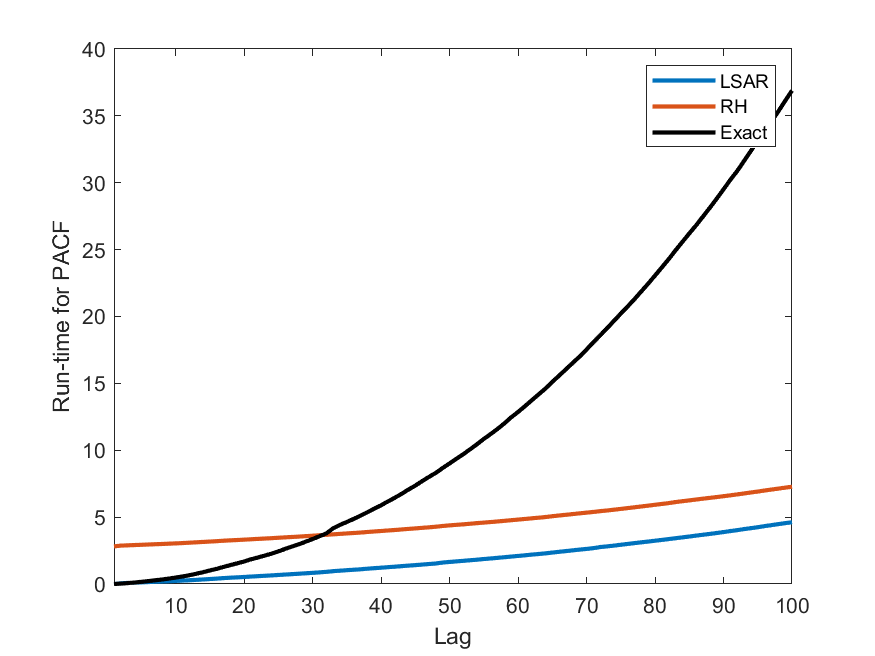}
		\caption{computation time}
		\label{fig:real_Time}
	\end{subfigure}
	\begin{subfigure}{.49\textwidth}
		\includegraphics[width=\textwidth]{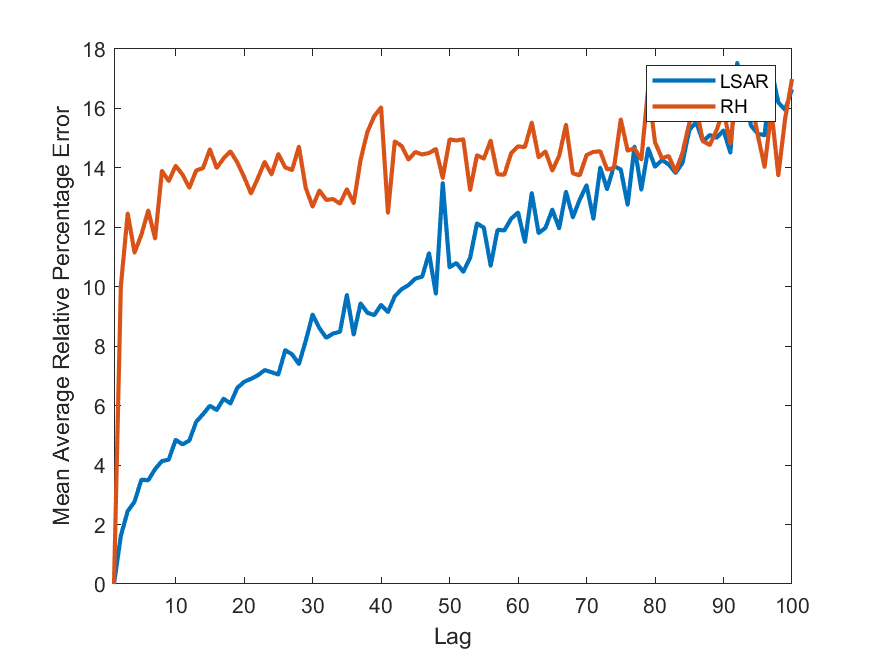}
		\caption{relative percentage error}
		\label{fig:real_Error}
	\end{subfigure}
	\caption[Computation Time and Relative Percentage Error of \texttt{LSAR}, \texttt{RH} and Exact Algorithms on Real Gas Sensor Data]{Figure (a) corresponding to the gas sensor data, shows the comparison between the computation time (in seconds) to generate the PACF for the \texttt{LSAR} algorithm (in blue), the \texttt{RH} algorithm (in red) and the exact computation of PACF (in black). Figure (b) shows the average relative percentage error over 50 runs of the algorithms, for each lag $h$}
	\label{fig:real}
\end{figure}

\cref{fig:real} displays the run time and the error of the estimated maximum likelihood error for the \texttt{LSAR} algorithm, the \texttt{RH} algorithm, and the exact calculation on the gas sensor data. 

To smooth out the error curves in \cref{fig:real_Error}, the algorithms were repeated 50 times and the mean of error at each lag was computed after excluding 5\% of the data values at each end of the data set. This was done to remove outliers.

The run times of all three algorithms, shown in \cref{fig:real_Time}, tell a similar story to that of \cref{sec:Time,sec:TimeYesOuts}. However, \cref{fig:real_Error} displays a different pattern of error from what we have previously seen.  Instead of steadily rising and being of similar magnitude to the \texttt{LSAR} algorithm, the \texttt{RH} algorithm jumps to 10\% error in the first lag before steadily rising. On the other hand, error in the parameters of \texttt{LSAR} algorithm is robust and consistent with the numerical results that we have presented in the synthetically generated data sections.

\begin{figure}[t]  % Fig 11
    \centering
	\begin{subfigure}{.32\textwidth}
		\includegraphics[width=\textwidth]{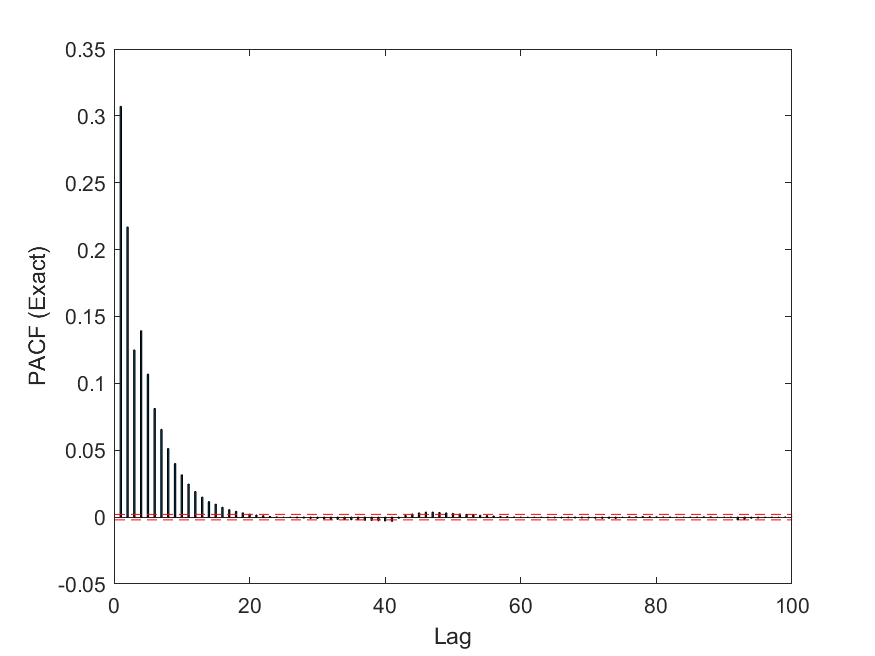}
		\caption{Exact}
		\label{figsout:Real_PACF_n}
	\end{subfigure}
	\begin{subfigure}{.32\textwidth}
		\includegraphics[width=\textwidth]{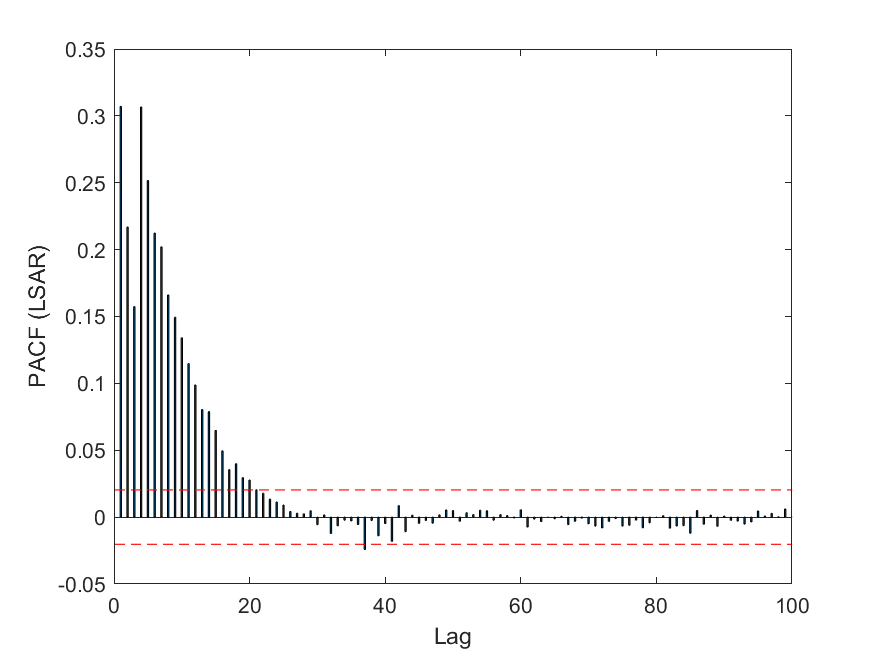}
		\caption{\texttt{LSAR}}
		\label{figsout:Real_PACF_s}
	\end{subfigure}
	\begin{subfigure}{.32\textwidth}
		\includegraphics[width=\textwidth]{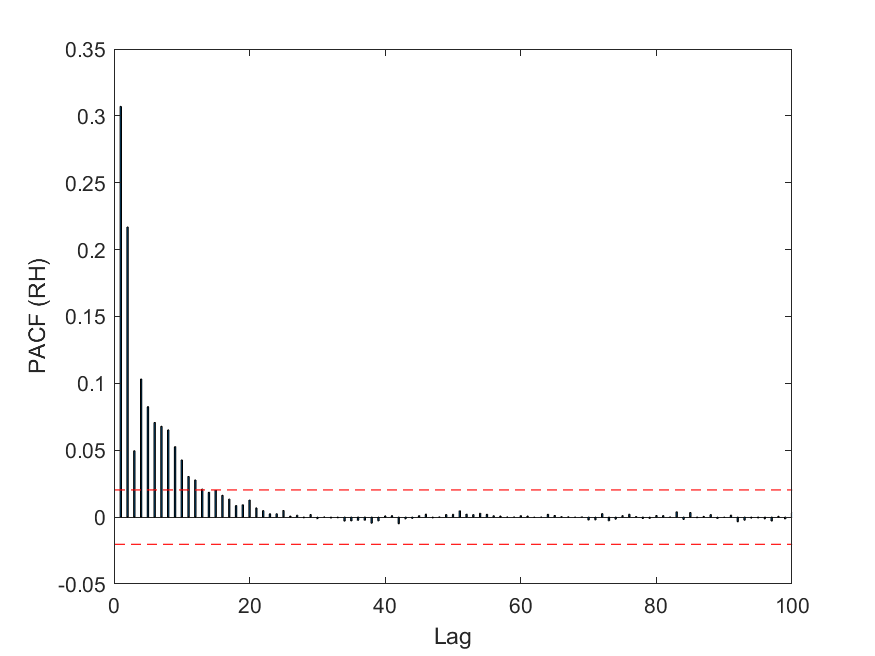}
		\caption{\texttt{RH}}
		\label{figsout:Real_PACF_d}
    \end{subfigure}
    \caption[PACF Plot Generated by \texttt{LSAR}, \texttt{RH} and Exact Algorithms on Gas Sensor Data]{Figures (a), (b) and (c) display the exact PACF plot, the PACF plot computed by the \texttt{LSAR} algorithm and the PACF plot computed using the \texttt{RH} algorithm respectively. Each plot corresponds to the gas sensor data of \cite{Huerta2016OnlineHA}.}
    \label{fig:RealPACF}
\end{figure}

This pattern of errors in the estimated parameters is reflected in the PACF plots produced by each of the algorithms (displayed in \cref{fig:RealPACF}). From \cref{figsout:Real_PACF_s}, the PACF plot generated by the \texttt{LSAR} algorithm excellently replicates the PACF plot of the exact algorithm, and it would seem that \texttt{AR}$(18)$ would be a good fit for this data set according to both the exact and \texttt{LSAR} algorithms. \cref{figsout:Real_PACF_d} on the other hand reaches the 95\% zero confidence bounds much earlier, suggesting that it would incorrectly estimate the order to fit to the data set.

%% file: 5Conclusion.tex
\section{\bf Conclusion}

We have examined the application of RandNLA to large time-series data. To do this we compared the \texttt{LSAR} and \texttt{RH} algorithms over a range of problem sizes with synthetic data and also some real data. As expected, because the algorithms solve subproblems with significantly smaller data matrices, the time to solve OLS problems associated with fitting an AR model was considerably reduced. In addition, the errors of the estimated model parameters were small and similar for both algorithms for each of the synthetic data sets. When applied to real time-series data, the two algorithms again had comparable run time; however, the \texttt{LSAR} algorithm elicited less error when estimating parameters.

The low error in the estimated parameters speaks to the utility of the algorithms for fitting an AR model. The AR fitting process involves two steps: estimating the order (from the PACF), then obtaining parameters of the model with best fitting order. Utility here refers to how the PACF plots generated by the algorithms provide it with accurate and usable information. For all synthetic models the PACF plot generated could be used to identify the order of the model, give or take some noise.

Overall, this paper displays the effectiveness of RandNLA in a time series context. We also see how the \texttt{LSAR} algorithm could provide a framework with which to adapt another Toeplitz least squares solver, the \texttt{RH} algorithm, to a time series context. Future work could look at adapting other Toeplitz least squares solvers such as those in \cite{Xi2014TOLS,van2003superfast} to a time series context and also compare the accuracy of these solvers when the problem involves ill-conditioned matrices.

%% file: main.bbl
\begin{thebibliography}{10}

\bibitem{abolghasemi2020ijpe}
M.~Abolghasemi, J.~Hurley, A.~Eshragh, and B.~Fahimnia.
\newblock Demand forecasting in the presence of systematic events: Cases in
  capturing sales promotions.
\newblock {\em International Journal of Production Economics}, 230:107892,
  2020.

\bibitem{avrachenkov2016aor}
K.~Avrachenkov, A.~Eshragh, and J.~Filar.
\newblock On transition matrices of {Markov} chains corresponding to
  hamiltonian cycles.
\newblock {\em Annals of Operations Research}, 243(1):19--35, 2016.

\bibitem{bean2015ajp}
N.~G. Bean, R.~Elliott, A.~Eshragh, and J.~V. Ross.
\newblock On binomial observation of continuous-time {M}arkovian population
  models.
\newblock {\em Journal of Applied Probability}, 52:457--472, 2015.

\bibitem{bean2016comminstat}
N.~G. Bean, A.~Eshragh, and J.~V. Ross.
\newblock Communications in statistics: Theory and methods.
\newblock {\em Annals of Operations Research}, 45(24):7161--7183, 2016.

\bibitem{clarkson2017low}
K.~L. Clarkson and D.~P. Woodruff.
\newblock Low-rank approximation and regression in input sparsity time.
\newblock {\em Journal of the ACM (JACM)}, 63(6):54, 2017.

\bibitem{cohen2015uniform}
M.~B. Cohen, Y.~T. Lee, C.~Musco, R.~Peng, and A.~Sidford.
\newblock Uniform sampling for matrix approximation.
\newblock In {\em Proceedings of the 2015 Conference on Innovations in
  Theoretical Computer Science}, pages 181--190, 2015.

\bibitem{drineas2012fast}
P.~Drineas, M.~Magdon-Ismail, M.~W. Mahoney, and D.~P. Woodruff.
\newblock Fast approximation of matrix coherence and statistical leverage.
\newblock {\em Journal of Machine Learning Research}, 13(Dec):3475--3506, 2012.

\bibitem{drineas2016randnla}
P.~Drineas and M.~W. Mahoney.
\newblock {RandNLA:} randomized numerical linear algebra.
\newblock {\em Communications of the ACM}, 59(6):80--90, 2016.

\bibitem{drineas2017lecture}
P.~Drineas and M.~W. Mahoney.
\newblock Lectures on randomized numerical linear algebra.
\newblock {\em CoRR}, abs/1712.08880, 2017.

\bibitem{eshragh2015modsim}
A.~Eshragh.
\newblock {F}isher information, stochastic processes and generating functions.
\newblock In {\em Proceedings of the 21$^{st}$ International Congress on
  Modeling and Simulation}, 2015.

\bibitem{eshragh2020plosone}
A.~Eshragh, S.~Alizamir, P.~Howley, and E.~Stojanovski.
\newblock Modeling the dynamics of the {COVID-19} population in {Australia}:
  {A} probabilistic analysis.
\newblock {\em PLoS ON}, 15:e0240153, 2020.

\bibitem{eshragh2011mor}
A.~Eshragh and J.~Filar.
\newblock Hamiltonian cycles, random walks and the geometry of the space of
  discounted occupational measures.
\newblock {\em Mathematics of Operations Research}, 36(2):258--270, 2011.

\bibitem{eshragh2011aor}
A.~Eshragh, J.~Filar, and M.~Haythorpe.
\newblock A hybrid simulation-optimization algorithm for the {H}amiltonian
  cycle problem.
\newblock {\em Annals of Operations Research}, 189:103--125, 2011.

\bibitem{eshragh2011es}
A.~Eshragh, J.~Filar, and A.~Nazari.
\newblock A projection-adapted {C}ross {E}ntropy ({PACE}) method for
  transmission network planning.
\newblock {\em Energy Systems}, 2(2):189--208, 2011.

\bibitem{eshragh2021enmo}
A.~Eshragh, B.~Ganim, T.~Perkins, and K.~Bandara.
\newblock The importance of environmental factors in forecasting {A}ustralian
  power demand.
\newblock {\em Environmental Modeling \& Assessment}, 2021.

\bibitem{eshragh2021rollage}
A.~Eshragh, G.~Livingston, T.~M. McCann, and L.~Yerbury.
\newblock {R}ollage: {E}fficient rolling average algorithm to estimate {ARMA}
  models for big time series data.
\newblock {\em arXiv preprint arXiv:2103.09175}, 2021.

\bibitem{eshragh2019lsar}
A.~Eshragh, F.~Roosta, A.~Nazari, and M.~W. Mahoney.
\newblock {LSAR}: Efficient leverage score sampling algorithm for the analysis
  of big time series data.
\newblock {\em arXiv preprint arXiv:1911.12321}, 2019.

\bibitem{fahimnia2018cor}
B.~Fahimnia, H.~Davarzani, and A.~Eshragh.
\newblock Performance comparison of three meta-heuristic algorithms for
  planning of a complex supply chain.
\newblock {\em Computers and Operations Research}, 89:241--252, 2018.

\bibitem{fahimnia2015ijpe}
B.~Fahimnia, J.~Sarkis, A.~Choudhary, and A.~Eshragh.
\newblock Tactical supply chain planning under a carbon tax policy scheme: {A}
  case study.
\newblock {\em International Journal of Production Economics}, 164:206--215,
  2015.

\bibitem{fahimnia2015omega}
B.~Fahimnia, J.~Sarkis, and A.~Eshragh.
\newblock Trade-off model for green supply chain planning: {A}
  leanness-versus-greenness analysis.
\newblock {\em International Journal of Production Economics}, 54:173--190,
  2015.

\bibitem{Hamilton1989}
J.~D. Hamilton.
\newblock A new approach to the economic analysis of nonstationary time series
  and the business cycle.
\newblock {\em Econometrica}, 57(2):357--384, 1989.

\bibitem{huerta2016UCI}
F.~Huerta and R.~Huerta.
\newblock Gas sensors for home activity monitoring data set.
\newblock
  \url{https://archive.ics.uci.edu/ml/datasets/Gas+sensors+for+home+activity+monitoring},
  2016.

\bibitem{Huerta2016OnlineHA}
R.~A. Huerta, T.~S. Mosqueiro, J.~Fonollosa, N.~F. Rulkov, and
  I.~Rodr{\'i}guez-Luj{\'a}n.
\newblock Online humidity and temperature decorrelation of chemical sensors for
  continuous monitoring.
\newblock {\em Chemometrics and Intelligent Laboratory Systems},
  157(15):169--176, 2016.

\bibitem{mahoney2011randomized}
M.~W. Mahoney.
\newblock Randomized algorithms for matrices and data.
\newblock {\em Foundations and Trends{\textregistered} in Machine Learning},
  2011.

\bibitem{shi2019sublinear}
X.~Shi and D.~P. Woodruff.
\newblock Sublinear time numerical linear algebra for structured matrices.
\newblock {\em Proceedings of the AAAI Conference on Artificial Intelligence},
  33(01):4918--4925, Jul. 2019.

\bibitem{van2003superfast}
M.~Van~Barel, G.~Heinig, and P.~Kravanja.
\newblock {A} superfast method for solving {T}oeplitz linear least squares
  problems.
\newblock {\em Linear Algebra and its Applications}, 366:441--457, June 2003.

\bibitem{woodruff2014sketching}
D.~P. Woodruff.
\newblock Sketching as a tool for numerical linear algebra.
\newblock {\em Foundations and Trends{\textregistered} in Theoretical Computer
  Science}, 2014.

\bibitem{Xi2014TOLS}
Y.~Xi, J.~Xia, S.~Cauley, and V.~Balakrishnan.
\newblock Superfast and stable structured solvers for {T}oeplitz least squares
  via randomized sampling.
\newblock {\em SIAM Journal on Matrix Analysis and Applications}, 35, 01 2014.

\bibitem{ye2016Toeplitz}
K.~Ye and L.~Lim.
\newblock Every matrix is a product of {Toeplitz} matrices.
\newblock {\em Foundations of Computational Mathematics}, 16(3):577--598, 2016.

\end{thebibliography}
